\documentclass[10pt,twocolumn,letterpaper]{article}

\usepackage{cvpr}              

\usepackage{listings} 

\usepackage[pagebackref,breaklinks,colorlinks,allcolors=cvprblue]{hyperref}

\usepackage[utf8]{inputenc}
\usepackage[T1]{fontenc}
\usepackage[english]{babel}
\usepackage{csquotes}
\usepackage{amsmath, amssymb, amsfonts}
\usepackage{graphicx}
\usepackage{float}
\usepackage{url}
\usepackage{hyperref}
\usepackage{booktabs}
\usepackage{multirow}
\usepackage{adjustbox}
\usepackage{array}
\usepackage{caption}
\usepackage{subcaption}
\usepackage{wrapfig}
\usepackage{enumitem}
\usepackage{longtable}
\usepackage{makecell}
\usepackage{lipsum}
\usepackage{multirow}
\usepackage{pifont}

\usepackage{colortbl}
\usepackage{tikz}
\usetikzlibrary{matrix,backgrounds}
\usepackage{pgfplots}

\usepackage{tocloft}
\usepackage[capitalize]{cleveref}
\definecolor{lightgreen}{RGB}{225, 253, 209}
\definecolor{lightred}{RGB}{255, 200, 201}

\definecolor{bluelink}{RGB}{0,113,188}
\definecolor{greenlink}{RGB}{0,188,113}
\hypersetup{
    colorlinks=true,
    citecolor=green!93!black,
    linkcolor=black,
    urlcolor=bluelink
}

\usepackage[most]{tcolorbox}

\newcommand{\prompt}[2]{
    \vspace{-0.1cm}
    \begin{tcolorbox}[
        colback=white!90!gray,
        colframe=teal!60!black,
        arc=5pt,
        boxsep=5pt,
        left=10pt,
        right=10pt,
        top=2pt,
        bottom=2pt,
        boxrule=0.8pt,
        drop shadow=gray!50!white,
        enhanced jigsaw
    ]
        \paragraph{\textbf{\textit{Prompt #1:}}} #2
    \vspace{-0.1cm}
    \end{tcolorbox}
    \vspace{-0.3cm}
}
\NewDocumentCommand\emojidizzy{}{
        \raisebox{-.2ex}{\includegraphics[scale=0.1]{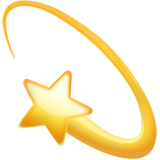}}
}
\NewDocumentCommand\emojidizzySmall{}{
        \raisebox{-.2ex}{\includegraphics[scale=0.07]{assets/dizzy_star.png}}
}
\NewDocumentCommand\dataset{}{
        \raisebox{-.2ex}{\includegraphics[scale=0.06]{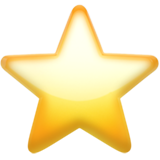}}
}
\NewDocumentCommand\benchmark{}{
        \raisebox{-.7ex}{\includegraphics[scale=0.15]{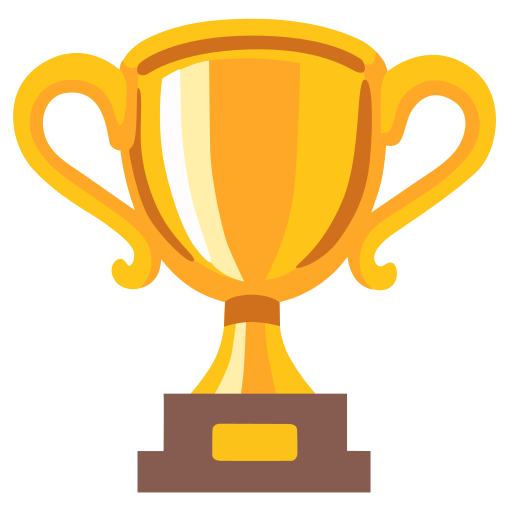}}
}
\definecolor{gold}{rgb}{0.84, 0.674, 0} 

\def\paperID{9042} 
\def\confName{CVPR}
\def\confYear{2025}

\title{\makebox[\textwidth][c]{\emojidizzy StarVector: Generating Scalable Vector Graphics Code from Images and Text}}

\author{
    Juan A. Rodriguez\textsuperscript{1,2,4} 
    Abhay Puri\textsuperscript{1} 
    Shubham Agarwal\textsuperscript{1, 2}
    Issam H. Laradji\textsuperscript{1, 5}
    Pau Rodriguez\textsuperscript{1}\\
    Sai Rajeswar\textsuperscript{1,2} 
    David Vazquez\textsuperscript{1}
    Christopher Pal\textsuperscript{1,2,3} 
    Marco Pedersoli\textsuperscript{1,4}
}

\begin{document}

\twocolumn[{
\renewcommand\twocolumn[1][]{#1}%
\maketitle

\begin{center}
\vspace{-2em}
\textsuperscript{1}ServiceNow Research 
\textsuperscript{2}Mila - Quebec AI Institute 
\textsuperscript{3}Canada CIFAR AI Chair\\
\textsuperscript{4}ÉTS, Montréal, Canada 
\textsuperscript{5}UBC, Vancouver, Canada \\


\texttt{\textcolor{gold}{\textbf{https://starvector.github.io/}}}
\end{center}

\includegraphics[width=1.0\textwidth]{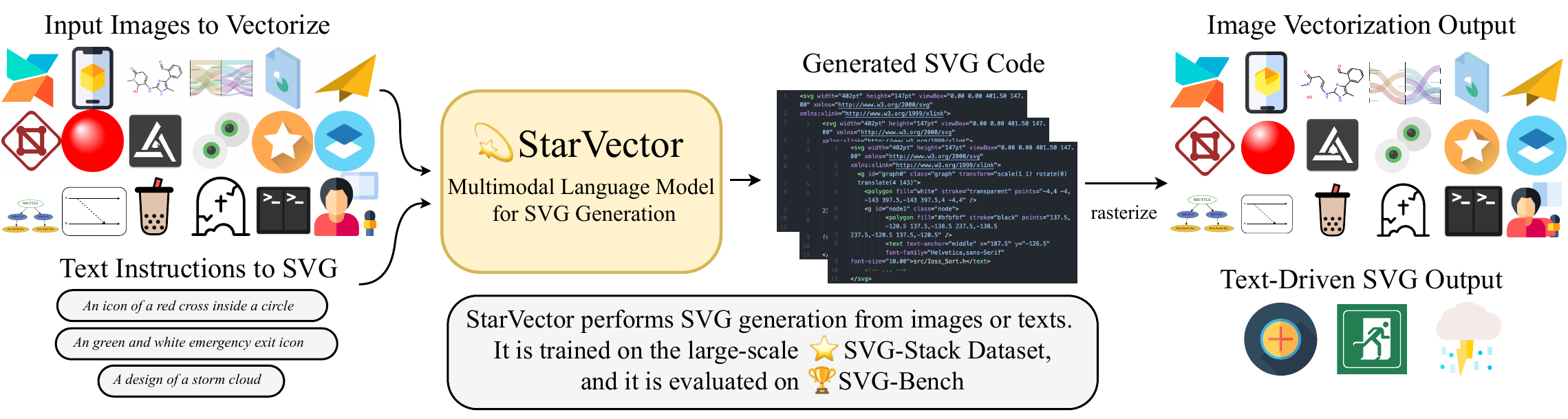}
\captionof{figure}{\textbf{StarVector: A foundation model for SVG generation.} StarVector’s multimodal architecture allows input from raster images or text instructions. It converts a variety of raster visuals, including icons, logos, and technical diagrams, into vector graphics or generates new SVGs from text. (Left) Inputs: raster images and text. (Right) Outputs: vectorized images (SVG)}
\vspace{1em}
\label{fig:teaser}
}]

\begin{abstract}
Scalable Vector Graphics (SVGs) are vital for modern image rendering due to their scalability and versatility. Previous SVG generation methods have focused on curve-based vectorization, lacking semantic understanding, often producing artifacts, and struggling with SVG primitives beyond \textit{path} curves. To address these issues, we introduce StarVector, a multimodal large language model for SVG generation. It performs image vectorization by understanding image semantics and using SVG primitives for compact, precise outputs. Unlike traditional methods, StarVector works directly in the SVG code space, leveraging visual understanding to apply accurate SVG primitives. To train StarVector, we create SVG-Stack, a diverse dataset of 2M samples that enables generalization across vectorization tasks and precise use of primitives like ellipses, polygons, and text. We address challenges in SVG evaluation, showing that pixel-based metrics like MSE fail to capture the unique qualities of vector graphics. We introduce SVG-Bench, a benchmark across 10 datasets, and 3 tasks: Image-to-SVG, Text-to-SVG generation, and diagram generation. Using this setup, StarVector achieves state-of-the-art performance, producing more compact and semantically rich SVGs.
\end{abstract}    
\section{Introduction}
\label{sec:intro}

\definecolor{tagcolor}{rgb}{0.75, 0.13, 0.13}
\definecolor{attrcolor}{rgb}{0.25, 0.5, 0.9}
\definecolor{valcolor}{rgb}{0.12, 0.55, 0.12}
\definecolor{bordercolor}{rgb}{0.5, 0.5, 0.5} 
\definecolor{bgcolor}{rgb}{0.95, 0.95, 0.95} 

\lstset{
    language=XML,
    basicstyle=\ttfamily\tiny, 
    keywordstyle=\color{tagcolor}\bfseries, 
    identifierstyle=\color{attrcolor},
    stringstyle=\color{valcolor},
    morekeywords={svg, rect, polygon, ellipse, circle, line, path}, 
    breaklines=true,
    breakatwhitespace=true, 
    postbreak=\mbox{}, 
    rulecolor=\color{bordercolor}, 
    columns=fullflexible
}

\begin{figure*}[ht]
    \centering
    \begin{tabular}{cc}
        \begin{minipage}{0.5\textwidth}
            \includegraphics[width=\textwidth]{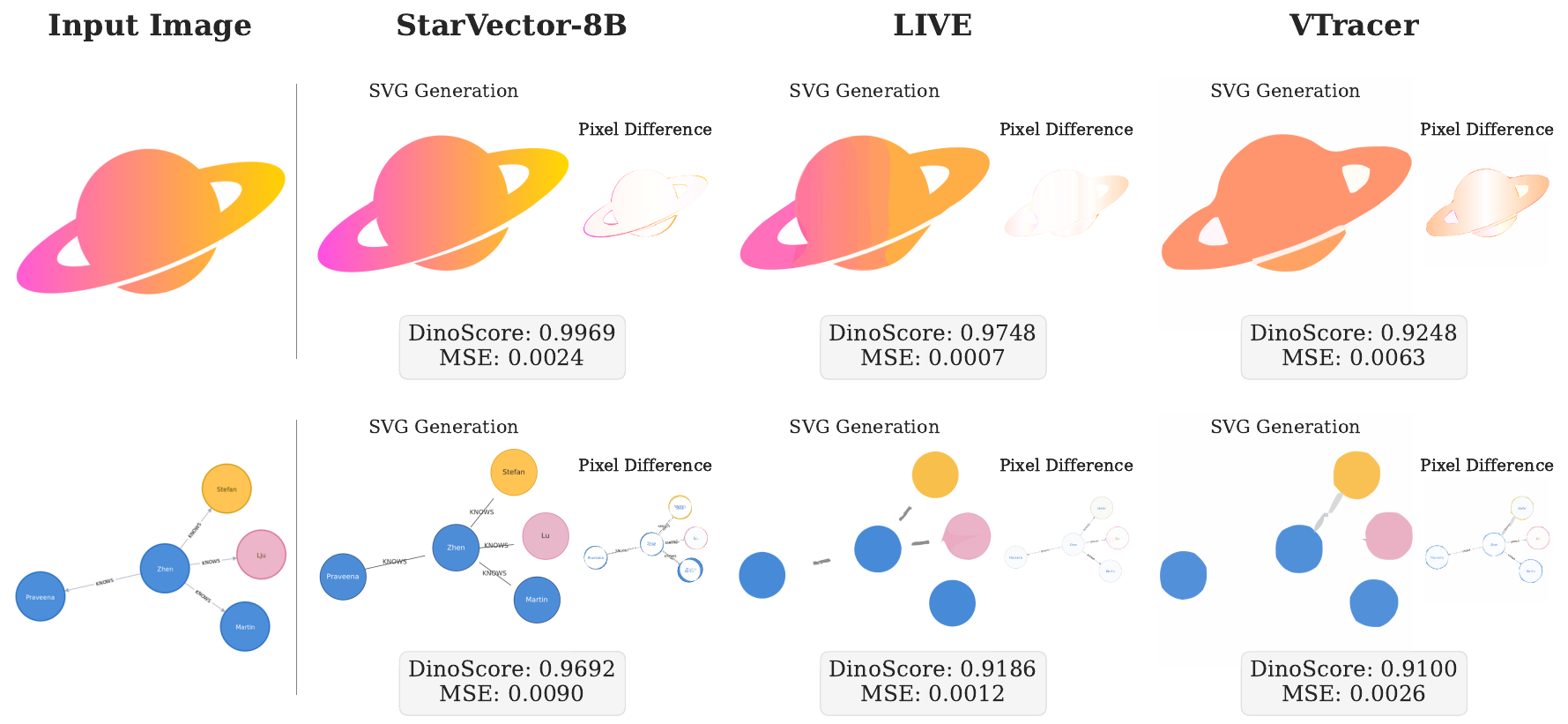} 
        \end{minipage}
        &
        \begin{minipage}{0.5\textwidth}
            \hspace{-0.05\textwidth}
            \begin{tabular}{cc}
                \begin{minipage}{0.15\textwidth}
                \centering
                    \includegraphics[width=1.5cm]{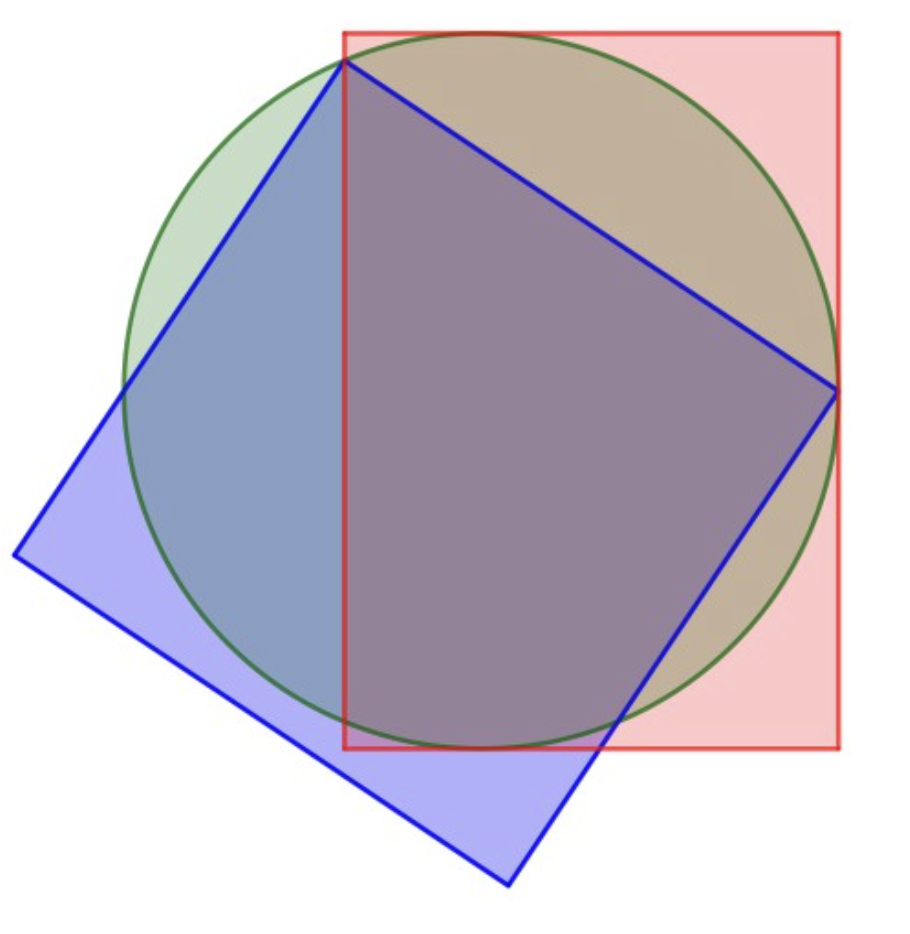}
                \end{minipage}
                 &
                 \hspace{-0.05\textwidth}
                 \begin{tabular}{c}
                \begin{minipage}{0.85\textwidth}
                \tiny\textbf{StarVector-8B}
                \vspace{-0.2cm}
                    \lstinputlisting[firstline=1, lastline=10]{figures/results/primitives-analysis/svg/sample3_starvector_small.svg}
                \end{minipage}
                \\
                \begin{minipage}{0.85\textwidth}
                \tiny\textbf{VTracer}
                \vspace{-0.2cm}
                    \lstinputlisting[firstline=1, lastline=10]{figures/results/primitives-analysis/svg/sample3_vtracer_small.svg}
                \end{minipage} \\
                \end{tabular}
            \end{tabular}
        \end{minipage}
    \end{tabular}
    \caption{\textbf{(Left) Image Vectorization} results using StarVector-8B, LIVE, and VTracer. Each row shows the input image, generated SVGs, and pixel-wise difference maps to highlight accuracy. StarVector-8B better preserves shapes, color gradients, and text, despite minor misplacements. Notably, \emph{MSE often misaligns with visual quality}, e.g., regarding the `planet' example, StarVector’s MSE (0.009) is higher than LIVE’s (0.0012) and VTracer’s (0.0039), yet StarVector preserves the color gradient. For the `diagram' example, StarVector preserves the text. DinoScore better reflects these details, consistently favoring StarVector. \textbf{(Right) Curve vs Primitive-based Vectorization.} SVG code generated by StarVector and VTracer for the given image. StarVector effectively leverages shape primitives, resulting in a compact vectorization. VTracer decomposes the image into numerous paths, resulting in a more complex result with less semantic clarity.}
\label{fig:mse-bad}
\end{figure*}

Vector graphics represent an archetypal form of image representation, where visual compositions are constituted by scalable primitive shapes~\citep{li2020differentiable, ma2022towards, ma2022towards, jain2023vectorfusion}. For modern image rendering, Scalable Vector Graphics (SVGs)~\citep{quint2003scalable} have become the standard for representing vector graphics. The SVG format~\citep{ferraiolo2000scalable} provides a comprehensive set of primitives and styling options. At its core, the \textit{path} represents basic curves ~\citep{quint2003scalable}. Combined with primitives like \textit{polygon} or \textit{ellipse}, SVGs define complex designs precisely.

The task of \textit{image vectorization}, i.e., converting pixel-based raster images into SVGs, stands as a fundamental challenge in vector graphics. The main challenge lies in developing methods that generalize across diverse domains, from fonts and logos to complex illustrations and diagrams~\cite{rodriguez2023ocr, rodriguez2023figgen, belouadi2023automatikz, belouadi2024detikzify}. Traditional approaches often rely on approximating images through multiple \textit{paths}~\citep{vtracer, autotrace, potrace, ma2022towards, li2020differentiable}. This strategy can be inefficient as shown in Fig. \ref{fig:mse-bad}~(Right). For instance, a circle shape could be represented as long \textit{path} or, more precisely and compactly, as a single \texttt{<circle/>} primitive. Similarly, text elements should be vectorized as editable \texttt{<text/>} primitives to retain the original textual content. This balance between curve-based shape approximation and accurately recognizing primitives has been previously unexplored and remains a core challenge in modern vectorization.

Vectorization approaches fall into two main categories: traditional image processing and deep learning (DL)-based methods. Image processing techniques~\citep{vtracer, autotrace, potrace} trace vector curves via pixel-level analysis but often generate overly complex representations with artifacts and lack semantic understanding (Figure~\ref{fig:mse-bad}). DL methods~\citep{carlier2020deepsvg, reddy2021im2vec, cao2023svgformer, wang2021deepvecfont} improve vector modeling using latent variable models and differentiable rendering~\citep{li2020differentiable, ma2022towards} but struggle with generalization and underutilize SVG primitives. This limits their effectiveness for complex SVGs like scientific diagrams and hinders their application in modern multimodal tasks such as text-driven SVG generation~\citep{ramesh2021zeroshot, rombach2021highresolution, yu2022scaling, esser2024scaling}.

Recent advancements in Multimodal Large Language Models (MLLMs)~\citep{alayrac2022flamingo, liu2023visual} have integrated visual understanding into transformer~\citep{vaswani2017attention} architectures while demonstrating strong code generation capabilities~\citep{li2023starcoder, lozhkov2024starcoder, TheC3, openai2023gpt4}. Building on these developments, \emph{we introduce image vectorization as an inverse rendering and code generation task}, leveraging MLLMs to generate SVG code directly from input images. This approach naturally encompasses the full range of SVG primitives, enhancing both semantic understanding and generation capabilities (Table~\ref{tab:related-work_comparison}).

We introduce \emojidizzySmall StarVector, a foundational MLLM for SVG generation. StarVector processes both images and text instructions to produce compilable SVG code, leveraging SVG primitives to accurately represent vector graphics. We build upon the \textbf{Star}Coder works~\citep{li2023starcoder, lozhkov2024starcoder} to connect the code generation research with SVG generation. Figure~\ref{fig:starvector} describes the model architecture. It integrates an image encoder that projects images into visual tokens, and a transformer language model for learning the relationships between instructions, visual features, and SVG code sequences, to perform image vectorization (Image-to-SVG) or text-driven SVG Generation (Text-to-SVG) tasks. StarVector performs primitive-aware vectorization through learned semantic understanding, effectively leveraging SVG primitives without explicit pixel reconstruction objectives. To address the lack of large-scale SVG datasets for training StarVector, we introduce SVG-Stack, with over 2M SVG samples paired with rendered images and text descriptions. 

Additionally, we find that conventional metrics like MSE fail to adequately assess vector graphics quality, as demonstrated in Figure~\ref{fig:mse-bad}. Instead, we propose DinoScore, a perceptual similarity metric that better correlates with visual human perception, and introduce SVG-Bench, an evaluation framework spanning 10 datasets and 3 tasks: Image-to-SVG, Text-to-SVG, and diagram generation.
\paragraph{Contributions}
\begin{enumerate}
    \item We introduce \textbf{\emojidizzySmall StarVector}, an MLLM capable of image vectorization and text-driven SVG Generation, uniquely preserving SVG primitives rather than producing multiple curves—a previously unexplored skill.
    \item We create \textbf{\dataset SVG-Stack}, a large-scale dataset with 2M samples, supporting Image-to-SVG and Text-to-SVG. 
    \item We develop \textbf{\benchmark SVG-Bench}, an MLLM benchmark with 10 datasets across 3 SVG tasks.
    \item We conduct extensive experiments and evaluations, including human assessments, demonstrating StarVector's advantages in primitive recognition and highlighting the limitations of pixel-wise metrics for SVG evaluation.
\end{enumerate}

\section{Related Work}
\label{sec:related}
\noindent\textbf{SVG Generation Methods.} Early image vectorization efforts~\citep{wiki:Comparison_of_raster-to-vector_conversion_software} primarily relied on traditional techniques~\citep{liao2012subdivision, xia2009patch, diebel2008bayesian} such as segmentation and polynomial curve fitting~\citep{li2020differentiable, vtracer, potrace, autotrace}. With advancements in deep learning, new approaches have emerged. For instance, SVG-VAE~\citep{lopes2019learned} employs a class-conditional Variational Autoencoder (VAE)~\citep{kingma2013auto} to predict a latent style vector and generate SVGs using an LSTM decoder~\citep{hochreiter1997long}. DeepSVG~\citep{carlier2020deepsvg} introduces a hierarchical VAE architecture using transformers for SVG path representation, while Im2Vec~\citep{reddy2021im2vec} converts pixel images into latent representations that can be decoded into paths with a recurrent neural network. However, these methods are limited to basic primitives like paths, resulting in a performance gap compared to traditional image processing methods. Concurrently, BeyondPixels~\citep{zhang2023pixelsexploringhumanreadablesvg} utilizes LLMs to predict command sequences in synthetic SVGs. These trends highlight the potential of LLM models for improved generalization in SVG generation.

Recent results in image generation using diffusion~\citep{ho2020denoising, rombach2022high} or autoregressive~\citep{esser2021taming, ramesh2021zero, yu2022scaling} models and the success of text-to-image generation~\citep{ramesh2021zeroshot, rombach2021highresolution, balaji2022eDiff-I} have inspired the research into text-conditioned SVG generation. VectorFusion~\citep{jain2023vectorfusion} leverages a strong text-to-image diffusion model to find the SVG via iterative optimization. Some work has focused on SVG editing from textual inputs~\citep{cai2023leveraging}. CLIPasso~\citep{vinker2022clipasso} uses a CLIP distance loss to iteratively refine SVG from sketches. These solutions are slow due to their iterative nature. IconShop~\citep{wu2023iconshop} trains a BERT~\citep{devlin2018bert} model for text-conditioned SVG generation of icons, but their method is restricted to using paths. 

In contrast, we train an MLLM on the inverse rendering image vectorization task, leveraging visual understanding to produce accurate SVG code with optimal use of primitives.

\vspace{2mm}
\noindent\textbf{SVG Benchmarks and Datasets.} Previous work on SVG datasets and benchmarks has been limited. Existing Image-to-SVG datasets mainly focus on fonts, icons, and emojis~\citep{carlier2020deepsvg, reddy2021im2vec, cao2023svgformer, wu2023iconshop, cai2023leveraging, wang2021deepvecfont}, offering limited variety for broader SVG types and primitives. Recent text-driven SVG generation efforts~\citep{Cloutre2019FIGRFI, wu2023iconshop, cai2023leveraging} leverage large language models~\cite{openai2023gpt4v} for synthetic captions, but limited dataset accessibility hampers reproducibility. Evaluation methods also face challenges. Existing pixel-based metrics like MSE~\cite{carlier2020deepsvg, reddy2021im2vec, lopes2019learned} fail to capture the fidelity and structure of SVGs, overlooking aspects like line definition and primitive usage (Figure~\ref{fig:mse-bad}, illustrates how MSE can be misleading). Recent benchmarking efforts~\citep{nishina2024svgeditbench, zou2024vgbench} focus on caption-based generation or editing but remain limited in scope.

To address these gaps, we introduce SVG-Stack, a large-scale dataset for diverse SVG generation tasks from both images and text, and SVG-Bench, a unified benchmark with datasets, tasks, and metrics tailored for SVGs. Together, they provide a solid foundation for improving SVG model training and evaluation.

\vspace{2mm}
\noindent\textbf{Multimodal Language Models.} Large language models (LLMs) have achieved great success in natural language processing (NLP)~\citep{vaswani2017attention, radford2018improving, brown2020language, openai2023gpt4, touvron2023llama, touvron2023llama2}, especially in code generation tasks~\cite{bisong2019google, husain2019codesearchnet, gao2020pile, kocetkov2022stack, dakhel2023github, chen2021evaluating, nijkamp2022codegen, li2023starcoder}. Recent trends in Multimodal Large Language Modeling (MLLM)~\citep{alayrac2022flamingo, liu2023visual, openai2023gpt4v, wang2024qwen2} have allowed infusing image understanding into text-only models, for tasks like visual-question answering (VQA)~\citep{antol2015vqa} or image captioning~\citep{li2022mplug, li2022blip, li2023blip}. Current approaches define an image encoder for computing visual tokens from images, that can be processed by an LLM~\citep{alayrac2022flamingo, liu2023visual, openai2023gpt4v, wang2024qwen2}. Multiple works use a Vision Transformer~\citep{dosovitskiy2020image} backbone with pre-trained weights like CLIP~\citep{radford2021learning}. We get inspiration from these architectures for training an MLLM on the tasks of image vectorization and text-driven SVG Generation (also referred to as Image-to-SVG and Text-to-SVG).

\section{\dataset SVG-Stack Dataset}
\label{svg-stack}
\begin{figure*}[!ht]
    \centering
    \includegraphics[width=1.0\linewidth]{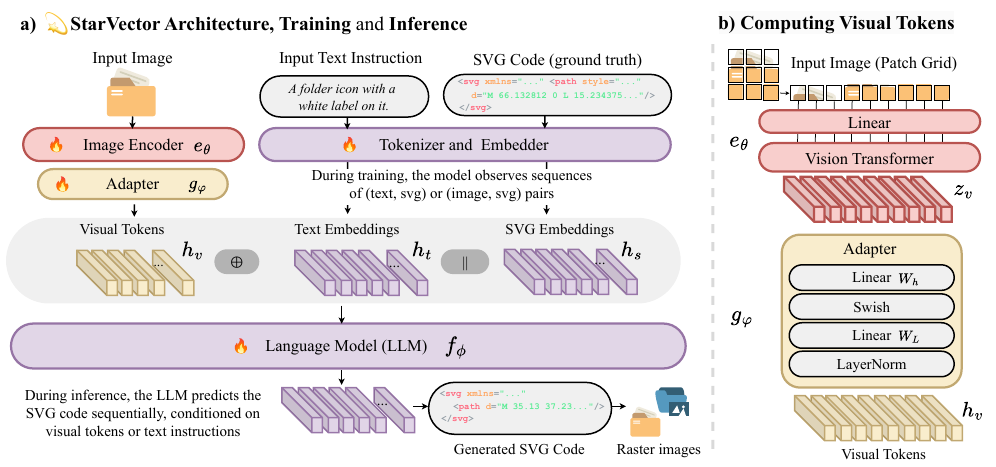}
    \caption{\textbf{a) StarVector Architecture}. Images are projected into visual tokens via an Image Encoder and Adapter, aligned with the Language Model’s hidden space. Text conditioning uses the LLM’s tokenizer and embedder. $\oplus$ denotes mutually exclusive addition of image or text features to SVG tokens, while $\parallel$ indicates sequence concatenation. During training, the model maps token sequences (visual or textual) to SVG code, and at inference, SVG code is generated sequentially. \textbf{b) Visual Token Computation}. We use a Vision Transformer (ViT) to process image patches into hidden features, which are projected through a non-linear Adapter to form visual tokens.}
    \label{fig:starvector} 
\end{figure*}
To address the lack of large-scale SVG datasets for training foundational SVG models, we introduce SVG-Stack, a dataset containing 2.1 million SVG samples for training, 108k for validation, and 5.7k for testing. Each SVG is paired with its corresponding raster image and descriptive text, making the dataset ideal for multi-modal learning. SVG-Stack is sourced from The Stack~\citep{kocetkov2022stack}, a diverse collection of code samples from various software languages. Our selection builds upon the initial filtering and de-duplication processes conducted in~\citep{kocetkov2022stack, allal2023santacoder, li2023starcoder}. We perform additional filtering to ensure non-duplicate SVG samples. 

These samples come from publicly available repositories on GitHub, providing a rich variety of real-world SVGs used across websites, graphic designs, and more. A key advantage of our approach is the inclusion of SVGs with diverse syntactic structures, varying templating approaches, different header formats, framework-specific implementations, and full support for SVG primitives. This diversity significantly enhances model generalization capabilities. This approach contrasts with previous datasets that primarily focused on a narrow subset of SVG types~\cite{lopes2019learned,carlier2020deepsvg, wu2023iconshop}. SVG-Stack represents the first large-scale pre-training dataset for SVG generation, with permissively licensed samples~\citep{kocetkov2022stack}, bringing together a broad spectrum of SVGs that closely mirror the diversity seen across the Web. Table~\ref{tab:all-datasets} and Figures~[\ref{fig:colored-datasets}, \ref{fig:simple-datasets}, \ref{fig:diagrams}] in Appendix~\ref{app:datasets} provide a comparison of available SVG datasets, showcasing the breadth of test examples found in SVG-Stack.

\vspace{2mm}
\noindent\textbf{Data Processing and Curation.} Our data processing begins with extracting SVG samples from TheStack~\citep{kocetkov2022stack}, followed by a comprehensive deduplication process based on filenames, SVG code, and metadata. We utilized CairoSVG~\citep{cairosvg} for rasterization, removing samples that produced completely white images. To optimize sequence length and improve visualization, we eliminated comments and XML headers from the SVG code.

\vspace{2mm}
\noindent\textbf{Synthetic Generation of Text Instructions.} To enable Text-to-SVG generation, we augment our 2M samples with synthetically generated textual captions describing the rasterized images. We employ open-source image captioning models, specifically BLIP2~\citep{li2023blip} and Llava~\citep{liu2023visual}, resulting in a comprehensive dataset of 4 million paired textual captions and SVGs. Detailed information about the annotation process is available in Appendix~\ref{app:nnotation-svg-stack}.

\vspace{2mm}
\noindent\textbf{Data Augmentation.} We implement several SVG-specific augmentation techniques to enhance model robustness. Rather than storing static-resolution raster images, we perform rasterization during data collation, enabling dynamic SVG transformations without significant training overhead. Our augmentation pipeline includes modifications to image resolution, rotation, translation, scaling, and color properties. We leverage open-source libraries svgpathtools~\citep{svgpathtools} and bs4~\citep{bs4} for SVG manipulation. Ablation studies demonstrate that these augmentation techniques significantly improve model performance.

\section{\emojidizzySmall StarVector}
\label{sec:starvector}
\subsection{Architecture}

StarVector is a foundational MLLM for SVG generation, trained for Image-to-SVG and Text-to-SVG Generation. It effectively uses image semantics to identify and utilize shape primitives, producing precise and compact SVG outputs. This capability emerges as the model learns to predict SVG code directly, while it is trained on a large SVG dataset (SVG-Stack). \emph{We frame the task of image vectorization as an inverse rendering and code generation problem}, where images are represented as visual tokens that precede the sequence of SVG code tokens. During generation, an image is converted into visual tokens, prompting StarVector to predict the SVG code following a vectorization trigger token. As depicted in Figure~\ref{fig:starvector}, our architecture employs a large language model (LLM) $f_{\phi}$ and an image encoder $e_{\theta}$, parameterized by trainable parameters $\phi$ and $\theta$, respectively. 

\vspace{2mm}
\noindent\textbf{Visual Tokens.} For each input image $x_v$, the Image Encoder provides flattened grid features $z_v = e_{\theta}(x_v)$. A Vision Transformer (ViT)~\citep{radford2021learning} is utilized to define $e_{\theta}(\cdot)$. All features from the last transformer layer are used, as we require high visual expressivity. LLM Adapter $g_{\varphi}$ is devoted to projecting the visual features $z_v$ into the dimensionality of the LLM, creating visual tokens $h_v$:
\begin{equation}
    h_v = g_{\varphi}(z_v), \\ 
    \text{where } z_v = e_{\theta}(x_v)
\end{equation}
As depicted in Figure~\ref{fig:starvector}-b, the LLM Adapter $g_{\varphi}$ performs a non-linear projection of the image embeddings into the LLM embedding space, producing a set of visual token embeddings (or visual tokens). This transformation aligns the image representations with the LLM, effectively bridging the visual and SVG code modalities. The Adapter is composed of a sequence of fully connected (FC) layers with Swish~\citep{ramachandran2017searching} activation function and Layer Normalization~\citep{lei2016layer}. We initialize the adapter parameters  $\varphi$ using the normal distribution. We initialize the image encoder parameters $\theta$ using the public weights of CLIP ViT-L/14~\citep{radford2021learning}.
\begin{equation}
    g_{\varphi}(z_v) = \text{LayerNorm}(W_L \cdot \text{Swish}(W_h \cdot z_v))
\end{equation}

\vspace{2mm}
\noindent\textbf{Language Modeling.} For each image $x_v$, we define $x_t$ as its corresponding textual caption and $x_s$ as the SVG code. For each sample, we have a tuple $(x_v, x_t, x_s)$. Training sequences are constructed by concatenation: $(x_v, x_s)$ for Image-to-SVG tasks and $(x_t, x_s)$ for Text-to-SVG tasks. To simplify, we use $x_c$ to represent the conditioning sequence, which is either $x_v$ or $x_t$, depending on the task. 

As depicted in Figure~\ref{fig:starvector}, both textual $x_t$ and SVG $x_s$ sequences are processed by a tokenizer and an embedder, which converts text strings to tokens and tokens to embeddings. This embedding operation has trainable parameters. We model the conditional probability of SVG sequences as:
\begin{equation} p(x_s \mid x_c) = \prod_{i=1}^{L} p(x_{s,i} \mid x_{s,<i}, x_c), \end{equation}
where $L$ is the length of the SVG sequence $x_s$. This formulation allows us to use a generative objective with next-token cross-entropy over the SVG sequence. During inference, only the conditioning sequence $x_c$ is given as input, and the SVG code is sampled autoregressively until the ending \texttt{<svg-end>} token is reached. 







\vspace{2mm}
\noindent\textbf{StarVector Variants.} We define two variants of StarVector to explore its scaling behavior, varying in image resolution, LLM parameter count, and context length. 
\vspace{1mm}
\begin{enumerate}
    \item \textbf{StarVector-1B} is initialized with a CLIP ViT-B/32~\cite{radford2021learning} image encoder, processing images at a 224×224 resolution to produce 257 visual tokens. The LLM uses StarCoder-1B~\cite{li2023starcoder} with a context length of 8192 tokens.
    
    \item \textbf{StarVector-8B} employs a SigLip (siglip-so400m-patch14-384)~\cite{zhai2023sigmoid} image encoder, processing images at a 384×384 resolution to yield 576 visual tokens. This model utilizes StarCoder2-7B, offering an expanded context length of 16k tokens. With a total of 8B parameters, we investigate how scaling can yield more precise and compact SVGs due to higher image resolution and enhanced LLM capacity.
\end{enumerate}





\vspace{2mm}
\noindent\textbf{Inference.} During generation, StarVector processes an input image \textit{or} text, converting it into tokens, and then predicts subsequent tokens auto-regressively to produce SVG code. This code is rasterized with CairoSVG~\citep{cairosvg}, generating an image. A key challenge during generation is ensuring both 1) syntactically valid SVGs and 2) SVGs optimized for compactness and precision. Decoding introduces stochasticity, and limited context length can result in incomplete SVGs. We find that StarVector is sensible to temperature, length penalty, and logit bias, i.e., adding more weight to certain tokens, like the \textit{<svg-end>} token which encourages valid SVG outputs (properly closed SVG). We introduce them as inference hyperparameters. To further improve quality, we generate $k$ samples with varied parameters, ranking them based on DinoScore.
\section{\benchmark SVG-Bench: Evaluation Suite for SVG}
\label{svg-bench}
In response to the limited benchmarks for SVG evaluation and to unify evaluation practices~\citep{lopes2019learned, carlier2020deepsvg, reddy2021im2vec}, we introduce SVG-Bench. This benchmark assesses SVG methods across Image-to-SVG, Text-to-SVG, and Diagram Generation tasks, encompassing various SVG types from simple graphics like icons and fonts to complex diagrams with multiple primitives. Dataset statistics are presented in Table~\ref{tab:all-datasets}.

\newcommand{\rot}{60} 
\begin{table*}[t]
    \centering
    \caption{\textbf{Image Vectorization Results.} Image processing methods (denoted by \textsuperscript{\textdagger}) excel in pixel-based metrics (SSIM and MSE) while StarVector models lead in semantic-based metrics (DinoScore and LPIPS). StarVector shows better performance in SVG-Stack, SVG-Fonts, and SVG-Icons but underperforms in SVG-Emoji due to limited training data. We highlight the token lengths of generated SVGs from different models, comparing them to the average token count in test examples (shown in gray below the ``Tokens'' header). Token counts close to the actual number are marked in green, while the largest counts are highlighted in red. \textit{Notably, methods that perform well on MSE tend to utilize a large number of tokens}, whereas StarVector shows remarkable compression.}

    \resizebox{1.0\textwidth}{!}{
    \setlength{\tabcolsep}{3pt} 
\begin{tabular}[t]{@{}lccccc|ccccc|ccccc|ccccc@{}} 
    \toprule
        & \multicolumn{5}{c}{\textbf{SVG-Stack}} & \multicolumn{5}{c}{\textbf{SVG-Fonts}} & \multicolumn{5}{c}{\textbf{SVG-Icons}} & \multicolumn{5}{c}{\textbf{SVG-Emoji}} \\
    
        \cmidrule(lr){2-6} \cmidrule(lr){7-11} \cmidrule(lr){12-16} \cmidrule(lr){17-21}

        \textbf{Method} 
        & \rotatebox{\rot}{\textbf{Dino $\uparrow$}} 
        & \rotatebox{\rot}{\textbf{LPIPS $\downarrow$}} 
        & \rotatebox{\rot}{\textbf{SSIM $\uparrow$}} 
        & \rotatebox{\rot}{\textbf{MSE $\downarrow$}} 
        & \rotatebox{\rot}{\shortstack{\textbf{Tokens} \\ \textcolor{gray}{\tiny{\textbf{2,822}}}}}
        & \rotatebox{\rot}{\textbf{Dino $\uparrow$}} 
        & \rotatebox{\rot}{\textbf{LPIPS $\downarrow$}} 
        & \rotatebox{\rot}{\textbf{SSIM $\uparrow$}} 
        & \rotatebox{\rot}{\textbf{MSE $\downarrow$}} 
        & \rotatebox{\rot}{\shortstack{\textbf{Tokens} \\ \textcolor{gray}{\tiny{\textbf{3,136}}}}}
        & \rotatebox{\rot}{\textbf{Dino $\uparrow$}} 
        & \rotatebox{\rot}{\textbf{LPIPS $\downarrow$}} 
        & \rotatebox{\rot}{\textbf{SSIM $\uparrow$}} 
        & \rotatebox{\rot}{\textbf{MSE $\downarrow$}} 
        & \rotatebox{\rot}{\shortstack{\textbf{Tokens} \\ \textcolor{gray}{\tiny{\textbf{3,305}}}}}
        & \rotatebox{\rot}{\textbf{Dino $\uparrow$}} 
        & \rotatebox{\rot}{\textbf{LPIPS $\downarrow$}} 
        & \rotatebox{\rot}{\textbf{SSIM $\uparrow$}} 
        & \rotatebox{\rot}{\textbf{MSE $\downarrow$}} 
        & \rotatebox{\rot}{\shortstack{\textbf{Tokens} \\ \textcolor{gray}{\tiny{\textbf{5,618}}}}} \\
        
        \midrule
        AutoTrace\textsuperscript{\textdagger} & 
        0.942 & 0.063 & 0.930 & \underline{0.009} &  \cellcolor{lightred} \textbf{59.1k}&0.954 & 0.025 & 0.968 & 0.006 &  \cellcolor{lightred} \textbf{30.8k}&0.946 & 0.053 & 0.937 & 0.014 &  \cellcolor{lightred} \textbf{56.7k}&\underline{0.975} & 0.077 & \underline{0.902} & 0.011 &  \cellcolor{lightred} \textbf{94.0k}\\
        
        Potrace\textsuperscript{\textdagger} & 
        0.898 & 0.139 & 0.856 & 0.036 &  7.5k&
        0.967 & \textbf{0.009} & \textbf{0.988} & \underline{0.002} &  4.2k&
        0.972 & \textbf{0.023} & \underline{0.973} & \textbf{0.004} &  12.0k&0.882 & 0.267 & 0.780 & 0.067 &  9.7k\\
        
        VTracer\textsuperscript{\textdagger} & 
       \underline{0.954} & 0.062 & 0.883 & 0.010 &  9.7k&0.964 & 0.027 & 0.888 & 0.009 &  4.5k&0.940 & 0.062 & 0.914 & 0.017 &  20.0k&\textbf{0.981} & \underline{0.074} & 0.894 & \underline{0.008} &  15.7k\\
        
        Im2Vec  & 
        0.692 & 0.291 & 0.765 & 0.181 &  \cellcolor{lightgreen} \textbf{4.3k}&0.733 & 0.140 & 0.837 & 0.135 &  \cellcolor{lightgreen} \textbf{4.3k}&0.754 & 0.150 & 0.889 & 0.055 &  \cellcolor{lightgreen} \textbf{4.3k}&0.732 & 0.465 & 0.774 & 0.126 &  \cellcolor{lightgreen} \textbf{3.8k}\\
        
        LIVE &
        0.934 & \underline{0.059} & \textbf{0.953} & \textbf{0.003} &  18.3k&
        0.956 & \underline{0.013} & \underline{0.977} & \textbf{0.001} &  18.3k&
        0.959 & 0.035 & \underline{0.973} & \textbf{0.004} &  18.2k&
        0.969 & \textbf{0.060} & \textbf{0.958} & \textbf{0.002} &  18.3k\\
        
        DiffVG & 
        0.810 & 0.156 & 0.856 & 0.019 &  19.7k&0.821 & 0.051 & 0.959 & 0.007 &  19.7k&0.952 & 0.056 & 0.956 & 0.015 &  19.8k&
        0.814 & 0.242 & 0.776 & 0.034 &  19.7k\\
        
        GPT-4-V  & 
        0.852 & 0.317 & 0.711 & 0.195 &  443&0.842 & 0.198 & 0.749 & 0.197 &  279&0.848 & 0.238 & 0.755 & 0.144 &  524&0.850 & 0.344 & 0.712 & 0.170 &  672\\

        \midrule[\heavyrulewidth]
        StarVector-1B & 
        0.926 & 0.149 & 0.840 & 0.078 & \cellcolor{lightgreen} \textbf{3.7k}&\underline{0.978} & 0.022 & 0.961 & 0.022 & \cellcolor{lightgreen} \textbf{2.4k}&
        \underline{0.975} & 0.040 & 0.931 & 0.026 & \cellcolor{lightgreen}\textbf{ 3.5k}&
        0.929 & 0.217 & 0.820 & 0.063 & \cellcolor{lightgreen} \textbf{4.8k}\\
        StarVector-8B & 
        \textbf{0.966} & \textbf{0.058} & \underline{0.947} & 0.026 & \cellcolor{lightgreen}\textbf{ 5.3k}&
        \textbf{0.982} & 0.030 & 0.946 & 0.029 & \cellcolor{lightgreen} \textbf{3.0k}&
        \textbf{0.984} & \underline{0.035} & \textbf{0.975} & \underline{0.012} & \cellcolor{lightgreen} \textbf{2.8k}&
        0.943 & 0.193 & 0.829 & 0.052 & \cellcolor{lightgreen} \textbf{6.7k}\\
\midrule
& \multicolumn{5}{c}{\textbf{SVG-Stack\textsubscript{sim}}} 
& \multicolumn{5}{c}{\textbf{SVG-Fonts\textsubscript{sim}}} 
& \multicolumn{5}{c}{\textbf{SVG-Icons\textsubscript{sim}}} 
& \multicolumn{5}{c}{\textbf{SVG-Emoji\textsubscript{sim}}} \\
\cmidrule(lr){2-6} \cmidrule(lr){7-11} \cmidrule(lr){12-16} \cmidrule(lr){17-21}

        AutoTrace\textsuperscript{\textdagger} & 0.945 & 0.063 & 0.922 & 0.018 &  \cellcolor{lightred} \textbf{74.1k}&0.928 & 0.125 & 0.886 & 0.050 & \cellcolor{lightgreen} 1.5k&0.915 & 0.111 & 0.901 & 0.044 & \cellcolor{lightgreen} 1.3k&0.963 & 0.090 & 0.874 & 0.029 &  \cellcolor{lightred} \textbf{134.8k}\\

        Potrace\textsuperscript{\textdagger} & \underline{0.970} & \textbf{0.022} & \underline{0.968} & \underline{0.006} &  12.2k&\underline{0.991} & \underline{0.012} & \underline{0.983} & \underline{0.003} &  7.7k&\underline{0.983} & \underline{0.025} & \underline{0.976} & \underline{0.004} &  10.4k&\textbf{0.992} & \textbf{0.037} & \textbf{0.951} & \textbf{0.008} &  26.7k \\

        VTracer\textsuperscript{\textdagger} & 0.935 & 0.061 & 0.914 & 0.020 &  16.0k&0.946 & 0.040 & 0.939 & 0.013 &  12.7k&0.945 & 0.043 & 0.946 & 0.012 &  11.9k&0.948 & \underline{0.063} & 0.911 & 0.021 &  16.2k \\
        
        Im2Vec & 0.725 & 0.186 & 0.892 & 0.046 &  \cellcolor{lightgreen} \textbf{4.3k}&0.857 & 0.184 & 0.833 & 0.096 &  284&0.860 & 0.207 & 0.792 & 0.129 &  453&0.695 & 0.179 & 0.898 & 0.045 &  \cellcolor{lightgreen} \textbf{3.7k}\\
        
        LIVE & 0.963 & \underline{0.039} & \textbf{0.974} & \textbf{0.005} &  18.3k&0.975 & 0.016 & \textbf{0.991} & \textbf{0.001} &  \cellcolor{lightred} \textbf{18.3k}&0.961 & 0.030 & \textbf{0.978} & \textbf{0.003} &  \cellcolor{lightred} \textbf{18.2k}&0.958 & 0.075 & \underline{0.934} & \underline{0.014} &  18.2k\\
        
        DeepSVG & 0.907 & 0.192 & 0.835 & 0.071 &  1.5k&0.928 & 0.125 & 0.886 & 0.050 &  1.5k&0.915 & 0.111 & 0.901 & 0.044 &  1.3k&0.822 & 0.209 & 0.841 & 0.074 &  1.8k\\
        
        GPT-4 V & 0.874 & 0.226 & 0.768 & 0.137 &  329&0.946 & 0.040 & 0.939 & 0.013 &  12.7k&0.945 & 0.043 & 0.946 & 0.012 &  11.9k&0.852 & 0.212 & 0.802 & 0.105 &  424\\
        \midrule
        StarVector-1B & 0.954 & 0.089 & 0.870 & 0.053 & \cellcolor{lightgreen} \textbf{2.9k}&- & - & - & - & -& - & - & - & - & -&\underline{0.977} & 0.073 & 0.897 & 0.043 & \cellcolor{lightgreen} \textbf{3.0k}\\
        StarVector-8B & \textbf{0.977} & 0.074 & 0.888 & 0.045 & \cellcolor{lightgreen} \textbf{2.1k}&\textbf{0.993} & \textbf{0.012} & 0.970 & 0.009 & \cellcolor{lightgreen} \textbf{1.3k}&\textbf{0.990} & \textbf{0.024} & 0.947 & 0.017 & \cellcolor{lightgreen} \textbf{2.7k}&0.903 & 0.163 & 0.791 & 0.091 & \cellcolor{lightgreen} \textbf{3.2k}\\
        \bottomrule
    \end{tabular}
    }
    \label{tab:consolidated-image-vectorization-results}
\end{table*}

\subsection{Tasks and Benchmarks}
SVG-Bench focuses on the following tasks. For details on dataset curation and visual examples, see Appendix~\ref{app:datasets}.

\begin{enumerate}
    \vspace{1mm}
    \item{\textbf{Image-to-SVG}}: This task evaluates converting images to SVGs across varying complexities. We introduce \textit{SVG-Fonts}, \textit{SVG-Emoji}, \textit{SVG-Icons}, and \textit{SVG-Stack}, increasing in complexity. While prior works used similar datasets~\citep{lopes2019learned, reddy2021im2vec, carlier2020deepsvg}, access has often been unclear. We provide standardized train, validation, and test splits, along with simplified versions containing only \textit{paths} for compatibility with certain methods.
    \vspace{1mm}    
    \item{\textbf{Text-to-SVG}}: We evaluate the model's ability to generate SVGs from text instructions. This includes the \textit{SVG-Stack} test set, which provides two textual descriptions per image, and the SVG-FIGR dataset, which is sourced from FIGR-8-SVG~\citep{Cloutre2019FIGRFI, wu2023iconshop} dataset, enabling the generation of simpler (path-only) icons from text.
    \vspace{1mm}
    \item{\textbf{Diagram Generation}}: We assess the model's performance in generating diagrams, a specific type of SVG that involves text, rectangles, and arrow primitives. For this, we create the \textit{SVG-Diagrams} test set by extracting samples from SVG-Stack, including textual captions.
\end{enumerate}
\paragraph{Evaluation Metrics.} To compute benchmark scores, we define the following metrics: For image vectorization tasks, we use Mean Squared Error (MSE), Structural Similarity Index (SSIM)~\citep{wang2004image, wang2009mean}, and Learned Perceptual Image Patch Similarity (LPIPS)~\citep{zhang2018unreasonable}. To address the limitations of pixel-based metrics (see Figure~\ref{fig:mse-bad}), we propose DinoScore~\citep{oquab2023dinov2}, which computes L2 distance between DinoV2 features. Token Length (Tokens) measures the size of the SVG samples. We use the StarCoder~\citep{li2023starcoder} tokenizer to tokenize SVG code and compute the average length. These metrics are also used for Diagram Generation. For Text-to-SVG, we build on text-to-image literature~\citep{theis2015note, radford2021learning, ramesh2022hierarchical, rombach2022high} and prior Text-to-SVG methods~\citep{wu2023iconshop, cai2023leveraging}, using FID~\citep{theis2015note}, FID-CLIP~\citep{wu2023iconshop}, and CLIP Score~\citep{radford2021learning} to measure image-text alignment.

\section{Experiments and Results}
\label{sec:experiments}

\begin{table}[t]
    \centering
    \caption{\textbf{Image Vectorization on SVG-Diagrams}. StarVector outperforms LIVE in DinoScore, LPIPS, and SSIM, while LIVE ranks best in MSE. However, visual results (Fig.~\ref{tab:svg-diagrams}) confirm that StarVector is the only effective method for SVG generation, underscoring the misalignment of MSE. Additionally, it remains competitive in terms of token length.}
    \setlength{\tabcolsep}{3pt} 
    \small 
    \begin{tabular}{@{}lccccc} 
        \toprule
        &  \multicolumn{5}{c}{\textbf{SVG-Diagrams}} \\
         \cmidrule(lr){2-6} 
        
        \textbf{Method}  & \textbf{Dino $\uparrow$} & \textbf{LPIPS $\downarrow$} & \textbf{SSIM $\uparrow$} & \textbf{MSE $\downarrow$} & \textbf{Tokens $\downarrow$} \\
        
        \midrule
        
         Autotrace\textsuperscript{\textdagger}  & 0.874 & 0.114 & \underline{0.883} & 0.013 & 90.6k\\
          Potrace\textsuperscript{\textdagger} & 0.875 & 0.153 & 0.862 & 0.026 &  22.6k\\
         VTracer\textsuperscript{\textdagger} & 0.882 & 0.116 & 0.877 & \underline{0.011} & 15.8k\\
         LIVE & 0.870 & 0.121 & 0.859 & \textbf{0.010} & 18.3k\\
         DiffVG & 0.822 & 0.170 & 0.859 & 0.019 & 19.8k\\
         \midrule
         StarVector-1B & \underline{0.943} & \underline{0.107} & 0.862 & 0.032 &  9.5k\\
         StarVector-8B & \textbf{0.959} & \textbf{0.093} & \textbf{0.890} & 0.027 & -\\
        
        \bottomrule
    \end{tabular}

    \vspace{5px}
    
    \label{tab:svg-diagrams}
\end{table}
\begin{table}[t]
    \centering
    \caption{\textbf{Usage of \textit{Paths} and Inference Time.} We ablate the use of the \textit{path} primitive across models. LIVE and DiffVG allow setting the number of paths, while VTracer, Autotrace, and StarVector dynamically determine them. More paths generally improve performance. LIVE achieves the best pixel metrics, but StarVector excels in DinoScore. We also report \textit{average inference time per sample}, noting that LIVE is significantly slower. Results are averaged across SVG-Bench datasets.}
    \resizebox{0.5\textwidth}{!}{
    \setlength{\tabcolsep}{3pt} 
    \resizebox{0.5\textwidth}{!}{
    \begin{tabular}{@{}lcccccc} 
        
        \textbf{Method} & \textbf{\# Paths} & \textbf{Dino $\uparrow$} & \textbf{LPIPS $\downarrow$} & \textbf{SSIM $\uparrow$} & \textbf{MSE $\downarrow$} & \textbf{Time (s) $\downarrow$}\\
        \toprule

        LIVE & 5 & 0.898&0.137 & 0.881 & 0.013 & 190 \\
         & 16 & 0.930&0.064&0.937&0.006 & 290 \\
         & 32 & 0.937&\underline{0.057}&\underline{0.944}&\underline{0.004} & 650 \\
         & 60 & 0.939&\textbf{0.053}&\textbf{0.947}&\textbf{0.003} & 1,412\\
    \midrule
        DiffVG & 15 & 0.781&0.205&0.819&0.066 & 21 \\
         & 60 & 0.844&0.135&0.881&0.018 & 31 \\
         & 120 & 0.895&0.107&0.907&0.013 & 45 \\
        \midrule
        Vtracer & 18 & 0.942&0.067&0.892&0.011 & 0.09 \\
        Potrace & - & 0.937&0.109&0.897&0.024& 10\\
        AutoTrace & 3k & 0.951&0.065&0.924&0.010 & 1 \\
         \midrule
         StarVector-1B & 8 & \underline{0.952}&0.107&0.883&0.044 & 41\\
        StarVector-8B & 10 & \textbf{0.963}&0.085&0.911&0.031 & 74\\
        \bottomrule
    \end{tabular}
    }
    }    
    \label{tab:paths-and-inference}
\end{table}

\begin{table}[t]
    \centering
    \caption{\textbf{Results on Text-to-SVG}: We report FID, FID-CLIP (FID-C), and CLIP Score (CLIP) on SVG-Stack and SVG-FIGR. StarVector models outperform all previous baselines in all metrics. We observe improvement when scaling StarVector from 1B to 8B. Results for DeepSVG+GAN \citep{carlier2020deepsvg, goodfellow2020generative} and Bert on SVG-FIGR are extracted from \cite{wu2023iconshop}, while StarVector is trained on the same data and splits. Missing scores are due to limited model access.}
    \resizebox{0.5\textwidth}{!}{
    \setlength{\tabcolsep}{3pt} 
    \begin{tabular}{@{}lccc|ccc@{}} 
        \toprule
        & \multicolumn{3}{c}{\textbf{SVG-FIGR}} 
        & \multicolumn{3}{c}{\textbf{SVG-Stack}} \\
         \cmidrule(lr){2-4} \cmidrule(lr){5-7}
        \textbf{Method}  & \textbf{FID $\downarrow$} & \textbf{FID-C $\downarrow$} & \textbf{CLIP $\uparrow$}
         & \textbf{FID $\downarrow$} & \textbf{FID-C $\downarrow$} & \textbf{CLIP $\uparrow$} \\
        \midrule
        DeepSVG+GAN  & - &12.011 & 21.783 & - & - & - \\
        Bert & - & 35.104 & 22.035 & - & - & - \\
        IconShop & - & 4.657 & 25.746 & - & - & - \\
        GPT-4  & 32.953 & 19.026 & 26.088 & 37.381 & 9.664 & 26.228 \\
        CodeLlama  & 29.002 & 22.536 & 26.227 & 34.777 & 11.152 & 25.532 \\
        \midrule
        StarVector-1B  & \underline{15.263} & \underline{3.834} & \underline{26.342} & \underline{28.374} & \underline{6.482} & \underline{29.372}\\
        StarVector-8B  & \textbf{10.067} & \textbf{1.308} & \textbf{27.366} & \textbf{25.828} & \textbf{4.645} & \textbf{31.307}\\
        \bottomrule
    \end{tabular}
    }
    
    \label{tab:merged_results_text2svg}
\end{table}

We train StarVector (1B and 8B versions) on the \textit{inverse rendering} vectorization task using SVG-Stack dataset. We then fine-tune on the other datasets mentioned in Section~\ref{svg-bench}, as well as for Text-to-SVG task. We evaluate StarVector and other methods on SVG-Bench, focusing on quantitative and qualitative performance, SVG primitive use, and compactness. The following sections present the experimental setup and results. Ablation studies on the architecture, data augmentation, and generation can be found in Appendix~\ref{app:ablations}.

\vspace{2mm}
\noindent\textbf{Baselines.} \noindent\textbf{Baselines.} For our comparisons, we consider the following model baselines: For Image-to-SVG, we evaluate top image processing algorithms such as Potrace~\citep{potrace}, Vtracer~\citep{vtracer}, and Autotrace~\cite{autotrace}. We also report on deep learning methods including DeepSVG (<5M parameters)~\citep{carlier2020deepsvg}, Im2Vec (<5M)~\citep{reddy2021im2vec}, and MLLMs like GPT-4 Vision (>100B)~\citep{openai2023gpt4v}. For Text-to-SVG, we consider methods like IconShop (>1B), DeepSVG+GAN (<5M), and BERT (>1B), as outlined in~\citep{wu2023iconshop}. Additionally, we evaluate LLMs such as CodeLlama-70b~\citep{touvron2023llama} and GPT-4 (>100B)~\citep{openai2023gpt4}. For more details, see Appendix~\ref{app:baselines}, and Table~\ref{tab:related-work_comparison} summarizes the SVG capabilities of all methods.

\vspace{2mm}
\noindent\textbf{Training and Inference.} We train StarVector-1B with a batch size of 128 and StarVector-8B with 512, using a learning rate of 1e-5 and the AdamW optimizer. StarVector-1B took 7 days on 8 A100 GPUs, while StarVector-8B took 10 days on 64 H100 GPUs, both completing 2 epochs. Full training details are in Appendix~\ref{app:training}. During inference, we generate $k=5$ SVG outputs with temperatures ranging from 0 to 1, selecting the one with the highest DinoScore. We set a logit bias of 10 for the \textit{<svg-end>} token and apply top-p nucleus sampling (0.9) with a length penalty of -0.5, using beam search with a size of 1. We utilize vLLM~\citep{kwon2023efficient} as the backend to accelerate the generation process.

\vspace{2mm}
\noindent\textbf{Experimental Setup.} StarVector models are initialized as described in Section~\ref{sec:starvector} and all their weights are unfrozen for the Image-to-SVG task using SVG-Stack. We then fine-tune these models on the train sets of SVG-Emoji, SVG-Fonts, and SVG-Icons, including their simplified versions. For the Text-to-SVG task, the image encoder is disregarded. Only the LLM is trained on SVG-Stack and FIRG-SVG. We perform a comprehensive evaluation on SVG-Bench, analyzing the performance of baseline models mentioned above, alongside the StarVector models. We reproduce all baselines by training them on the benchmark’s training sets, within the limits of their availability.

\subsection{Main Results}

\begin{figure}[t]
    \centering
    \includegraphics[width=1.0\linewidth]{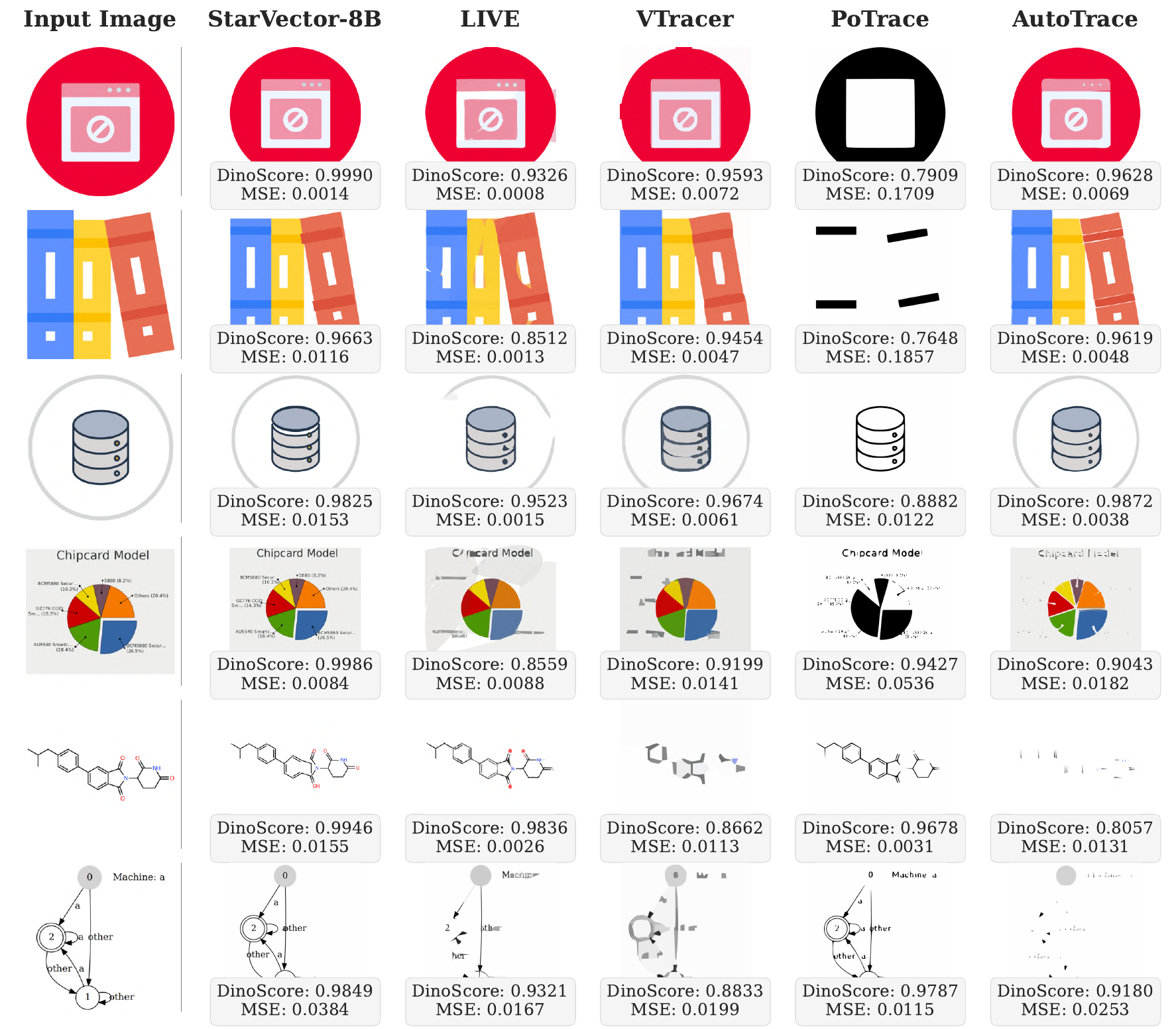}
    \caption{\textbf{Image vectorization results} on SVG-Stack (first 3 rows) and SVG-Diagrams (last 3 rows) test sets.}
    \label{fig:svg-stack-comparison}
\end{figure}

\vspace{2mm}
\noindent\textbf{Image Vectorization.} Table~\ref{tab:consolidated-image-vectorization-results} presents vectorization scores of models across 8 benchmarks. StarVector outperforms all other models in terms of DinoScore, achieving the highest score on six out of the eight benchmarks, thereby establishing its dominance in this metric. Results for the other metrics—LPIPS, SSIM, and MSE—are more varied. However, LIVE demonstrates superior performance on the SSIM and MSE metrics across all datasets. It is important to note that models that perform well on MSE tend to generate larger SVG files, as indicated by the number of tokens in the SVG code. This creates overly complex vectors with visible artifacts (see Fig.~\ref{fig:svg-stack-comparison}). Specifically, LIVE’s SVG outputs average around 18k tokens, while VTracer varies between 4.5k and 20k tokens. In contrast, StarVector averages approximately 3k tokens, closely matching the ground truth token count. This efficiency is primarily attributed to StarVector’s effective use of SVG primitives, as illustrated in Figure~\ref{fig:mse-bad} (right).

Figure~\ref{fig:svg-stack-comparison} presents qualitative results of models, on SVG-Stack and SVG-Diagrams. In terms of visual quality, LIVE, VTracer, and AutoTrace produce artifacts, especially when dealing with small details. Potrace offers more sharp results, but it is monochromatic. StarVector-8B produces superior results on shape preservation and definition. More results and qualitative samples can be found in Appendix~\ref{app:additional-results}.

\begin{figure}[t]
    \centering
    \includegraphics[width=1.0\linewidth]{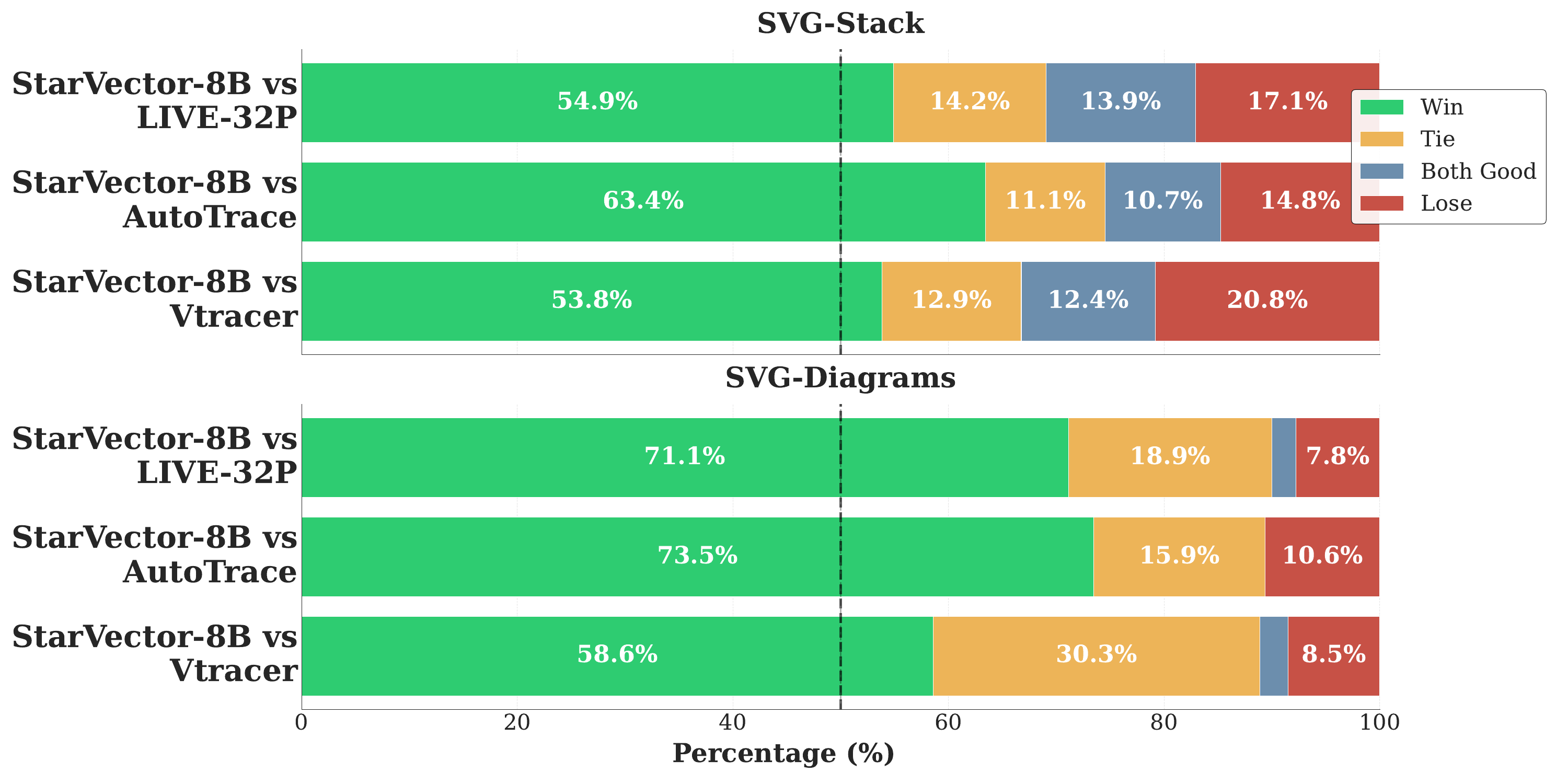}
    \caption{\textbf{Human evaluation}. StarVector-8B was evaluated against top-performing baselines: LIVE, AutoTrace, and Vtracer. Results consistently showed a strong human preference for SVG outputs generated by StarVector-8B over the baselines. In total, 1,948 evaluations were collected from 30 unique participants.}
    \label{fig:human-eval}
\end{figure}

\vspace{2mm}
\noindent\textbf{Why is pixel-based MSE not well-suited?}
Our results reveal significant limitations of pixel-based metrics (MSE, SSIM, LPIPS) for SVG quality assessment. While StarVector shows worse MSE scores compared to other methods in Table~\ref{tab:svg-diagrams}, visual inspection of the results in Figures~\ref{fig:svg-stack-comparison} and~\ref{fig:svg-diagrams-results} (especially on diagrams) demonstrates StarVector's superior quality. This discrepancy is particularly evident in the `planet' example (Figure~\ref{fig:mse-bad}), where StarVector preserves color gradients and line definition, yet receives poor MSE scores due to minor pixel misalignments. Human evaluation (Fig.~\ref{fig:human-eval}) shows a preference for StarVector's outputs over other models, contradicting these metrics.

Pixel-based metrics have limitations due to (a) the prevalence of constant background colors (allowing even empty SVGs to score reasonably), (b) sensitivity to small spatial misalignments, and (c) inability to capture non-smooth artifacts at corners. DinoScore proves more reliable and aligns consistently with visual quality, scoring StarVector higher on well-formed samples while penalizing poorly formatted ones, thanks to robust self-supervised training~\citep{oquab2023dinov2}.

\vspace{2mm}
\noindent\textbf{Diagram Generation.} Table~\ref{tab:svg-diagrams} shows results on SVG-Diagrams, and Figures~\ref{fig:svg-stack-comparison} and~\ref{fig:svg-diagrams-results} provide visual examples. Results highlight that \textbf{StarVector is the only method capable of performing diagram generation}, as it uniquely applies the required primitives like rectangles, arrows, and text, whereas other methods produce blobs and curves that attempt to replicate structure and color. Metrics such as DinoScore, LPIPS, capture this advantage, while MSE and SSIM remain poorly aligned. Human evaluations further confirmed the preference for StarVector's outputs.

\vspace{2mm}
\noindent\textbf{SVG Primitive Usage.} StarVector produces more compact SVGs by optimally using SVG primitives. This innovation combines visual semantic understanding and shape composition with direct SVG code generation, enabling decomposition into basic primitives. As shown in Figure~\ref{fig:mse-bad} (right), StarVector efficiently represents shapes using primitives, while VTracer relies on a large collection of \textit{paths}. See Appendix~\ref{app:app-primitives} for more examples (Figure~\ref{fig:primitives}) and SVG tag distribution analysis (Figure~\ref{fig:tag-stats}).

\vspace{2mm}
\noindent\textbf{Human Evaluation.} We conducted human evaluations comparing StarVector-8B with baseline results, involving participants from diverse backgrounds screened for conflicts of interest. Results in Figure~\ref{fig:human-eval} show a strong preference for StarVector-8B in all settings, particularly in the SVG-Diagrams tasks, highlighting a disconnect between pixel-based metrics (MSE, SSIM) and visual perception of SVG. While baselines prioritize pixel-perfect reconstruction, humans prefer StarVector's sharp, well-defined shapes and effective use of primitives (Figure~\ref{fig:tag-stats}). Spearman correlations between model metrics (MSE and DINO) and human evaluation show weak correlations for MSE (0.0596 and -0.1002), indicating it is not a strong predictor. In contrast, DinoScore exhibits stronger correlations, with values of -0.6193 and 0.6214, and a strong correlation of 0.7577 between DINO differences and human evaluation differences, highlighting DINO as a more reliable metric.

\vspace{2mm}
\noindent\textbf{Text-to-SVG.} Table~\ref{tab:merged_results_text2svg} shows StarVector outperforms baselines on SVG-FIGR and SVG-Stack. While qualitative results show potential, semantic accuracy often suffers due to data quality (see Appendix~\ref{app:text2svg}).

\section{Conclusions}
\label{sec:conclusions}

We introduced StarVector, an MLLM that excels in Image-to-SVG and Text-to-SVG generation, delivering superior visual quality, precise line and shape rendering, and optimal SVG primitive usage compared to baselines. To train StarVector, we created SVG-Stack, a diverse dataset that enables generalization across SVG types and primitives. Additionally, we developed SVG-Bench, a unified benchmark with tasks, datasets, and targeted metrics. Our study shows that traditional metrics inadequately capture SVG quality, leading us to propose more effective alternatives.

\textbf{Limitations and Future Work.} StarVector is constrained by its 16k token context, which is inadequate for complex SVGs. It relies primarily on code prediction with minimal visual feedback. Additionally, generation speed is limited by the LLM. Future work may investigate integrating pixel-based signals to overcome these challenges.

\textbf{Acknowledgements.} We sincerely thank Ghazwa Darwiche, Christian Hudon, Fanny Rancourt, and Marie-Ève Marchand for their invaluable administrative and technical support. This work was supported by the Natural Sciences and Engineering Research Council of Canada and Mitacs. Chris Pal acknowledges the Canada CIFAR AI Chair. We also appreciate the human verifiers for their contributions to assessment and data verification.

{
    \small
    \bibliographystyle{ieeenat_fullname}
    \bibliography{main}

\begin{thebibliography}{100}
\providecommand{\natexlab}[1]{#1}
\providecommand{\url}[1]{\texttt{#1}}
\expandafter\ifx\csname urlstyle\endcsname\relax
  \providecommand{\doi}[1]{doi: #1}\else
  \providecommand{\doi}{doi: \begingroup \urlstyle{rm}\Url}\fi

\bibitem[The()]{TheC3}
The claude 3 model family: Opus, sonnet, haiku.

\bibitem[Alayrac et~al.(2022)Alayrac, Donahue, Luc, Miech, Barr, Hasson, Lenc, Mensch, Millican, Reynolds, et~al.]{alayrac2022flamingo}
Jean-Baptiste Alayrac, Jeff Donahue, Pauline Luc, Antoine Miech, Iain Barr, Yana Hasson, Karel Lenc, Arthur Mensch, Katherine Millican, Malcolm Reynolds, et~al.
\newblock Flamingo: a visual language model for few-shot learning.
\newblock \emph{Advances in Neural Information Processing Systems}, 35:\penalty0 23716--23736, 2022.

\bibitem[Allal et~al.(2023)Allal, Li, Kocetkov, Mou, Akiki, Ferrandis, Muennighoff, Mishra, Gu, Dey, et~al.]{allal2023santacoder}
Loubna~Ben Allal, Raymond Li, Denis Kocetkov, Chenghao Mou, Christopher Akiki, Carlos~Munoz Ferrandis, Niklas Muennighoff, Mayank Mishra, Alex Gu, Manan Dey, et~al.
\newblock Santacoder: don't reach for the stars!
\newblock \emph{arXiv preprint arXiv:2301.03988}, 2023.

\bibitem[Antol et~al.(2015)Antol, Agrawal, Lu, Mitchell, Batra, Zitnick, and Parikh]{antol2015vqa}
Stanislaw Antol, Aishwarya Agrawal, Jiasen Lu, Margaret Mitchell, Dhruv Batra, C~Lawrence Zitnick, and Devi Parikh.
\newblock Vqa: Visual question answering.
\newblock In \emph{Proceedings of the IEEE international conference on computer vision}, pages 2425--2433, 2015.

\bibitem[Ba et~al.(2016)Ba, Kiros, and Hinton]{ba2016layer}
Jimmy~Lei Ba, Jamie~Ryan Kiros, and Geoffrey~E Hinton.
\newblock Layer normalization.
\newblock \emph{arXiv preprint arXiv:1607.06450}, 2016.

\bibitem[Balaji et~al.(2022)Balaji, Nah, Huang, Vahdat, Song, Zhang, Kreis, Aittala, Aila, Laine, Catanzaro, Karras, and Liu]{balaji2022eDiff-I}
Yogesh Balaji, Seungjun Nah, Xun Huang, Arash Vahdat, Jiaming Song, Qinsheng Zhang, Karsten Kreis, Miika Aittala, Timo Aila, Samuli Laine, Bryan Catanzaro, Tero Karras, and Ming-Yu Liu.
\newblock ediff-i: Text-to-image diffusion models with ensemble of expert denoisers.
\newblock \emph{arXiv preprint arXiv:2211.01324}, 2022.

\bibitem[Belouadi et~al.(2023)Belouadi, Lauscher, and Eger]{belouadi2023automatikz}
Jonas Belouadi, Anne Lauscher, and Steffen Eger.
\newblock Automatikz: Text-guided synthesis of scientific vector graphics with tikz.
\newblock \emph{arXiv preprint arXiv:2310.00367}, 2023.

\bibitem[Belouadi et~al.(2024)Belouadi, Ponzetto, and Eger]{belouadi2024detikzify}
Jonas Belouadi, Simone~Paolo Ponzetto, and Steffen Eger.
\newblock Detikzify: Synthesizing graphics programs for scientific figures and sketches with tikz.
\newblock \emph{arXiv preprint arXiv:2405.15306}, 2024.

\bibitem[Bisong and Bisong(2019)]{bisong2019google}
Ekaba Bisong and Ekaba Bisong.
\newblock Google bigquery.
\newblock \emph{Building Machine Learning and Deep Learning Models on Google Cloud Platform: A Comprehensive Guide for Beginners}, pages 485--517, 2019.

\bibitem[Brown et~al.(2020)Brown, Mann, Ryder, Subbiah, Kaplan, Dhariwal, Neelakantan, Shyam, Sastry, Askell, et~al.]{brown2020language}
Tom Brown, Benjamin Mann, Nick Ryder, Melanie Subbiah, Jared~D Kaplan, Prafulla Dhariwal, Arvind Neelakantan, Pranav Shyam, Girish Sastry, Amanda Askell, et~al.
\newblock Language models are few-shot learners.
\newblock \emph{Advances in neural information processing systems}, 33:\penalty0 1877--1901, 2020.

\bibitem[Bubeck et~al.(2023)Bubeck, Chandrasekaran, Eldan, Gehrke, Horvitz, Kamar, Lee, Lee, Li, Lundberg, et~al.]{bubeck2023sparks}
S{\'e}bastien Bubeck, Varun Chandrasekaran, Ronen Eldan, Johannes Gehrke, Eric Horvitz, Ece Kamar, Peter Lee, Yin~Tat Lee, Yuanzhi Li, Scott Lundberg, et~al.
\newblock Sparks of artificial general intelligence: Early experiments with gpt-4.
\newblock \emph{arXiv preprint arXiv:2303.12712}, 2023.

\bibitem[Cai et~al.(2023)Cai, Huang, Li, Wang, and Lee]{cai2023leveraging}
Mu Cai, Zeyi Huang, Yuheng Li, Haohan Wang, and Yong~Jae Lee.
\newblock Leveraging large language models for scalable vector graphics-driven image understanding.
\newblock \emph{arXiv preprint arXiv:2306.06094}, 2023.

\bibitem[Cao et~al.(2023)Cao, Wang, Echevarria, and Liu]{cao2023svgformer}
Defu Cao, Zhaowen Wang, Jose Echevarria, and Yan Liu.
\newblock Svgformer: Representation learning for continuous vector graphics using transformers.
\newblock In \emph{Proceedings of the IEEE/CVF Conference on Computer Vision and Pattern Recognition}, pages 10093--10102, 2023.

\bibitem[Carlier et~al.(2020)Carlier, Danelljan, Alahi, and Timofte]{carlier2020deepsvg}
Alexandre Carlier, Martin Danelljan, Alexandre Alahi, and Radu Timofte.
\newblock Deepsvg: A hierarchical generative network for vector graphics animation.
\newblock \emph{Advances in Neural Information Processing Systems}, 33:\penalty0 16351--16361, 2020.

\bibitem[Chen et~al.(2021)Chen, Tworek, Jun, Yuan, Pinto, Kaplan, Edwards, Burda, Joseph, Brockman, et~al.]{chen2021evaluating}
Mark Chen, Jerry Tworek, Heewoo Jun, Qiming Yuan, Henrique Ponde de~Oliveira Pinto, Jared Kaplan, Harri Edwards, Yuri Burda, Nicholas Joseph, Greg Brockman, et~al.
\newblock Evaluating large language models trained on code.
\newblock \emph{arXiv preprint arXiv:2107.03374}, 2021.

\bibitem[Clou{\^a}tre and Demers(2019)]{Cloutre2019FIGRFI}
Louis Clou{\^a}tre and Marc Demers.
\newblock Figr: Few-shot image generation with reptile.
\newblock \emph{ArXiv}, abs/1901.02199, 2019.

\bibitem[Dakhel et~al.(2023)Dakhel, Majdinasab, Nikanjam, Khomh, Desmarais, and Jiang]{dakhel2023github}
Arghavan~Moradi Dakhel, Vahid Majdinasab, Amin Nikanjam, Foutse Khomh, Michel~C Desmarais, and Zhen Ming~Jack Jiang.
\newblock Github copilot ai pair programmer: Asset or liability?
\newblock \emph{Journal of Systems and Software}, 203:\penalty0 111734, 2023.

\bibitem[Dao et~al.(2022)Dao, Fu, Ermon, Rudra, and R{\'e}]{dao2022flashattention}
Tri Dao, Dan Fu, Stefano Ermon, Atri Rudra, and Christopher R{\'e}.
\newblock Flashattention: Fast and memory-efficient exact attention with io-awareness.
\newblock \emph{Advances in Neural Information Processing Systems}, 35:\penalty0 16344--16359, 2022.

\bibitem[Deng et~al.(2009)Deng, Dong, Socher, Li, Li, and Fei-Fei]{deng2009imagenet}
Jia Deng, Wei Dong, Richard Socher, Li-Jia Li, Kai Li, and Li Fei-Fei.
\newblock Imagenet: A large-scale hierarchical image database.
\newblock In \emph{2009 IEEE conference on computer vision and pattern recognition}, pages 248--255. Ieee, 2009.

\bibitem[Devlin et~al.(2018)Devlin, Chang, Lee, and Toutanova]{devlin2018bert}
Jacob Devlin, Ming-Wei Chang, Kenton Lee, and Kristina Toutanova.
\newblock Bert: Pre-training of deep bidirectional transformers for language understanding.
\newblock \emph{arXiv preprint arXiv:1810.04805}, 2018.

\bibitem[Diebel(2008)]{diebel2008bayesian}
James~Richard Diebel.
\newblock \emph{Bayesian Image Vectorization: the probabilistic inversion of vector image rasterization}.
\newblock Stanford University, 2008.

\bibitem[Dosovitskiy et~al.(2020)Dosovitskiy, Beyer, Kolesnikov, Weissenborn, Zhai, Unterthiner, Dehghani, Minderer, Heigold, Gelly, et~al.]{dosovitskiy2020image}
Alexey Dosovitskiy, Lucas Beyer, Alexander Kolesnikov, Dirk Weissenborn, Xiaohua Zhai, Thomas Unterthiner, Mostafa Dehghani, Matthias Minderer, Georg Heigold, Sylvain Gelly, et~al.
\newblock An image is worth 16x16 words: Transformers for image recognition at scale.
\newblock \emph{arXiv preprint arXiv:2010.11929}, 2020.

\bibitem[Esser et~al.(2021)Esser, Rombach, and Ommer]{esser2021taming}
Patrick Esser, Robin Rombach, and Bjorn Ommer.
\newblock Taming transformers for high-resolution image synthesis.
\newblock In \emph{Proceedings of the IEEE/CVF conference on computer vision and pattern recognition}, pages 12873--12883, 2021.

\bibitem[Esser et~al.(2024)Esser, Kulal, Blattmann, Entezari, M{\"u}ller, Saini, Levi, Lorenz, Sauer, Boesel, et~al.]{esser2024scaling}
Patrick Esser, Sumith Kulal, Andreas Blattmann, Rahim Entezari, Jonas M{\"u}ller, Harry Saini, Yam Levi, Dominik Lorenz, Axel Sauer, Frederic Boesel, et~al.
\newblock Scaling rectified flow transformers for high-resolution image synthesis.
\newblock In \emph{Forty-first International Conference on Machine Learning}, 2024.

\bibitem[Ferraiolo et~al.(2000)Ferraiolo, Jun, and Jackson]{ferraiolo2000scalable}
Jon Ferraiolo, Fujisawa Jun, and Dean Jackson.
\newblock \emph{Scalable vector graphics (SVG) 1.0 specification}.
\newblock iuniverse Bloomington, 2000.

\bibitem[Gao et~al.(2020)Gao, Biderman, Black, Golding, Hoppe, Foster, Phang, He, Thite, Nabeshima, et~al.]{gao2020pile}
Leo Gao, Stella Biderman, Sid Black, Laurence Golding, Travis Hoppe, Charles Foster, Jason Phang, Horace He, Anish Thite, Noa Nabeshima, et~al.
\newblock The pile: An 800gb dataset of diverse text for language modeling.
\newblock \emph{arXiv preprint arXiv:2101.00027}, 2020.

\bibitem[Glorot and Bengio(2010)]{glorot2010understanding}
Xavier Glorot and Yoshua Bengio.
\newblock Understanding the difficulty of training deep feedforward neural networks.
\newblock In \emph{Proceedings of the thirteenth international conference on artificial intelligence and statistics}, pages 249--256. JMLR Workshop and Conference Proceedings, 2010.

\bibitem[Goodfellow et~al.(2020)Goodfellow, Pouget-Abadie, Mirza, Xu, Warde-Farley, Ozair, Courville, and Bengio]{goodfellow2020generative}
Ian Goodfellow, Jean Pouget-Abadie, Mehdi Mirza, Bing Xu, David Warde-Farley, Sherjil Ozair, Aaron Courville, and Yoshua Bengio.
\newblock Generative adversarial networks.
\newblock \emph{Communications of the ACM}, 63\penalty0 (11):\penalty0 139--144, 2020.

\bibitem[Ho et~al.(2020)Ho, Jain, and Abbeel]{ho2020denoising}
Jonathan Ho, Ajay Jain, and Pieter Abbeel.
\newblock Denoising diffusion probabilistic models.
\newblock \emph{Advances in neural information processing systems}, 33:\penalty0 6840--6851, 2020.

\bibitem[Hochreiter and Schmidhuber(1997)]{hochreiter1997long}
Sepp Hochreiter and J{\"u}rgen Schmidhuber.
\newblock Long short-term memory.
\newblock \emph{Neural computation}, 9\penalty0 (8):\penalty0 1735--1780, 1997.

\bibitem[Holtzman et~al.(2019)Holtzman, Buys, Du, Forbes, and Choi]{holtzman2019curious}
Ari Holtzman, Jan Buys, Li Du, Maxwell Forbes, and Yejin Choi.
\newblock The curious case of neural text degeneration.
\newblock \emph{arXiv preprint arXiv:1904.09751}, 2019.

\bibitem[Husain et~al.(2019)Husain, Wu, Gazit, Allamanis, and Brockschmidt]{husain2019codesearchnet}
Hamel Husain, Ho-Hsiang Wu, Tiferet Gazit, Miltiadis Allamanis, and Marc Brockschmidt.
\newblock Codesearchnet challenge: Evaluating the state of semantic code search.
\newblock \emph{arXiv preprint arXiv:1909.09436}, 2019.

\bibitem[Jain et~al.(2023)Jain, Xie, and Abbeel]{jain2023vectorfusion}
Ajay Jain, Amber Xie, and Pieter Abbeel.
\newblock Vectorfusion: Text-to-svg by abstracting pixel-based diffusion models.
\newblock In \emph{Proceedings of the IEEE/CVF Conference on Computer Vision and Pattern Recognition}, pages 1911--1920, 2023.

\bibitem[Kingma and Welling(2013)]{kingma2013auto}
Diederik~P Kingma and Max Welling.
\newblock Auto-encoding variational bayes.
\newblock \emph{arXiv preprint arXiv:1312.6114}, 2013.

\bibitem[Kocetkov et~al.(2022)Kocetkov, Li, Allal, Li, Mou, Ferrandis, Jernite, Mitchell, Hughes, Wolf, et~al.]{kocetkov2022stack}
Denis Kocetkov, Raymond Li, Loubna~Ben Allal, Jia Li, Chenghao Mou, Carlos~Mu{\~n}oz Ferrandis, Yacine Jernite, Margaret Mitchell, Sean Hughes, Thomas Wolf, et~al.
\newblock The stack: 3 tb of permissively licensed source code.
\newblock \emph{arXiv preprint arXiv:2211.15533}, 2022.

\bibitem[Kozea(2023)]{cairosvg}
Kozea.
\newblock Cairosvg.
\newblock \url{https://cairosvg.org/}, 2023.

\bibitem[Kwon et~al.(2023)Kwon, Li, Zhuang, Sheng, Zheng, Yu, Gonzalez, Zhang, and Stoica]{kwon2023efficient}
Woosuk Kwon, Zhuohan Li, Siyuan Zhuang, Ying Sheng, Lianmin Zheng, Cody~Hao Yu, Joseph Gonzalez, Hao Zhang, and Ion Stoica.
\newblock Efficient memory management for large language model serving with pagedattention.
\newblock In \emph{Proceedings of the 29th Symposium on Operating Systems Principles}, pages 611--626, 2023.

\bibitem[Lei~Ba et~al.(2016)Lei~Ba, Kiros, and Hinton]{lei2016layer}
Jimmy Lei~Ba, Jamie~Ryan Kiros, and Geoffrey~E Hinton.
\newblock Layer normalization.
\newblock \emph{ArXiv e-prints}, pages arXiv--1607, 2016.

\bibitem[Li et~al.(2022{\natexlab{a}})Li, Xu, Tian, Wang, Yan, Bi, Ye, Chen, Xu, Cao, et~al.]{li2022mplug}
Chenliang Li, Haiyang Xu, Junfeng Tian, Wei Wang, Ming Yan, Bin Bi, Jiabo Ye, Hehong Chen, Guohai Xu, Zheng Cao, et~al.
\newblock mplug: Effective and efficient vision-language learning by cross-modal skip-connections.
\newblock \emph{arXiv preprint arXiv:2205.12005}, 2022{\natexlab{a}}.

\bibitem[Li et~al.(2022{\natexlab{b}})Li, Li, Xiong, and Hoi]{li2022blip}
Junnan Li, Dongxu Li, Caiming Xiong, and Steven Hoi.
\newblock Blip: Bootstrapping language-image pre-training for unified vision-language understanding and generation.
\newblock In \emph{International Conference on Machine Learning}, pages 12888--12900. PMLR, 2022{\natexlab{b}}.

\bibitem[Li et~al.(2023{\natexlab{a}})Li, Li, Savarese, and Hoi]{li2023blip}
Junnan Li, Dongxu Li, Silvio Savarese, and Steven Hoi.
\newblock Blip-2: Bootstrapping language-image pre-training with frozen image encoders and large language models.
\newblock \emph{arXiv preprint arXiv:2301.12597}, 2023{\natexlab{a}}.

\bibitem[Li et~al.(2023{\natexlab{b}})Li, Allal, Zi, Muennighoff, Kocetkov, Mou, Marone, Akiki, Li, Chim, et~al.]{li2023starcoder}
Raymond Li, Loubna~Ben Allal, Yangtian Zi, Niklas Muennighoff, Denis Kocetkov, Chenghao Mou, Marc Marone, Christopher Akiki, Jia Li, Jenny Chim, et~al.
\newblock Starcoder: may the source be with you!
\newblock \emph{arXiv preprint arXiv:2305.06161}, 2023{\natexlab{b}}.

\bibitem[Li et~al.(2020)Li, Luk{\'a}{\v{c}}, Gharbi, and Ragan-Kelley]{li2020differentiable}
Tzu-Mao Li, Michal Luk{\'a}{\v{c}}, Micha{\"e}l Gharbi, and Jonathan Ragan-Kelley.
\newblock Differentiable vector graphics rasterization for editing and learning.
\newblock \emph{ACM Transactions on Graphics (TOG)}, 39\penalty0 (6):\penalty0 1--15, 2020.

\bibitem[Liao et~al.(2012)Liao, Hoppe, Forsyth, and Yu]{liao2012subdivision}
Zicheng Liao, Hugues Hoppe, David Forsyth, and Yizhou Yu.
\newblock A subdivision-based representation for vector image editing.
\newblock \emph{IEEE transactions on visualization and computer graphics}, 18\penalty0 (11):\penalty0 1858--1867, 2012.

\bibitem[Liu et~al.(2023)Liu, Li, Wu, and Lee]{liu2023visual}
Haotian Liu, Chunyuan Li, Qingyang Wu, and Yong~Jae Lee.
\newblock Visual instruction tuning.
\newblock \emph{arXiv preprint arXiv:2304.08485}, 2023.

\bibitem[Liu et~al.(2022)Liu, Mao, Wu, Feichtenhofer, Darrell, and Xie]{liu2022convnet}
Zhuang Liu, Hanzi Mao, Chao-Yuan Wu, Christoph Feichtenhofer, Trevor Darrell, and Saining Xie.
\newblock A convnet for the 2020s.
\newblock In \emph{Proceedings of the IEEE/CVF conference on computer vision and pattern recognition}, pages 11976--11986, 2022.

\bibitem[Lopes et~al.(2019)Lopes, Ha, Eck, and Shlens]{lopes2019learned}
Raphael~Gontijo Lopes, David Ha, Douglas Eck, and Jonathon Shlens.
\newblock A learned representation for scalable vector graphics.
\newblock In \emph{Proceedings of the IEEE/CVF International Conference on Computer Vision}, pages 7930--7939, 2019.

\bibitem[Loshchilov and Hutter(2017)]{loshchilov2017decoupled}
Ilya Loshchilov and Frank Hutter.
\newblock Decoupled weight decay regularization.
\newblock \emph{arXiv preprint arXiv:1711.05101}, 2017.

\bibitem[Lozhkov et~al.(2024)Lozhkov, Li, Allal, Cassano, Lamy-Poirier, Tazi, Tang, Pykhtar, Liu, Wei, et~al.]{lozhkov2024starcoder}
Anton Lozhkov, Raymond Li, Loubna~Ben Allal, Federico Cassano, Joel Lamy-Poirier, Nouamane Tazi, Ao Tang, Dmytro Pykhtar, Jiawei Liu, Yuxiang Wei, et~al.
\newblock Starcoder 2 and the stack v2: The next generation.
\newblock \emph{arXiv preprint arXiv:2402.19173}, 2024.

\bibitem[Ma et~al.(2022)Ma, Zhou, Xu, Sun, Filev, Orlov, Fu, and Shi]{ma2022towards}
Xu Ma, Yuqian Zhou, Xingqian Xu, Bin Sun, Valerii Filev, Nikita Orlov, Yun Fu, and Humphrey Shi.
\newblock Towards layer-wise image vectorization.
\newblock In \emph{Proceedings of the IEEE/CVF Conference on Computer Vision and Pattern Recognition}, pages 16314--16323, 2022.

\bibitem[{Martin Weber}(2024)]{autotrace}
{Martin Weber}.
\newblock {Autotrace}.
\newblock \url{https://github.com/autotrace/autotrace}, 2024.

\bibitem[Murray and Chiang(2018)]{murray2018correcting}
Kenton Murray and David Chiang.
\newblock Correcting length bias in neural machine translation.
\newblock \emph{arXiv preprint arXiv:1808.10006}, 2018.

\bibitem[Nijkamp et~al.(2022)Nijkamp, Pang, Hayashi, Tu, Wang, Zhou, Savarese, and Xiong]{nijkamp2022codegen}
Erik Nijkamp, Bo Pang, Hiroaki Hayashi, Lifu Tu, Huan Wang, Yingbo Zhou, Silvio Savarese, and Caiming Xiong.
\newblock Codegen: An open large language model for code with multi-turn program synthesis.
\newblock \emph{arXiv preprint arXiv:2203.13474}, 2022.

\bibitem[Nishina and Matsui(2024)]{nishina2024svgeditbench}
Kunato Nishina and Yusuke Matsui.
\newblock Svgeditbench: A benchmark dataset for quantitative assessment of llm's svg editing capabilities.
\newblock \emph{arXiv preprint arXiv:2404.13710}, 2024.

\bibitem[OpenAI(2023)]{openai2023gpt4}
OpenAI.
\newblock Gpt-4 technical report, 2023.

\bibitem[{OpenAI}(2023)]{openai2023gpt4v}
{OpenAI}.
\newblock {GPT-4V(ision) System Card}.
\newblock \url{https://cdn.openai.com/papers/GPTV_System_Card.pdf}, 2023.
\newblock Accessed: 2023-11-05.

\bibitem[Oquab et~al.(2023)Oquab, Darcet, Moutakanni, Vo, Szafraniec, Khalidov, Fernandez, Haziza, Massa, El-Nouby, et~al.]{oquab2023dinov2}
Maxime Oquab, Timoth{\'e}e Darcet, Th{\'e}o Moutakanni, Huy Vo, Marc Szafraniec, Vasil Khalidov, Pierre Fernandez, Daniel Haziza, Francisco Massa, Alaaeldin El-Nouby, et~al.
\newblock Dinov2: Learning robust visual features without supervision.
\newblock \emph{arXiv preprint arXiv:2304.07193}, 2023.

\bibitem[Paszke et~al.(2017)Paszke, Gross, Chintala, Chanan, Yang, DeVito, Lin, Desmaison, Antiga, and Lerer]{paszke2017automatic}
Adam Paszke, Sam Gross, Soumith Chintala, Gregory Chanan, Edward Yang, Zachary DeVito, Zeming Lin, Alban Desmaison, Luca Antiga, and Adam Lerer.
\newblock Automatic differentiation in {PyTorch}.
\newblock In \emph{NeurIPS-W}, 2017.

\bibitem[{Peter Selinger}(2024)]{potrace}
{Peter Selinger}.
\newblock {Potrace}.
\newblock \url{https://github.com/tatarize/potrace}, 2024.

\bibitem[Quint(2003)]{quint2003scalable}
Antoine Quint.
\newblock Scalable vector graphics.
\newblock \emph{IEEE MultiMedia}, 10\penalty0 (3):\penalty0 99--102, 2003.

\bibitem[Radford et~al.(2018)Radford, Narasimhan, Salimans, Sutskever, et~al.]{radford2018improving}
Alec Radford, Karthik Narasimhan, Tim Salimans, Ilya Sutskever, et~al.
\newblock Improving language understanding by generative pre-training.
\newblock \emph{openAI}, 2018.

\bibitem[Radford et~al.(2021)Radford, Kim, Hallacy, Ramesh, Goh, Agarwal, Sastry, Askell, Mishkin, Clark, et~al.]{radford2021learning}
Alec Radford, Jong~Wook Kim, Chris Hallacy, Aditya Ramesh, Gabriel Goh, Sandhini Agarwal, Girish Sastry, Amanda Askell, Pamela Mishkin, Jack Clark, et~al.
\newblock Learning transferable visual models from natural language supervision.
\newblock In \emph{International conference on machine learning}, pages 8748--8763. PMLR, 2021.

\bibitem[Ramachandran et~al.(2017)Ramachandran, Zoph, and Le]{ramachandran2017searching}
Prajit Ramachandran, Barret Zoph, and Quoc~V Le.
\newblock Searching for activation functions.
\newblock \emph{arXiv preprint arXiv:1710.05941}, 2017.

\bibitem[Ramesh et~al.(2021{\natexlab{a}})Ramesh, Pavlov, Goh, Gray, Voss, Radford, Chen, and Sutskever]{ramesh2021zero}
Aditya Ramesh, Mikhail Pavlov, Gabriel Goh, Scott Gray, Chelsea Voss, Alec Radford, Mark Chen, and Ilya Sutskever.
\newblock Zero-shot text-to-image generation.
\newblock In \emph{International Conference on Machine Learning}, pages 8821--8831. PMLR, 2021{\natexlab{a}}.

\bibitem[Ramesh et~al.(2021{\natexlab{b}})Ramesh, Pavlov, Goh, Gray, Voss, Radford, Chen, and Sutskever]{ramesh2021zeroshot}
Aditya Ramesh, Mikhail Pavlov, Gabriel Goh, Scott Gray, Chelsea Voss, Alec Radford, Mark Chen, and Ilya Sutskever.
\newblock Zero-shot text-to-image generation, 2021{\natexlab{b}}.

\bibitem[Ramesh et~al.(2022)Ramesh, Dhariwal, Nichol, Chu, and Chen]{ramesh2022hierarchical}
Aditya Ramesh, Prafulla Dhariwal, Alex Nichol, Casey Chu, and Mark Chen.
\newblock Hierarchical text-conditional image generation with clip latents.
\newblock \emph{arXiv preprint arXiv:2204.06125}, 1\penalty0 (2):\penalty0 3, 2022.

\bibitem[Reddy et~al.(2021)Reddy, Gharbi, Lukac, and Mitra]{reddy2021im2vec}
Pradyumna Reddy, Michael Gharbi, Michal Lukac, and Niloy~J Mitra.
\newblock Im2vec: Synthesizing vector graphics without vector supervision.
\newblock \emph{arXiv preprint arXiv:2102.02798}, 2021.

\bibitem[Richardson(2023)]{bs4}
Leonard Richardson.
\newblock Beautifulsoup.
\newblock \url{https://www.crummy.com/software/BeautifulSoup/}, 2023.

\bibitem[Rodriguez et~al.(2023{\natexlab{a}})Rodriguez, Vazquez, Laradji, Pedersoli, and Rodriguez]{rodriguez2023figgen}
Juan~A Rodriguez, David Vazquez, Issam Laradji, Marco Pedersoli, and Pau Rodriguez.
\newblock Figgen: Text to scientific figure generation.
\newblock \emph{arXiv preprint arXiv:2306.00800}, 2023{\natexlab{a}}.

\bibitem[Rodriguez et~al.(2023{\natexlab{b}})Rodriguez, Vazquez, Laradji, Pedersoli, and Rodriguez]{rodriguez2023ocr}
Juan~A Rodriguez, David Vazquez, Issam Laradji, Marco Pedersoli, and Pau Rodriguez.
\newblock Ocr-vqgan: Taming text-within-image generation.
\newblock In \emph{Proceedings of the IEEE/CVF Winter Conference on Applications of Computer Vision}, pages 3689--3698, 2023{\natexlab{b}}.

\bibitem[Rodriguez et~al.(2024)Rodriguez, Botzer, Vazquez, Pal, Pedersoli, and Laradji]{rodriguez2024intentgpt}
Juan~A Rodriguez, Nicholas Botzer, David Vazquez, Christopher Pal, Marco Pedersoli, and Issam Laradji.
\newblock Intentgpt: Few-shot intent discovery with large language models.
\newblock \emph{arXiv preprint arXiv:2411.10670}, 2024.

\bibitem[Rombach et~al.(2021)Rombach, Blattmann, Lorenz, Esser, and Ommer]{rombach2021highresolution}
Robin Rombach, Andreas Blattmann, Dominik Lorenz, Patrick Esser, and Björn Ommer.
\newblock High-resolution image synthesis with latent diffusion models, 2021.

\bibitem[Rombach et~al.(2022)Rombach, Blattmann, Lorenz, Esser, and Ommer]{rombach2022high}
Robin Rombach, Andreas Blattmann, Dominik Lorenz, Patrick Esser, and Bj{\"o}rn Ommer.
\newblock High-resolution image synthesis with latent diffusion models.
\newblock In \emph{Proceedings of the IEEE/CVF conference on computer vision and pattern recognition}, pages 10684--10695, 2022.

\bibitem[S.(2023)]{svgpathtools}
Andy S.
\newblock svgpathtools.
\newblock \url{https://github.com/mathandy/svgpathtools}, 2023.

\bibitem[Schuhmann et~al.(2022)Schuhmann, Beaumont, Vencu, Gordon, Wightman, Cherti, Coombes, Katta, Mullis, Wortsman, et~al.]{schuhmann2022laion}
Christoph Schuhmann, Romain Beaumont, Richard Vencu, Cade Gordon, Ross Wightman, Mehdi Cherti, Theo Coombes, Aarush Katta, Clayton Mullis, Mitchell Wortsman, et~al.
\newblock Laion-5b: An open large-scale dataset for training next generation image-text models.
\newblock \emph{Advances in Neural Information Processing Systems}, 35:\penalty0 25278--25294, 2022.

\bibitem[Shao et~al.(2017)Shao, Gouws, Britz, Goldie, Strope, and Kurzweil]{shao2017generating}
Louis Shao, Stephan Gouws, Denny Britz, Anna Goldie, Brian Strope, and Ray Kurzweil.
\newblock Generating high-quality and informative conversation responses with sequence-to-sequence models.
\newblock \emph{arXiv preprint arXiv:1701.03185}, 2017.

\bibitem[Shazeer(2019)]{shazeer2019fast}
Noam Shazeer.
\newblock Fast transformer decoding: One write-head is all you need.
\newblock \emph{arXiv preprint arXiv:1911.02150}, 2019.

\bibitem[Spector and Re(2023)]{spector2023accelerating}
Benjamin Spector and Chris Re.
\newblock Accelerating llm inference with staged speculative decoding.
\newblock \emph{arXiv preprint arXiv:2308.04623}, 2023.

\bibitem[Srivastava et~al.(2014)Srivastava, Hinton, Krizhevsky, Sutskever, and Salakhutdinov]{srivastava2014dropout}
Nitish Srivastava, Geoffrey Hinton, Alex Krizhevsky, Ilya Sutskever, and Ruslan Salakhutdinov.
\newblock Dropout: a simple way to prevent neural networks from overfitting.
\newblock \emph{The journal of machine learning research}, 15\penalty0 (1):\penalty0 1929--1958, 2014.

\bibitem[Tang et~al.(2021)Tang, Zhang, Xing, Ding, and Xu]{tang2021perlin}
Chengjun Tang, Kun Zhang, Chunfang Xing, Yong Ding, and Zengmin Xu.
\newblock Perlin noise improve adversarial robustness.
\newblock \emph{arXiv preprint arXiv:2112.13408}, 2021.

\bibitem[Theis et~al.(2015)Theis, Oord, and Bethge]{theis2015note}
Lucas Theis, A{\"a}ron van~den Oord, and Matthias Bethge.
\newblock A note on the evaluation of generative models.
\newblock \emph{arXiv preprint arXiv:1511.01844}, 2015.

\bibitem[Touvron et~al.(2023{\natexlab{a}})Touvron, Lavril, Izacard, Martinet, Lachaux, Lacroix, Rozi{\`e}re, Goyal, Hambro, Azhar, et~al.]{touvron2023llama}
Hugo Touvron, Thibaut Lavril, Gautier Izacard, Xavier Martinet, Marie-Anne Lachaux, Timoth{\'e}e Lacroix, Baptiste Rozi{\`e}re, Naman Goyal, Eric Hambro, Faisal Azhar, et~al.
\newblock Llama: Open and efficient foundation language models.
\newblock \emph{arXiv preprint arXiv:2302.13971}, 2023{\natexlab{a}}.

\bibitem[Touvron et~al.(2023{\natexlab{b}})Touvron, Martin, Stone, Albert, Almahairi, Babaei, Bashlykov, Batra, Bhargava, Bhosale, et~al.]{touvron2023llama2}
Hugo Touvron, Louis Martin, Kevin Stone, Peter Albert, Amjad Almahairi, Yasmine Babaei, Nikolay Bashlykov, Soumya Batra, Prajjwal Bhargava, Shruti Bhosale, et~al.
\newblock Llama 2: Open foundation and fine-tuned chat models.
\newblock \emph{arXiv preprint arXiv:2307.09288}, 2023{\natexlab{b}}.

\bibitem[Vaswani et~al.(2017)Vaswani, Shazeer, Parmar, Uszkoreit, Jones, Gomez, Kaiser, and Polosukhin]{vaswani2017attention}
Ashish Vaswani, Noam Shazeer, Niki Parmar, Jakob Uszkoreit, Llion Jones, Aidan~N Gomez, {\L}ukasz Kaiser, and Illia Polosukhin.
\newblock Attention is all you need.
\newblock \emph{Advances in neural information processing systems}, 30, 2017.

\bibitem[Vijayakumar et~al.(2016)Vijayakumar, Cogswell, Selvaraju, Sun, Lee, Crandall, and Batra]{vijayakumar2016diverse}
Ashwin~K Vijayakumar, Michael Cogswell, Ramprasath~R Selvaraju, Qing Sun, Stefan Lee, David Crandall, and Dhruv Batra.
\newblock Diverse beam search: Decoding diverse solutions from neural sequence models.
\newblock \emph{arXiv preprint arXiv:1610.02424}, 2016.

\bibitem[Vinker et~al.(2022)Vinker, Pajouheshgar, Bo, Bachmann, Bermano, Cohen-Or, Zamir, and Shamir]{vinker2022clipasso}
Yael Vinker, Ehsan Pajouheshgar, Jessica~Y Bo, Roman~Christian Bachmann, Amit~Haim Bermano, Daniel Cohen-Or, Amir Zamir, and Ariel Shamir.
\newblock Clipasso: Semantically-aware object sketching.
\newblock \emph{ACM Transactions on Graphics (TOG)}, 41\penalty0 (4):\penalty0 1--11, 2022.

\bibitem[{Vision Cortex}(2023)]{vtracer}
{Vision Cortex}.
\newblock {VTracer}.
\newblock \url{https://www.visioncortex.org/vtracer-docs}, 2023.

\bibitem[Wang et~al.(2024)Wang, Bai, Tan, Wang, Fan, Bai, Chen, Liu, Wang, Ge, et~al.]{wang2024qwen2}
Peng Wang, Shuai Bai, Sinan Tan, Shijie Wang, Zhihao Fan, Jinze Bai, Keqin Chen, Xuejing Liu, Jialin Wang, Wenbin Ge, et~al.
\newblock Qwen2-vl: Enhancing vision-language model's perception of the world at any resolution.
\newblock \emph{arXiv preprint arXiv:2409.12191}, 2024.

\bibitem[Wang and Lian(2021)]{wang2021deepvecfont}
Yizhi Wang and Zhouhui Lian.
\newblock Deepvecfont: Synthesizing high-quality vector fonts via dual-modality learning.
\newblock \emph{ACM Transactions on Graphics (TOG)}, 40\penalty0 (6):\penalty0 1--15, 2021.

\bibitem[Wang and Bovik(2009)]{wang2009mean}
Zhou Wang and Alan~C Bovik.
\newblock Mean squared error: Love it or leave it? a new look at signal fidelity measures.
\newblock \emph{IEEE signal processing magazine}, 26\penalty0 (1):\penalty0 98--117, 2009.

\bibitem[Wang et~al.(2004)Wang, Bovik, Sheikh, and Simoncelli]{wang2004image}
Zhou Wang, Alan~C Bovik, Hamid~R Sheikh, and Eero~P Simoncelli.
\newblock Image quality assessment: from error visibility to structural similarity.
\newblock \emph{IEEE transactions on image processing}, 13\penalty0 (4):\penalty0 600--612, 2004.

\bibitem[Wikipedia(2024)]{wiki:Comparison_of_raster-to-vector_conversion_software}
Wikipedia.
\newblock {Comparison of raster-to-vector conversion software} --- {W}ikipedia{,} the free encyclopedia.
\newblock \url{http://en.wikipedia.org/w/index.php?title=Comparison\%20of\%20raster-to-vector\%20conversion\%20software&oldid=1185354750}, 2024.
\newblock [Online; accessed 07-March-2024].

\bibitem[Wolf et~al.(2019)Wolf, Debut, Sanh, Chaumond, Delangue, Moi, Cistac, Rault, Louf, Funtowicz, et~al.]{wolf2019huggingface}
Thomas Wolf, Lysandre Debut, Victor Sanh, Julien Chaumond, Clement Delangue, Anthony Moi, Pierric Cistac, Tim Rault, R{\'e}mi Louf, Morgan Funtowicz, et~al.
\newblock Huggingface's transformers: State-of-the-art natural language processing.
\newblock \emph{arXiv preprint arXiv:1910.03771}, 2019.

\bibitem[Wu et~al.(2023)Wu, Su, Ma, and Liao]{wu2023iconshop}
Ronghuan Wu, Wanchao Su, Kede Ma, and Jing Liao.
\newblock Iconshop: Text-based vector icon synthesis with autoregressive transformers.
\newblock \emph{arXiv preprint arXiv:2304.14400}, 2023.

\bibitem[Xia et~al.(2009)Xia, Liao, and Yu]{xia2009patch}
Tian Xia, Binbin Liao, and Yizhou Yu.
\newblock Patch-based image vectorization with automatic curvilinear feature alignment.
\newblock \emph{ACM Transactions on Graphics (TOG)}, 28\penalty0 (5):\penalty0 1--10, 2009.

\bibitem[Yu et~al.(2022)Yu, Xu, Koh, Luong, Baid, Wang, Vasudevan, Ku, Yang, Ayan, et~al.]{yu2022scaling}
Jiahui Yu, Yuanzhong Xu, Jing~Yu Koh, Thang Luong, Gunjan Baid, Zirui Wang, Vijay Vasudevan, Alexander Ku, Yinfei Yang, Burcu~Karagol Ayan, et~al.
\newblock Scaling autoregressive models for content-rich text-to-image generation.
\newblock \emph{arXiv preprint arXiv:2206.10789}, 2\penalty0 (3):\penalty0 5, 2022.

\bibitem[Zhai et~al.(2023)Zhai, Mustafa, Kolesnikov, and Beyer]{zhai2023sigmoid}
Xiaohua Zhai, Basil Mustafa, Alexander Kolesnikov, and Lucas Beyer.
\newblock Sigmoid loss for language image pre-training.
\newblock In \emph{Proceedings of the IEEE/CVF International Conference on Computer Vision}, pages 11975--11986, 2023.

\bibitem[Zhang et~al.(2018)Zhang, Isola, Efros, Shechtman, and Wang]{zhang2018unreasonable}
Richard Zhang, Phillip Isola, Alexei~A Efros, Eli Shechtman, and Oliver Wang.
\newblock The unreasonable effectiveness of deep features as a perceptual metric.
\newblock In \emph{Proceedings of the IEEE conference on computer vision and pattern recognition}, pages 586--595, 2018.

\bibitem[Zhang et~al.(2023)Zhang, Liu, Zhang, Cheng, and Wang]{zhang2023pixelsexploringhumanreadablesvg}
Tong Zhang, Haoyang Liu, Peiyan Zhang, Yuxuan Cheng, and Haohan Wang.
\newblock Beyond pixels: Exploring human-readable svg generation for simple images with vision language models, 2023.

\bibitem[Zou et~al.(2024)Zou, Cai, Zhang, and Lee]{zou2024vgbench}
Bocheng Zou, Mu Cai, Jianrui Zhang, and Yong~Jae Lee.
\newblock Vgbench: Evaluating large language models on vector graphics understanding and generation.
\newblock \emph{arXiv preprint arXiv:2407.10972}, 2024.

\end{thebibliography}
}
\newpage
\clearpage
\maketitlesupplementary

In the following sections, we provide additional details on the datasets used in this paper, present further experiments, and describe our baselines in detail. We also discuss the StarVector architecture, its training process, and the method for sampling SVG code from the model. Additionally, we provide more insights into SVG-Bench, including the proposed datasets and the different baselines within the evaluation setup. Finally, qualitative results are presented to showcase the strengths and limitations of our foundational model.

\section{SVG Datasets in SVG-Bench}\label{app:datasets}
\begin{table*}[t]
    \centering
        \caption{\textbf{Summary of datasets.} We offer a summary of statistics about the datasets used in our training and evaluation experiments. This datasets are included in SVG-Bench. The subscript \textit{sim} stands for the simplified version of the dataset, as required by some baselines.}
    \resizebox{1.0\textwidth}{!}{
    \begin{tabular}{lccccccccc}
        \hline
        \textbf{Dataset}  & \textbf{Train} & \textbf{Val} & \textbf{Test} & \textbf{Source} & \textbf{Token Length} & \textbf{SVG Primitives} &\textbf{Annotation}\\
        \hline
        
        \textbf{SVG-Stack} & 2,1M & 108k & 5,7k & \multirow{3}{*}{TheStack~\citep{kocetkov2022stack}} & 1,822 $\pm$ 1,808 & All & Caption\\
        
        \textbf{SVG-Stack}\textsubscript{sim}  & 601k & 30,k & 1,5k &  & 2k $\pm$ 918 & Vector path &  Caption\\
        
        \textbf{SVG-Diagrams} &-& - & 472 && 3,486 $\pm$ 1,918 & All &  Caption\\
        \midrule
        \textbf{SVG-Fonts}  & 1,8M & 91,5k & 4,8k & \multirow{2}{*}{Glypazzn~\citep{lopes2019learned}} & 2,121 $\pm$ 1,868 & Vector path & Font type\\
        \textbf{SVG-Fonts}\textsubscript{sim}  & 1,4M & 71,7k & 3,7k & & 1,722 $\pm$ 723 & Vector path & Font type\\
        \midrule                
        \textbf{SVG-Emoji}& 8,7k & 667 & 668 & \multirow{2}{*}{\makecell{OpenMoji, 
        NotoEmoji, TweMoji}} & 2,551 $\pm$ 1,805 & All & Class\\
        \textbf{SVG-Emoji}\textsubscript{sim} & 580 & 57 & 96 & & 2,448 $\pm$ 1,026 & Vector Path & Class\\
               \midrule
        \textbf{SVG-Icons} & 80,4k & 6,2k & 2,4k & \multirow{2}{*}{DeepSVG~\citep{carlier2020deepsvg}} &  2,449 $\pm$ 1,543 & Vector path & -\\
        \textbf{SVG-Icons}\textsubscript{sim}  & 80,435 & 2,836 & 1,277 &   &  2,005 $\pm$ 824 & Vector path & -\\
        \midrule
        \textbf{SVG-FIGR} \textsuperscript{\textdaggerdbl} & 270k & 27k & 3k & IconShop~\citep{wu2023iconshop} & 5,342 $\pm$ 2,345  & Vector path& Class, Caption\\
        
        \hline
    \end{tabular}
    }
    \label{tab:all-datasets}
\end{table*}

\begin{figure*}
    \centering
    \includegraphics[width=1.0\linewidth]{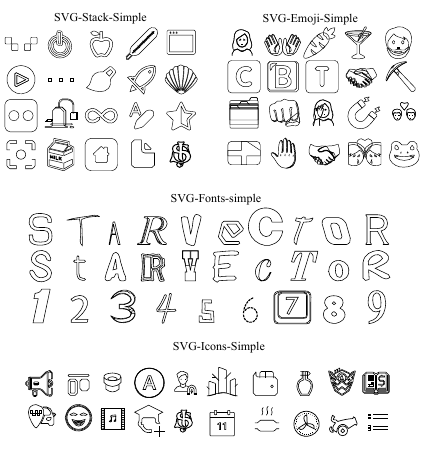}
    \caption{\textbf{Simplified Datasets in SVG Bench.} Ground truth test examples from the simplified test sets of \textbf{SVG-Stack}, \textbf{SVG-Emoji}, \textbf{SVG-Fonts}, and \textbf{SVG-Icons}. We show SVG images.}
    \label{fig:simple-datasets}
\end{figure*}

Here we describe available SVG datasets in the recent literature. We extend our description of the datasets used for training and evaluating StarVector and other baselines. Earlier SVG datasets proposed in the literature (mainly datasets of emojis and fonts) were not easily accessible due to broken URLs and no direct entry point. Therefore, we provide them as part of SVGBench for easy reproducibility. We introduce splits for train, validation, and testing. The train set is used to optimize the parameter weights of the network. The validation is used for tuning sampling hyperparameters, and the test is used for evaluation. Our model can handle up to 8k context tokens. Therefore, our datasets only consider examples with up to 8,192 tokens. See Table~\ref{tab:all-datasets} for a complete description of the datasets. See Figures~\ref{fig:diagrams}, \ref{fig:colored-datasets}, \ref{fig:simple-datasets} for ground truth examples of the test sets of SVG-Bench.

\subsection{Datasets with Simplified SVGs.} We create simplified versions of our four main datasets, i.e. emojis, icons, fonts, and SVG-Stack. This is done because DeepSVG~\citep{carlier2020deepsvg} requires a simplification of the SVG in its input. The simplification consists of eliminating complex primitives and using only vector \textit{paths}. Also, color and shapes are abstracted only to use simple line strokes.

\subsection{Creating the SVG-Fonts Dataset}
\begin{figure*}
    \centering
    \includegraphics[width=1.0\linewidth]{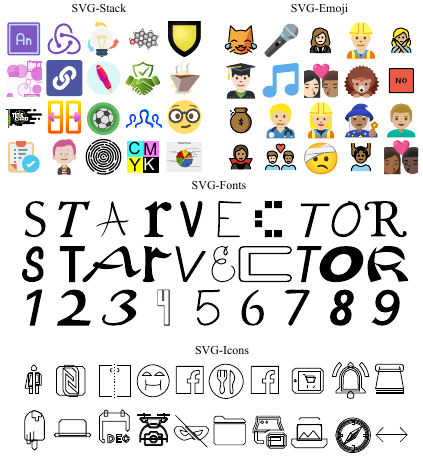}
    \caption{\textbf{Datasets in SVG-Bench.} Ground truth test examples from the test sets of \textbf{SVG-Stack}, \textbf{SVG-Emoji}, \textbf{SVG-Fonts}, and \textbf{SVG-Icons}. We show SVG images.}
    \label{fig:colored-datasets}
\end{figure*}

To construct the SVG-Font dataset, we replicate the procedure described in SVG-VAE~\citep{lopes2019learned}\footnote{\url{https://github.com/magenta/magenta/tree/main/magenta/models/svg_vae}}, which provides a list of public URLs containing open font packages. We download these packages, excluding any with broken URLs. The TTF files are then converted to SFD format, and we further use InkScape\footnote{\url{https://inkscape.org/}} to convert them into SVG code. Samples from the test set are shown in Figure~\ref{fig:colored-datasets} (bottom-left). These samples contain only path elements and represent a narrow range of images, each consisting of a single character per image, making them an ideal test case for SVG generation~\citep{lopes2019learned, carlier2020deepsvg, cao2023svgformer}.

\subsection{Creating SVG-Diagrams}
\begin{figure}[t]
    \centering
    \includegraphics[width=1.0\linewidth]{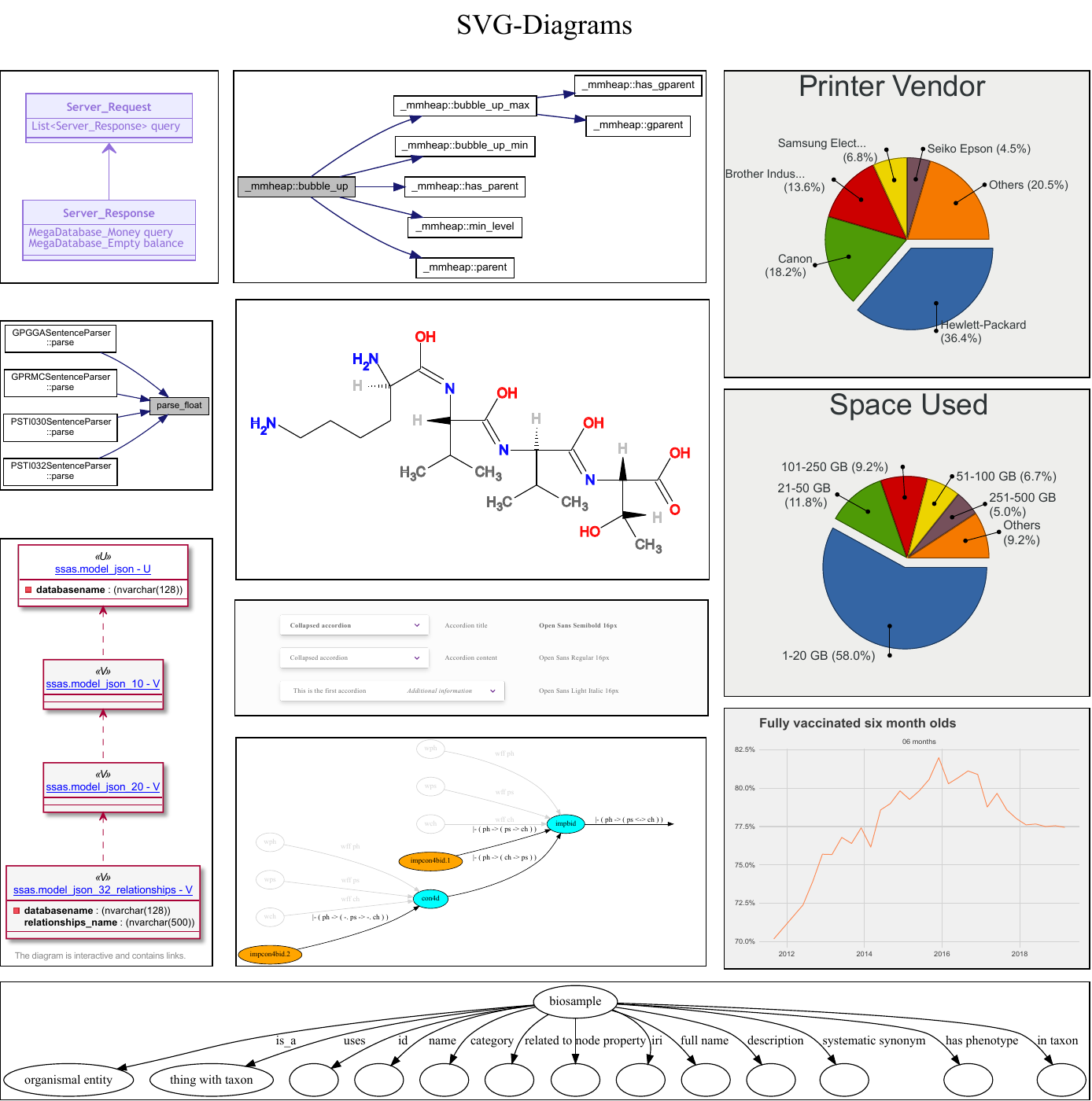}
    \caption{SVG Diagrams examples. These are ground truth SVG examples from the test set. They are presented as SVG, showing the challenge of understanding intricate structures and small texts, with images of variate aspect ratios.}
    \label{fig:diagrams}
\end{figure}

We introduce a novel SVG dataset, SVG-Diagrams, which focuses exclusively on diagrams, graphs, workflows, and other designs characterized by discrete elements such as boxes, arrows, and text. To construct this dataset, we filtered all SVGs containing the text element. Table~\ref{tab:all-datasets} provides detailed statistics about the dataset, and Figure~\ref{fig:diagrams} illustrates test samples from SVG-Diagrams.

This new benchmark for diagram generation is highly relevant, as it addresses a use case that cannot be tackled by traditional image-processing models or methods limited to the path primitive. Only approaches capable of leveraging SVG code can fully exploit the use of primitives to generate such structured designs effectively.

\subsection{Generating Synthetic Captions on SVG-Stack }\label{app:nnotation-svg-stack}

To generate textual instructions for vector images, we process SVG-Stack images using visual captioning models. This approach provides a textual description for each image, enabling us to fine-tune our model to follow textual instructions. For this task, we leverage off-the-shelf AI captioners, specifically BLIP2~\citep{li2023blip} and Llava~\citep{liu2023visual}. 

Through prompt engineering, we guide these models in performing the captioning task with reasonable quality. The prompt we used is shared in Prompt~\ref{svg-stack-captioning}. After automatically captioning all SVG samples in SVG-Stack, we compute the CLIP Score for the text-image pairs generated by the two models—producing two captions per image. Using a CLIP Score threshold of 30, we filter out text captions that fall below this threshold.
\vspace{1cm}
\prompt{1. Utilized with BLIP2 and Llava for SVG Captioning}{You are a helpful assistant. Your task is to caption the input images with a concise and clear description that represents what are the contents of the image.}\label{svg-stack-captioning}
\vspace{1cm}
\subsection{Data Augmentation for SVG}
\label{app:starvector-augmentations}
We introduce several data augmentation operations on SVGs that aim to perform minor modifications to the SVG code and rasterize it to get a new sample while training. We include rotation, color, and curve noise. We evaluate this setting on datasets with fewer samples, namely SVG-Emoji and SVG-Icons, as the other two datasets are large enough to do not overfit. Results are shown in Table~\ref{tab:augmentations}. Both datasets display improvements using these augmentations. We see a substantial uplift for SVG-Emoji, which has limited training data.

We introduce several augmentation operations to SVGs to apply slight changes that help our model learn to generate more precise results — for instance, being able to capture exact colors from the image and encode them in hexadecimal code to insert it in the \texttt{fill} attribute of the SVG element. Applying rotations or adding noise to the curve's control points helps the model learn to precisely capture the position of the edges or thickness of the stroke.

We perform random rotations in an angle range. We perform color changes by first parsing the element's color using the \texttt{fill} attribute and adding slight white Gaussian noise to the RGB values. We propose curve noise by injecting a small Perlin~\cite{tang2021perlin} noise into the control points in Bézier curves. We also experimented with adding Gaussian noise, which resulted in much less natural results. We apply this noise by uniformly sampling a scalar from the interval between 0.01 and 0.05 and use it to scale the noise.
 
We apply these augmentations directly on the SVG code, which involves parsing the XML code and accessing the attributes and arguments of the primitives defined. We use the libraries \texttt{BeautifulSoup}\footnote{\url{https://www.crummy.com/software/BeautifulSoup/bs4/doc/}} and \texttt{SvgPathTools}\footnote{\url{https://github.com/mathandy/svgpathtools}}. Some primitives are simplified using our augmentations. 

\section{SVG Methods and Baselines}\label{app:baselines}

Here, we describe the previous methods and baselines used to compare StarVector's performance in Image-to-SVG and Text-to-SVG generation tasks. We consider previous deep learning-based methods and image-processing methods. We evaluate the baselines with publicly available code in our proposed setup. 

\definecolor{mygreen}{rgb}{0.173, 0.549, 0.204}
\newcommand{\green}[1]{\textcolor{mygreen}{#1}}
\definecolor{myred}{rgb}{0.737, 0.141, 0.173}
\newcommand{\crossred}[1]{\textcolor{myred}{#1}}
\begin{table*}
\centering
\caption{\textbf{Summary of SVG Methods.} We compare SVG generation methods based on \textit{Image Processing}, \textit{Latent Variable}, \textit{Differentiable Rendering}, and \textit{Multimodal LLM}, evaluating their performance in \textit{SVG Generalization} and \textit{SVG Generation Tasks}. The \textit{SVG Generalization} column shows whether a model generates diverse SVG types (e.g., icons, logos, complex shapes) with \green{\ding{51}} or specializes in a subtype (e.g., emojis, fonts) with \crossred{\ding{55}}. The \textbf{SVG Primitive Coverage} column indicates access to all SVG primitives. The table also evaluates \textbf{Image Vectorization}, \textbf{Text to SVG}, and \textbf{Diagram Generation}, using \green{\ding{51}} for support and \crossred{\ding{55}} for limitations. $^*$DeepSVG requires modifications for image input.}

\resizebox{1.0\textwidth}{!}{
\begin{tabular}[t]{
    >{\raggedright\arraybackslash}p{3cm} 
    >{\raggedright\arraybackslash}p{2.5cm} 
    >{\centering\arraybackslash}p{2cm}
    >{\centering\arraybackslash}p{2.5cm} 
    >{\centering\arraybackslash}p{2.7cm} 
    >{\centering\arraybackslash}p{2.9cm} 
    >{\centering\arraybackslash}p{2.5cm} 
    >{\centering\arraybackslash}p{2cm} 
    >{\centering\arraybackslash}p{2cm}
}
\hline

& & & & \multicolumn{2}{c}{{\textit{SVG Coverage and Generalization}}} & \multicolumn{3}{c}{{\textit{SVG Generation Tasks}}} \\

 \textbf{Method Type} & \textbf{Model} &  \textbf{Input} & \textbf{Train Supervision} & \textbf{SVG Generalization} &  \textbf{SVG Primitive Coverage} &  \textbf{Image Vectorization} &  \textbf{Text to SVG} &  \textbf{Diagram Generation} \\

\hline
\multirow{3}{*}{{\textbf{Image Processing}}} 
& \textbf{Vtracer} & Image & Image & \green{\ding{51}} & \crossred{\ding{55}} & \green{\ding{51}} & \crossred{\ding{55}} & \crossred{\ding{55}} \\
& \textbf{Autotrace} & Image & Image & \green{\ding{51}} & \crossred{\ding{55}} & \green{\ding{51}} & \crossred{\ding{55}} & \crossred{\ding{55}} \\
& \textbf{Potrace} & Image & Image &\green{\ding{51}} & \crossred{\ding{55}} & \green{\ding{51}} & \crossred{\ding{55}} & \crossred{\ding{55}} \\
\hline

\multirow{3}{*}{{\textbf{Latent Variable}}} 
& \textbf{Im2Vec} & Image & Image & \crossred{\ding{55}} & \crossred{\ding{55}} & \green{\ding{51}} & \crossred{\ding{55}} & \crossred{\ding{55}} \\
& \textbf{DeepSVG} & SVG & Vector & \crossred{\ding{55}} & \crossred{\ding{55}} & \green{\ding{51}}$^*$ & \crossred{\ding{55}} & \crossred{\ding{55}} \\
& \textbf{SVGFormer} & SVG & Vector & \crossred{\ding{55}} & \crossred{\ding{55}} & \green{\ding{51}} & \crossred{\ding{55}} & \crossred{\ding{55}} \\
\hline

\multirow{4}{*}{{\textbf{Diff. Rendering}}} 
& {\textbf{DiffVG}} & Image & Image & \green{\ding{51}} & \crossred{\ding{55}} & \green{\ding{51}} & \crossred{\ding{55}} & \crossred{\ding{55}} \\
& {\textbf{LIVE}} & Image & Image &\green{\ding{51}} & \crossred{\ding{55}} & \green{\ding{51}} & \crossred{\ding{55}} & \crossred{\ding{55}} \\
& {\textbf{SAMVG}} & Image, Text & Image &\green{\ding{51}} & \crossred{\ding{55}} & \green{\ding{51}} & \crossred{\ding{55}} & \crossred{\ding{55}} \\
& {\textbf{SVGDreamer}} & Image, Text & Image &\green{\ding{51}} & \crossred{\ding{55}} & \crossred{\ding{55}} & \green{\ding{51}} & \crossred{\ding{55}} \\
\hline

\multirow{4}{*}{{\textbf{Multimodal LLM}}} 
& \textbf{GPT-4 V} & Image, Text & SVG & \green{\ding{51}} & \green{\ding{51}} & \crossred{\ding{55}} & \green{\ding{51}} & \green{\ding{51}} \\
& \textbf{CodeLlama} & Image & SVG &  \green{\ding{51}} & \green{\ding{51}} & \crossred{\ding{55}} & \green{\ding{51}} & \green{\ding{51}} \\
& \textbf{IconShop} & Text & SVG & \crossred{\ding{55}} & \crossred{\ding{55}} & \crossred{\ding{55}} & \green{\ding{51}} & \crossred{\ding{55}} \\
\rowcolor{gray!5}
& \textbf{StarVector} & Image, Text & SVG &  \green{\ding{51}} & \green{\ding{51}} & \green{\ding{51}} & \green{\ding{51}} & \green{\ding{51}} \\
\hline
\end{tabular}
}

\label{tab:related-work_comparison}
\end{table*}

\subsection{Image-to-SVG Baselines}

We reproduce all previous approaches on our proposed SVG-Bench benchmark, as the available results stem from an unclear version of the fonts, emojis, and icons datasets. For theImage-to-SVG task, we consider several baseline methods across deep learning and image processing approaches.

In the deep learning methods category, we start with DeepSVG\cite{carlier2020deepsvg}, Im2Vec\cite{reddy2021im2vec}, and LIVE\cite{ma2022towards}, using the official implementations with the hyperparameters proposed by the authors, and applying their pre- and post-processing code as required. Additionally, we incorporate the recent GPT-4 Vision\cite{openai2023gpt4v}, which is capable of processing images as input and generating corresponding SVG code as output.

For the image processing-based methods, which do not rely on data-driven learning, we consider VTracer\cite{vtracer}, Autotracer\cite{autotrace}, and Potrace~\cite{potrace}, running them on the test sets of SVG-Bench.

\noindent\textbf{{Autotrace}}\footnote{\url{ https://potrace.sourceforge.net/}}~\cite{autotrace} is a tool designed for converting images to vector graphics, similar to Potrace. It supports various input formats and can output to several vector formats. Autotrace's key feature is its ability to transform pixelated images into smooth, scalable vectors, making it ideal for upgrading images for various applications without losing detail or clarity. Our experiments leverage the Python bindings\footnote{\url{https://github.com/lemonyte/pyautotrace}} implementation of AutoTrace.

\vspace{1cm}\noindent\textbf{{Potrace}}\footnote{\url{https://potrace.sourceforge.net/}}~\cite{potrace} is a utility designed to convert images into refined, scalable vector graphics. It accepts input in various bitmap formats and outputs to a selection of vector formats. This functionality is particularly valuable for generating SVG of scanned imagery, such as logos and handwritten documents. We employ a Python library\footnote{\url{https://github.com/tatarize/potrace}}, which acts as a wrapper around the original C implementation of Potrace.

\vspace{1cm}\noindent\textbf{{VTracer}}\footnote{\url{https://github.com/visioncortex/vtracer}}~\cite{vtracer} is an image processing algorithm to convert images to SVGs. This 3-step pipeline algorithm relies on the hierarchical clustering of images, which are traced into vectors. First, pixels are converted into paths and then simplified into polygons. In the last step, polygons are smoothened and approximated with a Bezier curve fitter. We use the Python library\footnote{\url{https://github.com/etjones/vtracer_py}} for experiments, a wrapper over the Rust implementation. Similar to Im2Vec, we scale down all the images to 128X128 resolution. We use all the default values for the image processing engine, which generates a multi-colored SVG.

\vspace{1cm}\noindent\textbf{{Im2Vec}}~\citep{reddy2021im2vec} uses an end-to-end VAE, trained using only image supervision to produce vector graphics. The input rasterized image is encoded to a `global' latent vector and passed to an RNN to produce latent code for each path. The path decoder decodes these codes into Bezier paths to generate the output SVG. We used the publicly available code\footnote{\url{https://github.com/preddy5/Im2Vec}} to report the results. 

We scaled all the images to $128\times128$ resolution to be compatible with the Im2Vec model. We used a learning rate of $5 \times 10^{-4}$ and a batch size of 8. We implemented a custom post-processing operation for converting the vector parameters obtained during Im2Vec inference to obtain compilable SVG code.

\vspace{1cm}\noindent\textbf{{LIVE}}, (Layer-wise Image Vectorization)~\citep{ma2022towards} is a method for progressively generating SVGs that closely fit a given raster image by recursively adding and optimizing closed vector \textit{paths}. Using a differentiable renderer (based on DiffVG~\citep{li2020differentiable}), LIVE enables direct optimization of paths under raster image supervision while controlling shape complexity by adjusting the number of path segments. It introduces component-wise path initialization, identifying key visual components to ensure efficient topology extraction and minimize redundant shapes. LIVE achieves high-quality reconstructions with fewer paths, reducing SVG file size compared to other approaches. Nevertheless, its test time optimization approach makes it time-consuming during generation. We utilized their official open-source implementation \footnote{\url{https://github.com/ma-xu/LIVE}} with the proposed hyperparameters. This method requires to define a constant number of \textit{paths}; the more paths defined, the more accurate. We have performed an ablation on the number of paths (see Table~\ref{tab:paths-and-inference}) and found that paths=32 is an optimal value that brings good visual results. However, it takes more than 10 minutes to generate a single SVG, which makes it slow for a professional use case.

\vspace{1cm}\noindent\textbf{{DiffVG}}~\citep{li2020differentiable} is a landmark in vector graphics research, pioneering deep learning-based methods with the first differentiable vector graphics rasterization pipeline. By leveraging a combination of anti-aliasing techniques and gradient-based optimization, DiffVG ensures differentiability. Unlike methods relying on non-differentiable curve-to-mesh conversions, DiffVG employs a forward-backward rasterization process, where the forward pass generates anti-aliased images and the backward pass computes gradients with respect to vector graphic parameters. Using the official implementation\footnote{\url{https://github.com/BachiLi/diffvg}} and proposed hyperparameters, we ablate the number of \textit{paths}, finding \textit{paths}=60 to be optimal. DiffVG balances versatility and performance, achieving approximately 30 seconds per generation while excelling in differentiable rendering tasks.

\vspace{1cm}\noindent\textbf{{DeepSVG}}~\cite{carlier2020deepsvg} was introduced as a hierarchical path-based VAE encoder-decoder transformer architecture. Here, input paths are encoded separately using a path encoder and aggregated using a second encoder to produce a latent vector. The decoder uses this latent vector to output the path representations, which provide actual draw commands and arguments. We used the open-source code\footnote{\url{https://github.com/alexandre01/deepsvg}} to reproduce the results on different datasets. However, since the DeepSVG framework only allows simplified SVGs, we report results on the `simplified' test sets (see Table~\ref{tab:consolidated-image-vectorization-results}). 

\begin{figure}[!h]
    \centering
    \includegraphics[width=1.0\columnwidth]{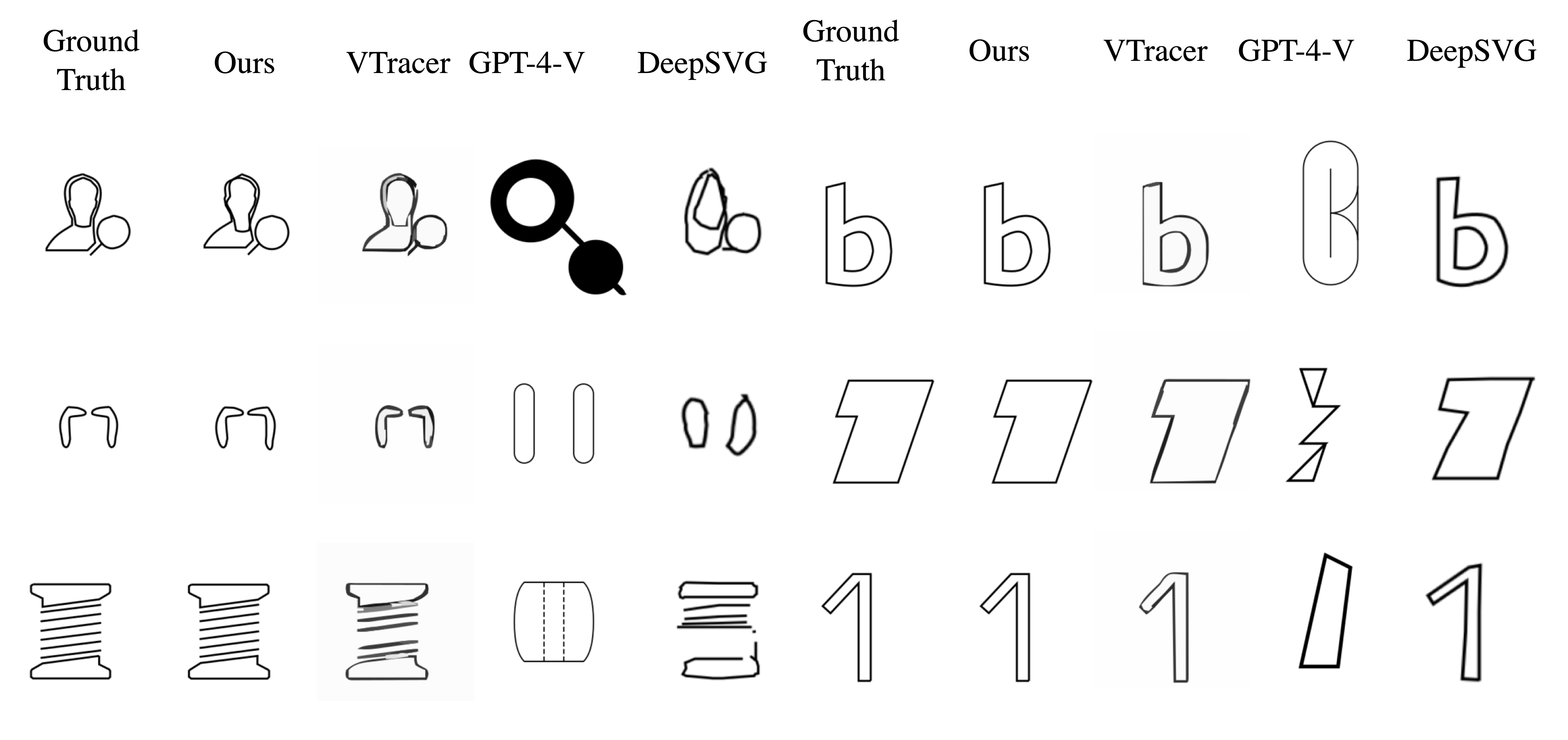}
    \caption{\textbf{Image-to-SVG} results on simplified SVG-Icons and SVG-Fonts test set.}
    \label{fig:results-icons-simplified}
\end{figure}

This model can only handle simplified SVGs composed of simple line strokes and splines (see examples in Figure~\ref{fig:results-icons-simplified}). Further, it can only process SVGs with eight groups (i.e., groups of shapes or parent nodes) and vector paths of at most 30 commands. To reproduce the DeepSVG baseline, we use the original hyperparameters, including a learning rate of $1e-3$ and a number of epochs of 50. We use a batch size of 200, except for the smaller emoji dataset, where we experiment with a batch size of 50.

\vspace{1cm}\noindent\textbf{{GPT-4 Vision.}} We use GPT-4V~\cite{openai2023gpt4v} by inserting an image and zero-shot prompting to generate SVG code. Here, we show how one can use prompt engineering~\cite{brown2020language, bubeck2023sparks, openai2023gpt4v} to condition the model to generate executable SVG code representing the given image. Prompt~\ref{gpt4-prompt} was used for this endeavor. We use the OpenAI library\footnote{\url{https://platform.openai.com/docs/libraries}}.

\begin{figure}[!h]
    \centering
    \includegraphics[width=1.0\columnwidth]{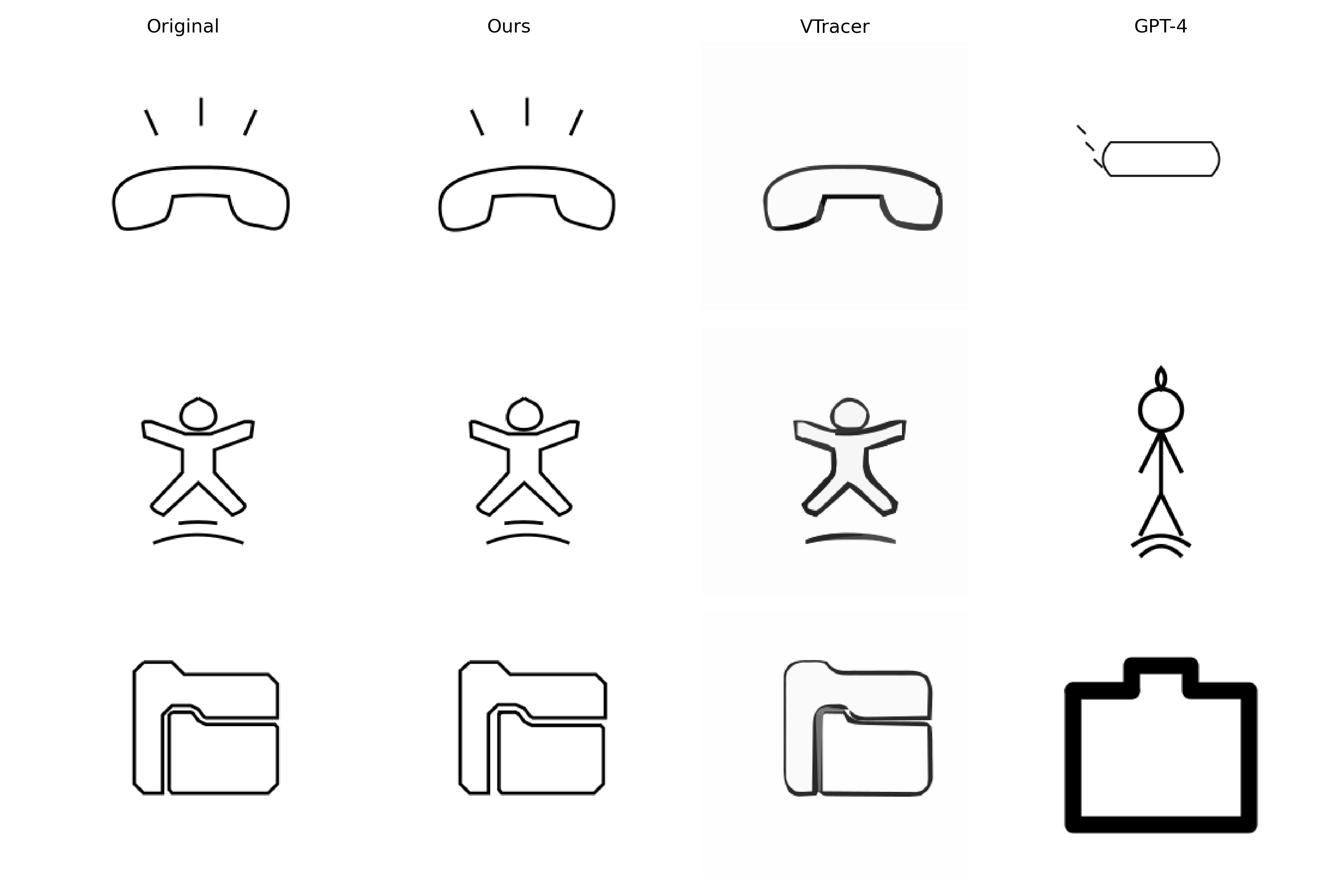}
    \caption{\textbf{Image-to-SVG} results on SVG-Icons test set.}
    \label{fig:results-icons}
\end{figure}

\begin{figure}[!h]
    \centering
    \includegraphics[width=0.98\linewidth]{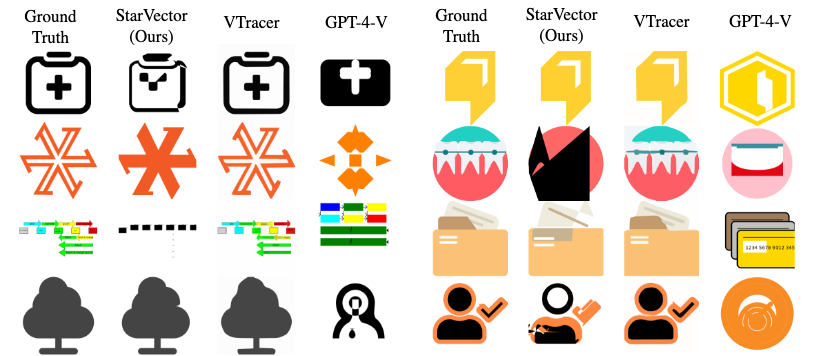}
    \caption{\textbf{Image-to-SVG} results on SVG-Stack test set. We show cherry-picked failure examples of StarVector.}
    \label{fig:failure-stack}
\end{figure}

\begin{figure}[!h]
    \centering
    \includegraphics[width=0.98\linewidth]{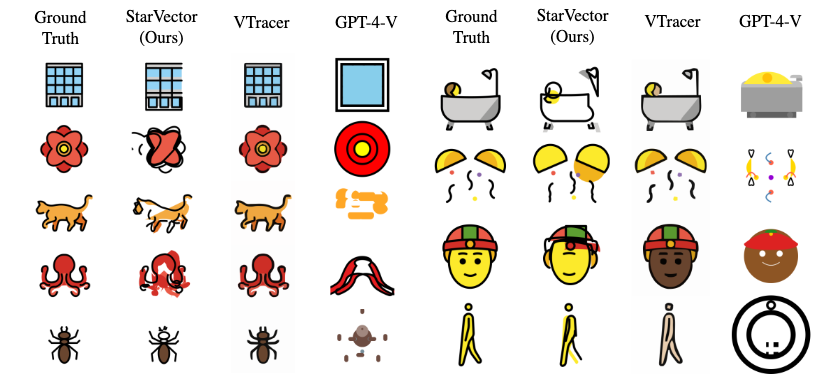}
    \caption{\textbf{Image-to-SVG} results on SVG-Emoji test set. We show cherry-picked failure examples of StarVector.}
    \label{fig:failure-emoji}
\end{figure}

\subsection{Text-to-SVG Generation Baselines}

For the Text-conditioned SVG generation task, we select baselines based on works that contain reproducible methodologies in public datasets or public code repositories. We reproduce baseline models from their official repositories, respecting the proposed hyperparameters.

\vspace{1cm}\noindent\textbf{{CodeLlama, }}~\cite{touvron2023llama, touvron2023llama2} has shown great success in general coding benchmarks. To the best of our knowledge, CodeLlama has seen SVGs during training. Hence, it is reasonable to consider it a strong baseline for text-conditioned SVG generation. We use Anyscale endpoints\footnote{\url{https://app.endpoints.anyscale.com/}} to generate CodeLlama results.

\vspace{1cm}\noindent\textbf{{GPT 4}}, is a closed source LLM that shows state-of-the-art results in many NLP sceneraios~\citep{openai2023gpt4, rodriguez2024intentgpt, bubeck2023sparks}. We evaluate GPT-4's 0-shot ability in generating SVGs when prompted with text inputs. We use OpenAI API\footnote{\url{https://platform.openai.com/docs/guides/gpt}} to generate results for GPT-4 in the 0-shot setting. Prompt~\ref{prompt-text2svg} was used for the Text-to-SVG task.

\vspace{5px}

\prompt{2. Used on GPT4-V VLM for Image-to-SVG Translation}{You are a helpful assistant. Your task is to help researchers write SVG code to reconstruct the provided image as accurately as possible. You should also provide a caption for the image. You are dedicated to solving the task of Image-to-SVG conversion for a robust system. Therefore, you must always respond with the best SVG code you can create. Feel free to use multiple paths to generate a compliant SVG code within a maximum of 8000 tokens. You should present the SVG code that best reconstructs the input image enclosed in triple quotes.}\label{gpt4-prompt}

\vspace{5px}

\prompt{3. Used on GPT4 and CodeLlama for Text-to-SVG Generation}{You are a helpful assistant assisting researchers in generating SVG code from textual descriptions. You will be provided with details to guide your SVG creation. Your task is to write SVG code that accurately represents the given textual information to the fullest extent possible. You are committed to solving the task of SVG generation for a robust system, so always strive to produce the best SVG code you can. Feel free to use multiple paths and any necessary shapes, colors, or lines to generate compilable SVG code within a maximum of 9000 tokens. The goal is to ensure the resulting SVG, when rasterized, best represents the described content. Respond only with the SVG code, enclosed in triple quotes, that directly corresponds to the provided textual description. Avoid adding any explanation or commentary.
}\label{prompt-text2svg}

\vspace{5px}

\vspace{1cm}\noindent\textbf{{IconShop}}
IconShop~\citep{wu2023iconshop} uses a transformer-based architecture to encode path commands and learn to model SVG path sequences autorregressively. It has shown excellent results in simplified icon scenarios and provides a good solution to Text-to-SVG generation by extending the FIGR-SVG dataset with captions. We have access to their dataset and original splits and have trained our model on that data using a pre-trained checkpoint (trained on SVG-Stack). We have extracted the results from IconShop and included them here to compare our method.

\section{Additional Experiments and Results}\label{app:additional-results}
\begin{figure*}[!ht]
    \centering
    \includegraphics[width=0.96\textwidth, trim=2.5cm 6.8cm 2.5cm 2.5cm, clip]{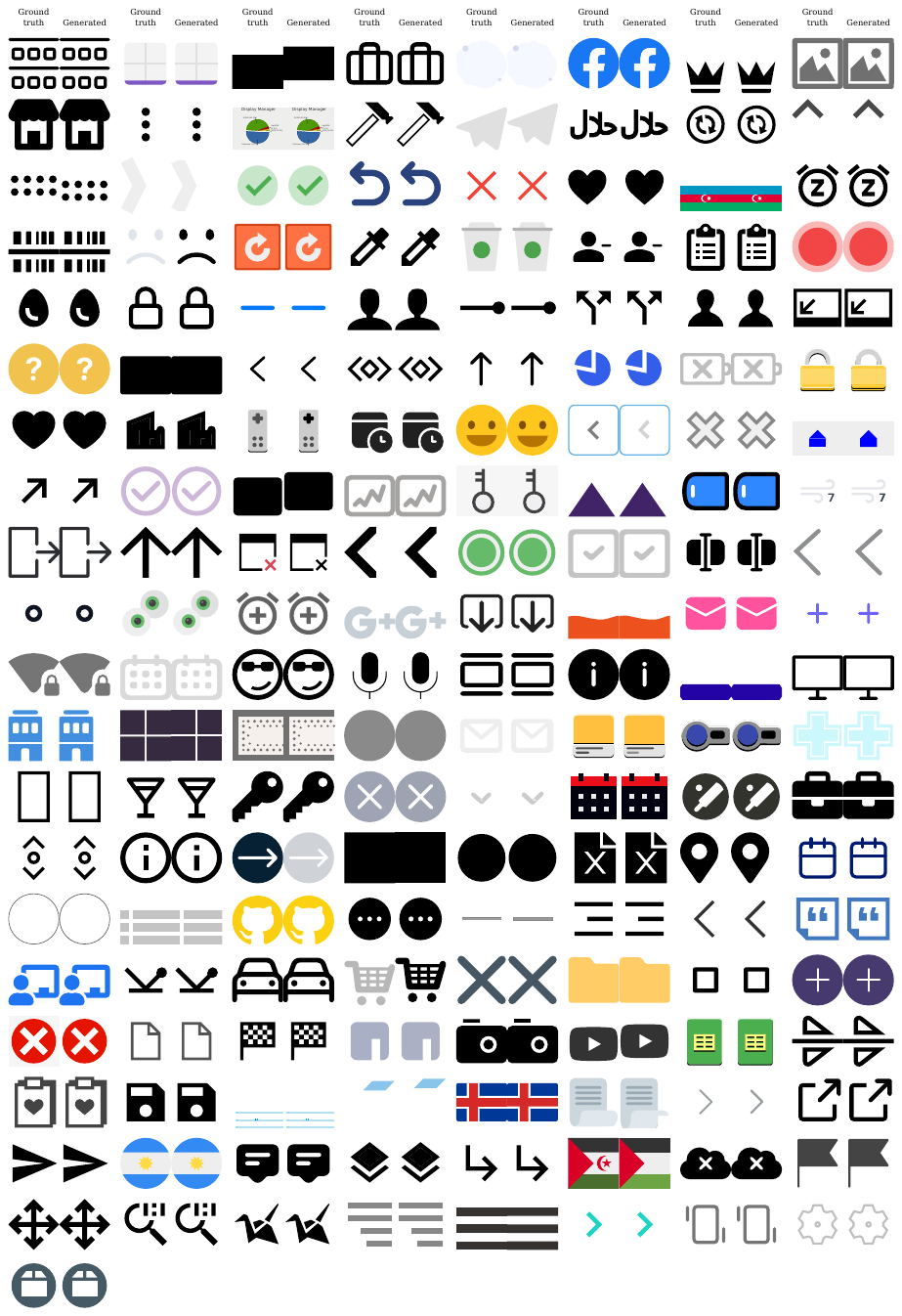}
    \caption{\textbf{Image-to-SVG Results on SVG-Stack}. We present vectorizations of StarVector-1B on the test set of SVG-Stack. Left is input raster image, right is the SVG image (in SVG format).}
    \label{fig:grid-svg-stack-vectorization-sv1}
\end{figure*}

\begin{figure*}[!ht]
    \centering
    \includegraphics[width=0.98\linewidth, trim=0 16.7cm 0 0, clip]{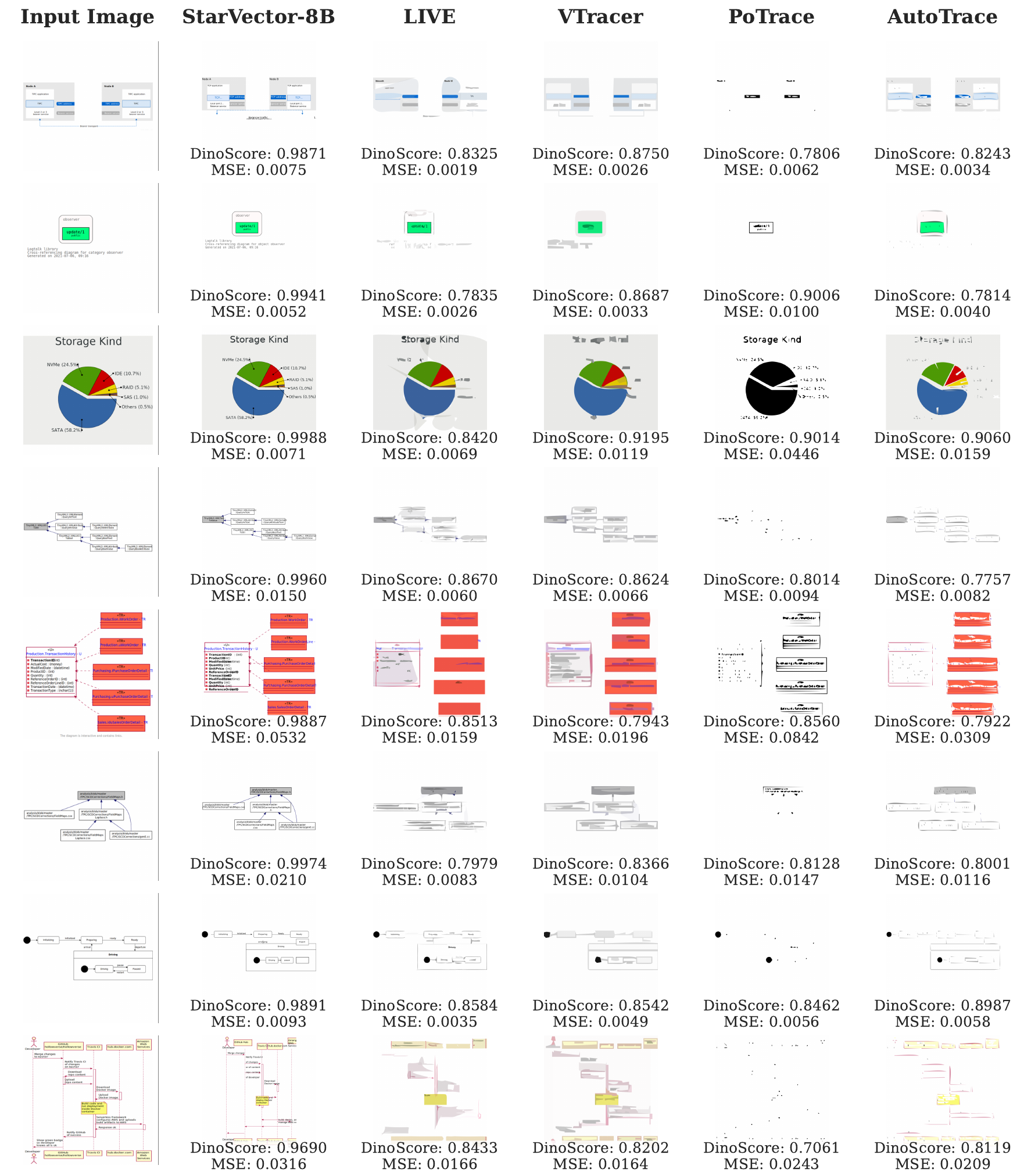}
    \caption{We compare the results from StarVector-8B with those from the most powerful baselines. Notably, StarVector is \textit{the only} method capable of producing acceptable results that preserve both structural integrity and textual content by utilizing a variety of SVG primitives. In contrast, other methods tend to generate blobs and curves that merely attempt to fit the structure and color of the original image. We present two metric scores for each sample: DinoScore and MSE. MSE consistently yields higher scores for other methods, as they focus on fitting vectors to the image as accurately as possible. While StarVector may not achieve perfect reconstruction, its results are preferred for their semantic fidelity. This highlights the limitations of MSE and the importance of DinoScore in capturing these aspects.}\label{fig:svg-diagrams-results}
\end{figure*}

\begin{figure*}[!ht]
    \centering
    \includegraphics[width=0.95\textwidth]{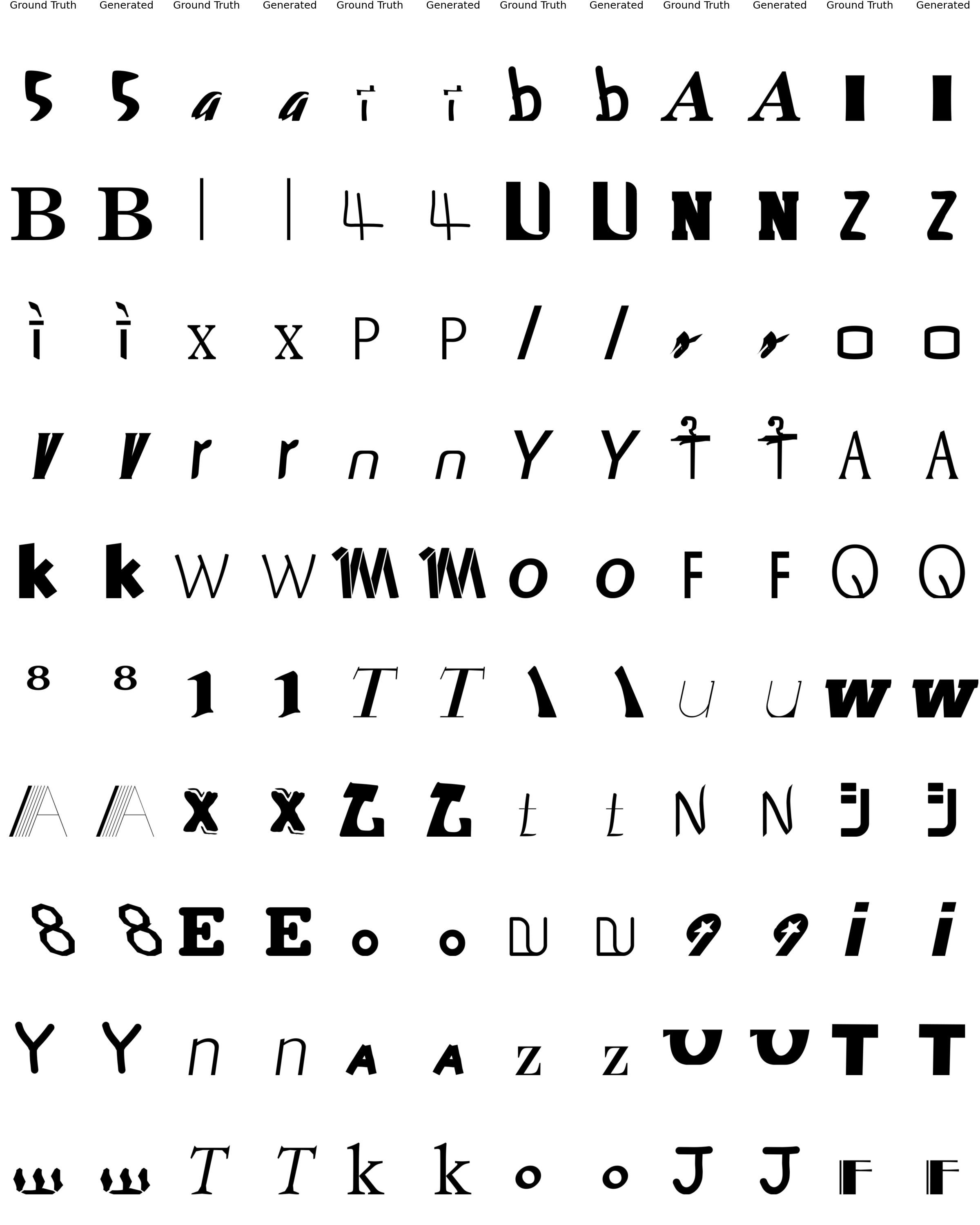}
    \caption{\textbf{Image-to-SVG results on SVG-Fonts} test set. Results are remarkably good, obtaining perfect font reconstructions. Intricate details are preserved. This is because the dataset is very large, above 1M samples. This shows that if having access to a large dataset, StarVector can learn high-quality SVG generation.}
    \label{fig:grid-fonts}
\end{figure*}

\begin{figure*}[!ht]
    \centering
    \includegraphics[width=0.95\textwidth]{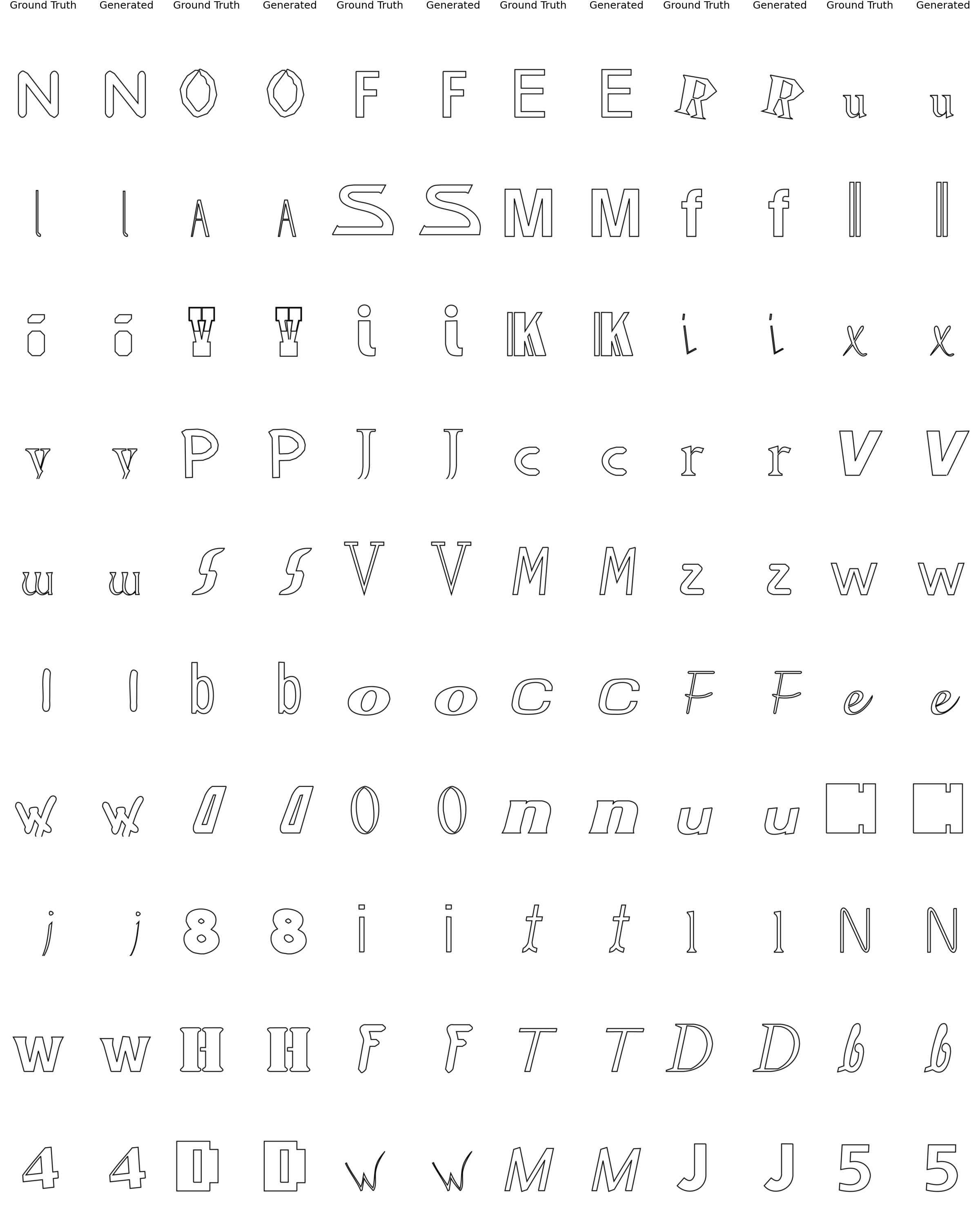}
    \caption{\textbf{Image-to-SVG results on SVG-Fonts simplified} test set.}
    \label{fig:grid-fonts-simplified}
\end{figure*}

\begin{figure*}[!ht]
    \centering
    \includegraphics[width=0.95\textwidth]{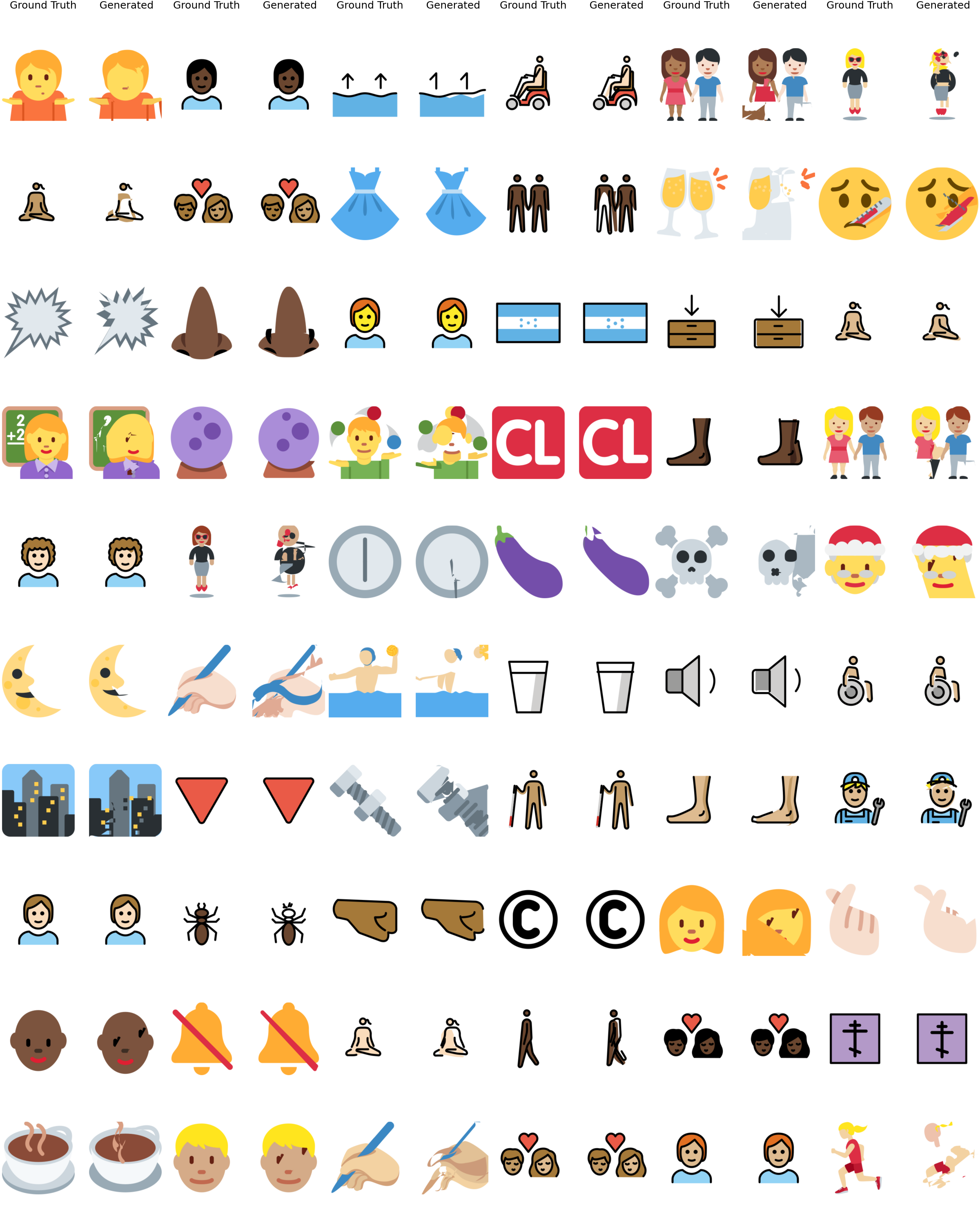}
    \caption{\textbf{Image-to-SVG results on SVG-Emoji} test set. Results are mostly wrong in this benchmark, due to the small training dataset of approximately 8k examples.}
    \label{fig:grid-emoji}
\end{figure*}

\begin{figure*}[!ht]
    \centering
    \includegraphics[width=0.95\textwidth]{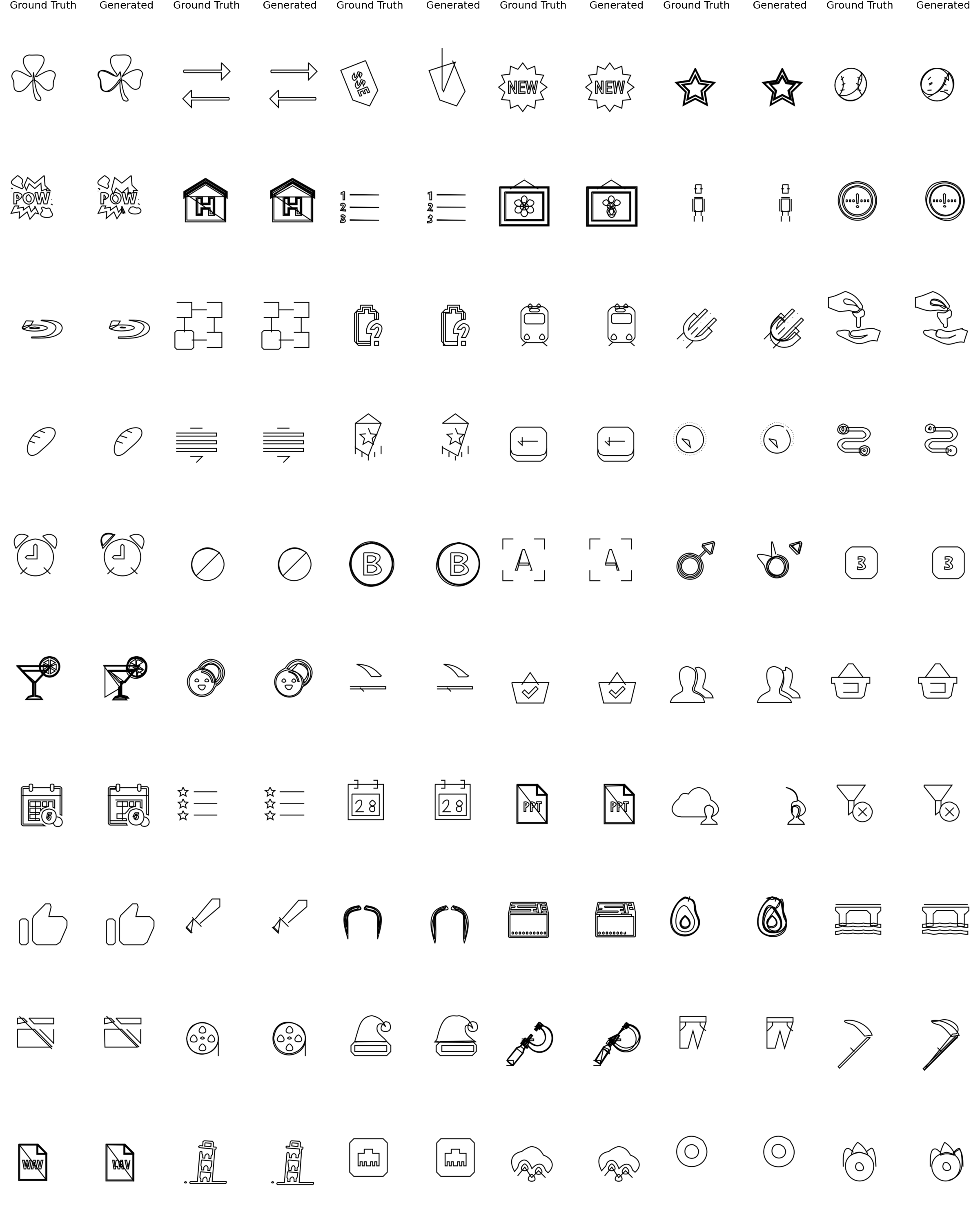}
    \caption{\textbf{Image-to-SVG results on SVG-Icons} test set.}
    \label{fig:grid-icons}
\end{figure*}

\subsubsection{Image-to-SVG Results}\label{app:svg-stack}

We show additional Image-to-SVG results from StarVector. Figures [\ref{fig:grid-svg-stack-vectorization-sv1} - \ref{fig:grid-icons}] show substantial qualitative samples generated by StarVector on all the proposed datasets. All results are computed in the test sets. We can observe the weaknesses and strengths of our model. Simplified datasets (Figures~\ref{fig:grid-fonts-simplified}) are near-perfectly converted to SVG. In the case of icons, in Figure~\ref{fig:grid-icons}, sometimes the model runs out of SVG code tokens, and the image is incomplete. Results on SVG-Emoji~\ref{fig:grid-emoji} show impressive performance in estimating the shape's color and semantics. However, it lacks fine-grained and accurate positioning of objects, i.e., in some examples, the model loses track of the coherent position and form of shapes. These problems result from insufficient emoji samples, i.e., less than 10,000 training examples. This problem can be alleviated by scaling up the current model in the number of parameters (currently 1.4 billion), training data for pre-training, and computing resources. 

\paragraph{Comparing parameter count of models}. The number of learnable parameters in deep learning based models often correlates with performance. Pre-LLM models like Im2Vec and DeepSVG use significantly fewer parameters (up to 5M) compared to StarVector and GPT-based models, which operate in the billions. While pre-LLM models can produce accurate results, they lack the generalization ability of LLM-based approaches. Comparing StarVector with GPT models reveals that high-fidelity SVG generation is achievable with just 1B parameters, whereas GPT-4V lacks specific training for this task. Future models will likely incorporate SVG data in training, but current results already demonstrate that LLM-based approaches offer superior generalization and scalability for SVG generation, at the cost of utilizing more parameters.

\paragraph{Context Length Limitation.}  The model’s architecture imposes a clear limitation on context length, which significantly impacts training and testing data pipelines, as well as the skills the model can learn. Our experiments show that the model scales effectively with increasing context lengths, from 8k to 16k, indicating that this limitation must be addressed with techniques for handling longer contexts—an area LLMs are expected to improve. For fair comparisons, we restricted our benchmark tests to a context length of 8k and evaluated all baselines within this setting. However, the benchmarks also provide versions with longer context lengths to assess future models, as increased length generally correlates with more complex SVGs. We did not observe substantial differences in scores between the 4k and 10k token settings, primarily because the data in our benchmark can typically be represented using an average of 3k tokens.

\paragraph{Limitations on Complex SVG Structures.} StarVector encounters challenges with complex SVG structures, intricate shapes, and detailed illustrations primarily due to limitations in its architecture. Currently, the model’s image encoder handles images by simply padding and resizing them to fixed dimensions of 224 or 384 pixels. This approach may not adequately capture the nuances of complex diagrams. A potential improvement would be to implement a dynamic image processing system akin to those found in newer Vision-Language Models (VLMs), which could enhance the model's ability to interpret and generate intricate SVGs more effectively. Additionally, improving data cleaning processes is crucial, as the model sometimes produces hallucinated information due to noise in the input data, such as URLs or base64-encoded images. Addressing these issues through architectural enhancements and more robust data preprocessing could significantly improve StarVector's performance on complex SVG tasks.

\paragraph{Generalization to Non-Standard SVGs.} StarVector's ability to generalize to non-standard SVGs—those not represented in its main training distribution—poses a significant challenge. While the model performs well on common styles and primitives encountered during training, it struggles with more unique or unconventional SVGs. This is primarily due to the model's training data, which tends to focus on widely used shapes and designs. As a result, StarVector may exhibit a bias towards these common styles, leading to suboptimal performance when faced with SVGs that feature unusual structures or less frequent elements.

To assess StarVector's generalization capabilities, we evaluated its performance on various datasets that include non-standard SVGs. The results indicate that while the model can produce reasonable outputs for some non-standard examples, it often falls short in accurately capturing the intricacies of less familiar styles. This limitation suggests that the model's training set lacks sufficient diversity to encompass the full range of potential SVG designs.

To address these concerns, future work should focus on expanding the training dataset to include a wider variety of SVG styles and structures. Incorporating data from niche applications and artistic domains could enhance the model's ability to generate SVGs across a broader spectrum of design elements. Additionally, techniques such as domain adaptation and transfer learning could be explored to improve generalization to non-standard SVGs, allowing StarVector to adapt more effectively to unfamiliar inputs.

\begin{table*}[!h]
    \centering
    \resizebox{0.98\textwidth}{!}{
    \setlength{\tabcolsep}{3pt} 
    \begin{tabular}{@{}lccc|ccc|ccc|ccc@{}} 
        \toprule
        & \multicolumn{3}{c}{\textbf{SVG-Fonts}} 
        & \multicolumn{3}{c}{\textbf{SVG-Emojis}} 
        & \multicolumn{3}{c}{\textbf{SVG-Icons}} 
        & \multicolumn{3}{c}{\textbf{SVG-Stack}} \\
         \cmidrule(lr){2-4} \cmidrule(lr){5-7} \cmidrule(lr){8-10} \cmidrule(lr){11-13}
        \textbf{Sampling technique}  & \textbf{LPIPS $\downarrow$} & \textbf{SSIM $\uparrow$} & \textbf{MSE $\downarrow$}
         & \textbf{LPIPS} & \textbf{SSIM} & \textbf{MSE}
         & \textbf{LPIPS} & \textbf{SSIM} & \textbf{MSE}
         & \textbf{LPIPS} & \textbf{SSIM} & \textbf{MSE} \\

        \midrule
    
    Greedy  & 0.019 & 0.969 & 0.013  & 0.251 & 0.731 & 0.071  & 0.059 & 0.912 & 0.028  & 0.157 & 0.797 & 0.067\\
    + Beam Search (B=5)  & 0.018 & 0.970 & 0.012  & 0.250 & 0.732 & 0.070  & 0.058 & 0.913 & 0.027  & 0.156 & \textbf{0.798} & \textbf{0.066} \\
    \rowcolor{gray!15}
    Nucleus Sampling (T=0.5)  & \textbf{0.013} & \textbf{0.976} & \textbf{0.008}  & \textbf{0.202} & \textbf{0.778} & \textbf{0.051}  & \textbf{0.043} & \textbf{0.923} & \textbf{0.022}  & \textbf{0.153} & 0.785 & 0.072 \\
        
    Nucleus Sampling (T=1.0)  & 0.015 & 0.975 & 0.009  & 0.244 & 0.742 & 0.067  & 0.053 & 0.917 & 0.025  & 0.161 & 0.786 & 0.069 \\
    
    + Beam-Search (B=5)  & 0.034 & 0.948 & 0.027  & 0.244 & 0.742 & 0.068  & 0.065 & 0.913 & 0.027  & 0.195 & 0.766 & 0.089 \\
    
    + Beam-Search (B=10)  & 0.040 & 0.943 & 0.031  & 0.251 & 0.742 & 0.072  & 0.071 & 0.910 & 0.028  & 0.175 & 0.762 & 0.079 \\
    \hline

    \end{tabular}
    }
    \vspace{5px}
    \caption{\textbf{Ablation study on sampling strategies}. We experimented using greedy decoding and added a beam search with B=5. We test nucleus sampling~\cite{holtzman2019curious} using top p=0.9, with temperatures T=0.5 and T=1.0. The two final rows describe beam search with nucleus sampling at T=1.0. See \url{huggingface.com/blog/how-to-generate} for reference on these sampling techniques.}
    \label{tab:sampling}
    
\end{table*}

\subsection{Ablation Studies}\label{app:ablations}

We performed ablations on the image encoder type, the data augmentation pipeline, inference techniques, and generation parameters. Most of our ablations were performed on the StarVector-1B model for faster iteration, and we empirically find they work well on the larger StarVector-8B.

\begin{figure}[!ht]
    \centering
    \includegraphics[width=1.0\columnwidth]{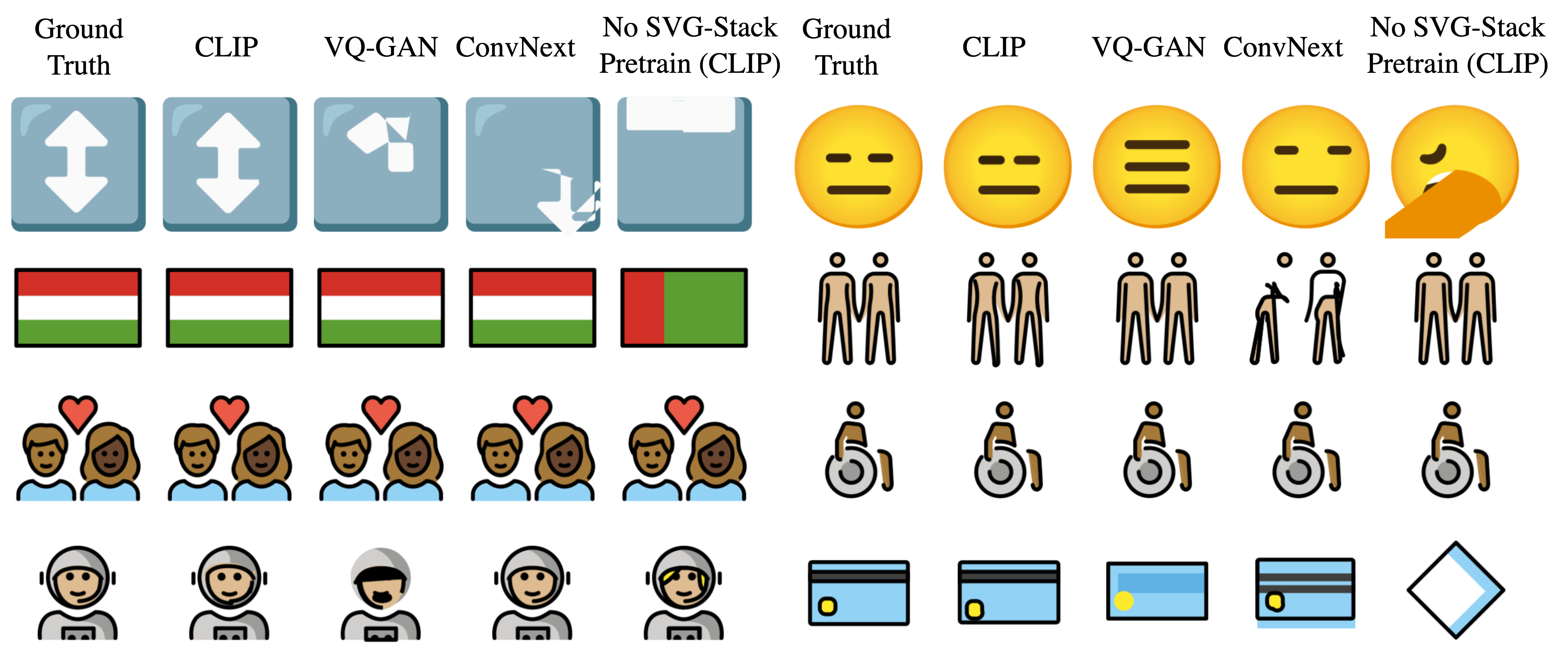}
    \caption{\textbf{Ablation of Image Encoders} Image vectorization results using different visual encoders on SVG-emoji test set. CLIP is the image encoder that delivers the best results, whereas VQ-GAN and ConvNet often miss relevant semantics of the image. No SVG-Stack Pretrain (CLIP) refers to an ablation where we use CLIP out of the box, without unfreezing its weights.}
    \label{fig:ablation-emoji}
\end{figure}

\begin{figure}[!ht]
    \centering
    \includegraphics[width=1.0\columnwidth]{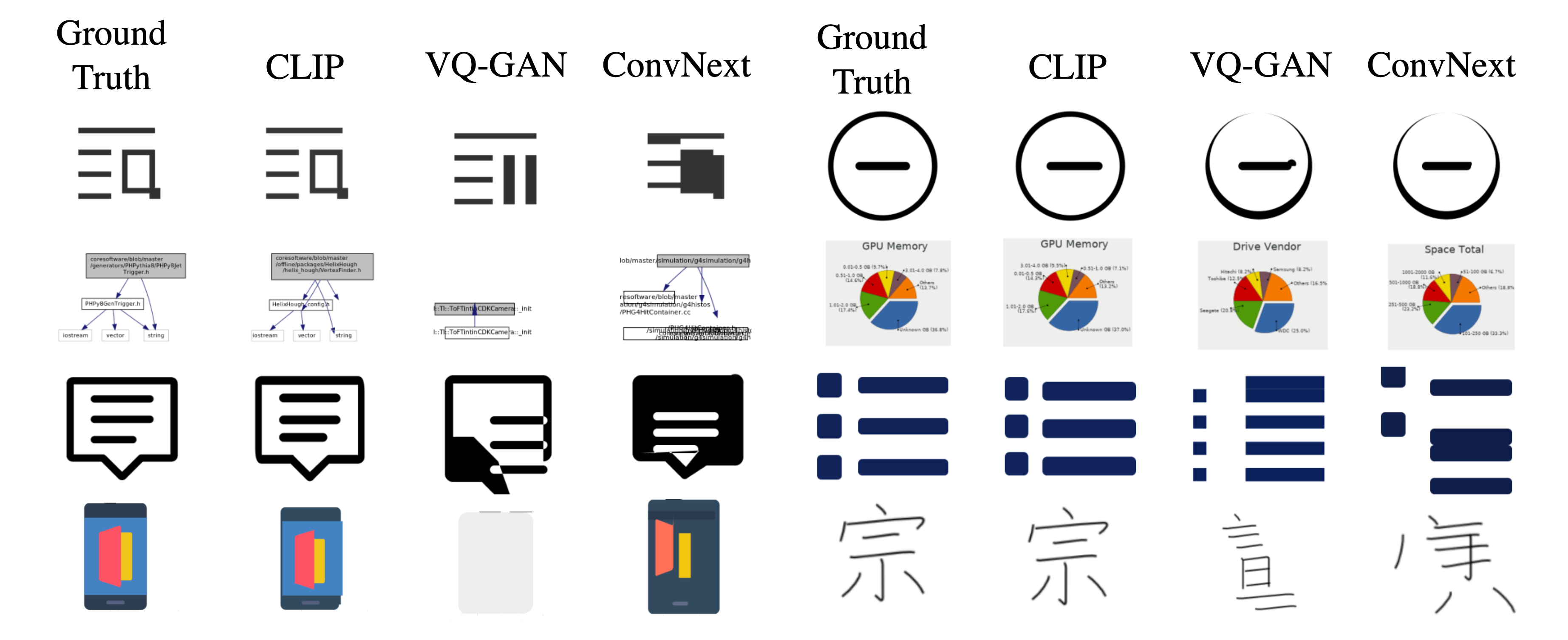}
    \caption{\textbf{Ablation of Image Encoders} Image vectorization results using different visual encoders on SVG-Stack test set. CLIP offers the best results. VQ-GAN and ConvNet often miss relevant semantics of the image.}
    \label{fig:ablation-stack}
\end{figure}

\begin{figure}[!ht]
    \centering
    \includegraphics[width=1.0\columnwidth]{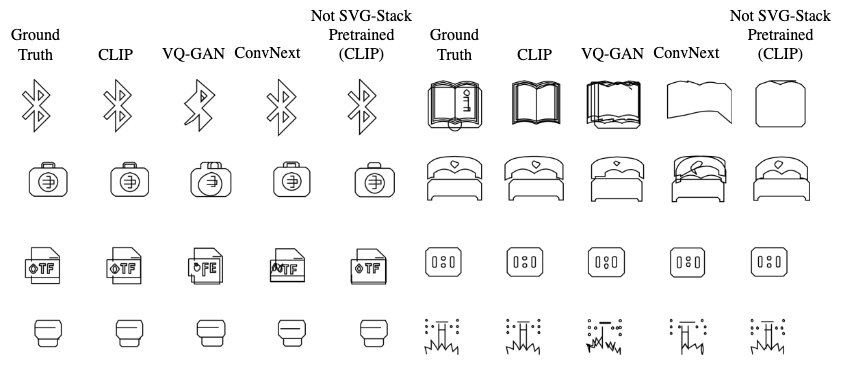}
    \caption{\textbf{Ablation of Image Encoders} Image vectorization results using different visual encoders on SVG-Icons test set. CLIP brings the best visual results. VQ-GAN and ConvNext are not able to capture correctly the details for correct vectorization. No SVG-Stack Pretrain (CLIP) refers to an ablation where we use CLIP out of the box, without unfreezing its weights. Notably, better results are obtained when training the CLIP image encoder on SVG-Stack.}
    \label{fig:ablation-icons}
\end{figure}

\begin{figure}[!ht]
    \centering
    \includegraphics[width=1.0\columnwidth]{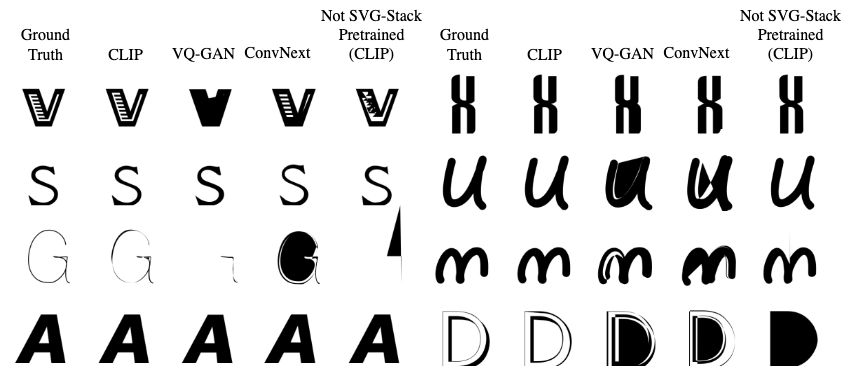}
    \caption{\textbf{Ablation of Image Encoders} Image vectorization results using different visual encoders on SVG-Fonts test set. As in the other datasets tested, CLIP brings the best visual results, as others are not able to provide perfect vector reconstruction when intricate details are present. No SVG-Stack Pretrain (CLIP) refers to an ablation where we use CLIP out of the box, without unfreezing its weights. Notably, better results are obtained when training the CLIP image encoder on SVG-Stack.}
    \label{fig:ablation-fonts}
\end{figure}

\vspace{1cm}
\noindent\textbf{{Image Encoder Ablation.}} The choice of image encoder for the problem of Image-to-SVG is highly impactful, as it determines how well visual information from raster images can be preserved in a representation suitable for precise reconstruction in the SVG space. We ablated the visual encoders by replacing them with VQGAN~\citep{esser2021taming}, ConvNext~\citep{liu2022convnet}, and CLIP ViT-B/32~\citep{radford2021learning}, in our StarVector-1B proposed architecture. This setup evaluates three commonly used approaches in visual representation learning~\citep{radford2021learning, esser2021taming, rombach2022high}. In our experiments, CLIP consistently outperformed across all metrics for various datasets (see Table~\ref{tab:results-ablation-fonts-emoji}). Figures [\ref{fig:ablation-emoji}–\ref{fig:ablation-fonts} further illustrate how VQGAN and ConvNext tend to lose local details during generation, even while maintaining semantic relevance.

\vspace{1cm}\noindent\textbf{{Pre-training on SVG-Stack.}} Pre-training on the SVG-Stack is highly beneficial for the downstream datasets with small data. Table~\ref{tab:augmentations} shows the uplift on all the metrics for different datasets. Qualitatively, we can also see that pre-training helps the model to identify the nuanced details from the images. For the case of SVG-Emoji, pre-training is a vital requirement, as it overfits without it due to limited data. Figure~\ref{fig:grid-emoji} shows that the model relies on colors and shapes to generate the SVG.

\begin{figure*}[!t]
    \centering
    \includegraphics[width=0.98\linewidth]{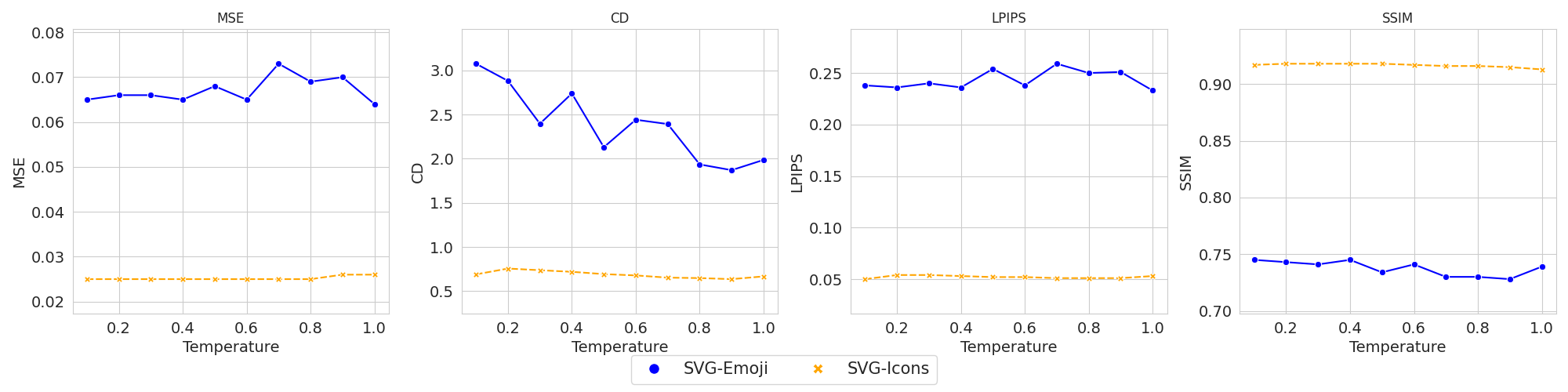}
    \caption{\textbf{Ablation study on sampling temperature}. We tested the performance impact of StarVector-1B when changing the sampling temperature. Results are computed for SVG-Emoji and SVG-Icons validation sets.}\label{temperature}
\end{figure*}

\vspace{1cm}\noindent\textbf{{Ablation on Generation Hyperparameters.}} We explore the impact of different generation hyperparameters on the StarVector-1B model. After an initial exploration to empirically determine the most relevant hyperparameters, we focus our ablation on these. We find that temperature and the number of beams in beam search significantly affect performance. The model is evaluated across various configurations (see Table~\ref{tab:sampling} and Figure~\ref{temperature}). Our results show that a beam search size of 5 achieves the best outcomes, albeit with increased memory usage and runtime. Similarly, nucleus sampling with a top-p of 0.9 and a temperature of 0.5 delivers the best overall performance.

\subsection{Text-to-SVG Results}\label{app:text2svg}

\begin{figure*}
    \centering
    \includegraphics[width=0.98\linewidth]{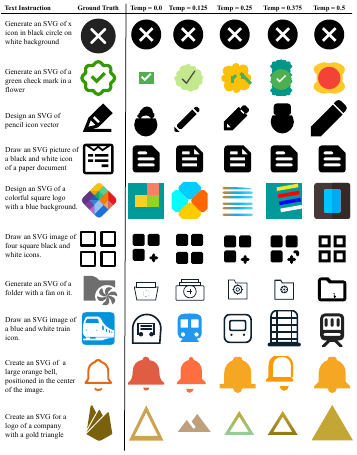}
    \caption{\textbf{Text-to-SVG Results}. We show successful Text-to-SVG results using StarVector-8B. We sample 5 different temperatures as an ablation, showing the sensitivity of this parameter during generation. Results are presented in SVG (not raster images)}
    \label{fig:text2svg-success}
\end{figure*}

\begin{figure*}[!h]
    \centering
    \includegraphics[width=0.98\textwidth, trim=2.5cm 12cm 6.5cm 2.5cm, clip]{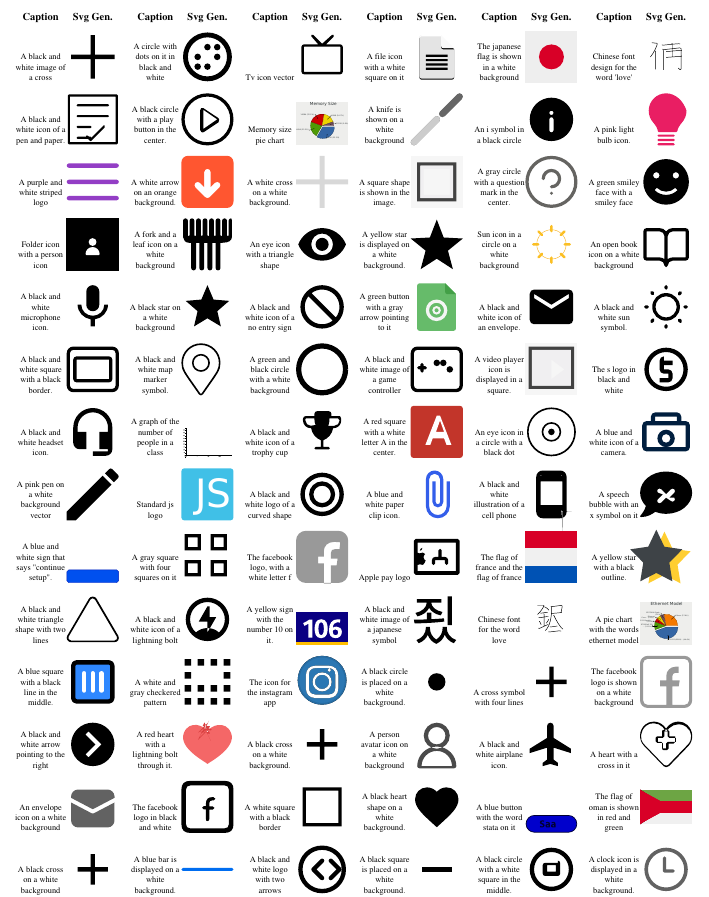}
    \caption{\textbf{Text-to-SVG Generation} results using StarVector-1B on SVG-Stack test set (i).}
    \label{fig:grid-svg-stack-text2svg-sv1-1}
\end{figure*}

\begin{figure*}[!h]
    \includegraphics[width=0.98\textwidth, trim=2.5cm 12cm 6.5cm 2.7cm, clip]{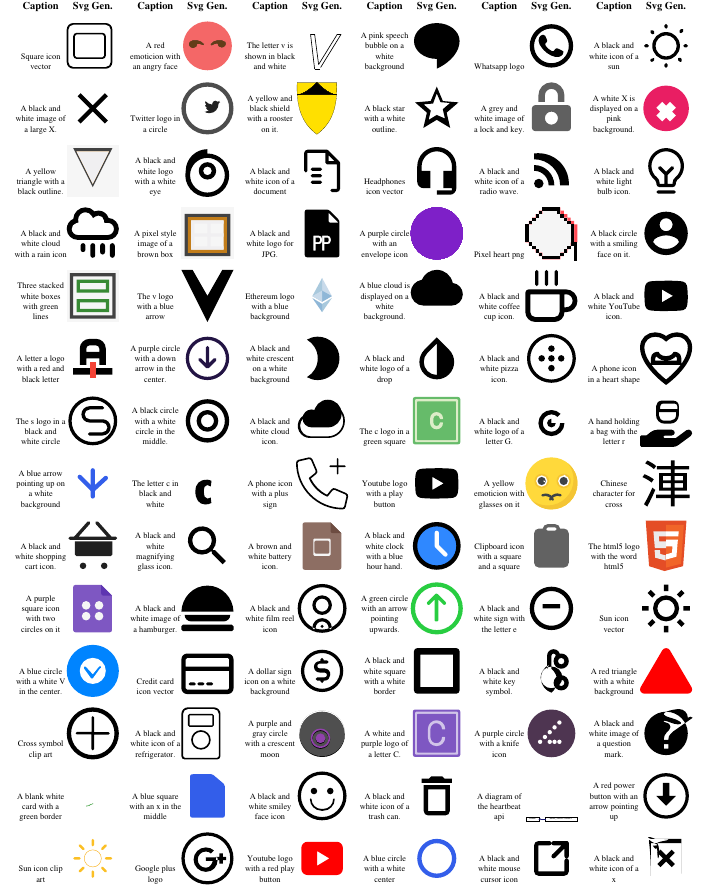}
    \caption{\textbf{Text-to-SVG Generation} results using StarVector-1B on SVG-Stack test set(ii).}
    \label{fig:grid-svg-stack-text2svg-sv1-2}
\end{figure*}

\begin{figure*}[!h]
    \includegraphics[width=0.98\textwidth, trim=2.5cm 12cm 6.5cm 2.7cm, clip]{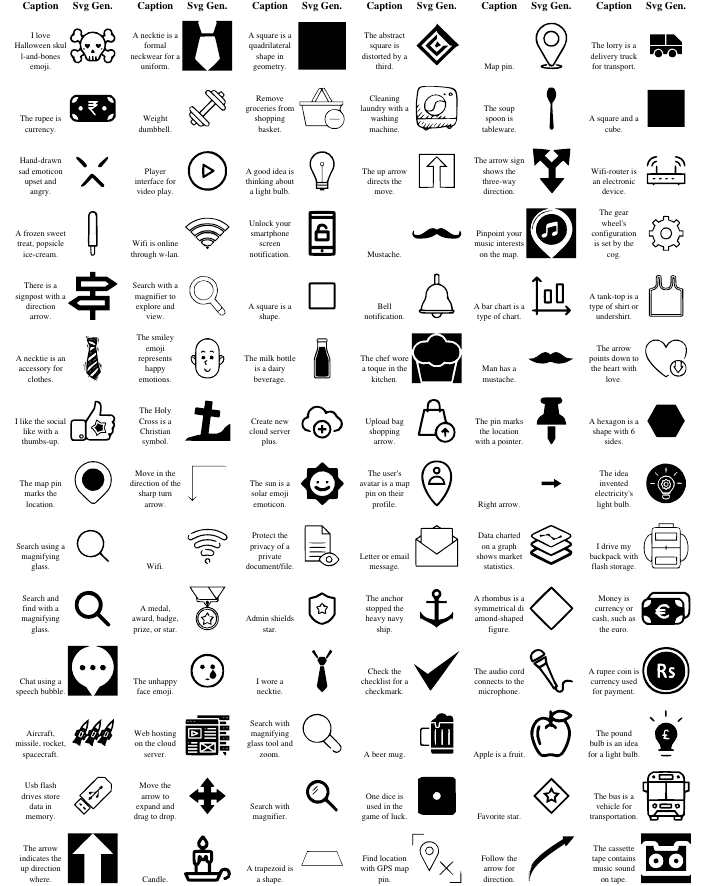}
    \caption{\textbf{Text-to-SVG generation} on FIGR-SVG test set (i).}
    \label{fig:grid-figr-text2svg-1}
\end{figure*}

\begin{figure*}[!h]
    \includegraphics[width=0.98\textwidth, trim=2.5cm 12cm 6.5cm 2.7cm, clip]{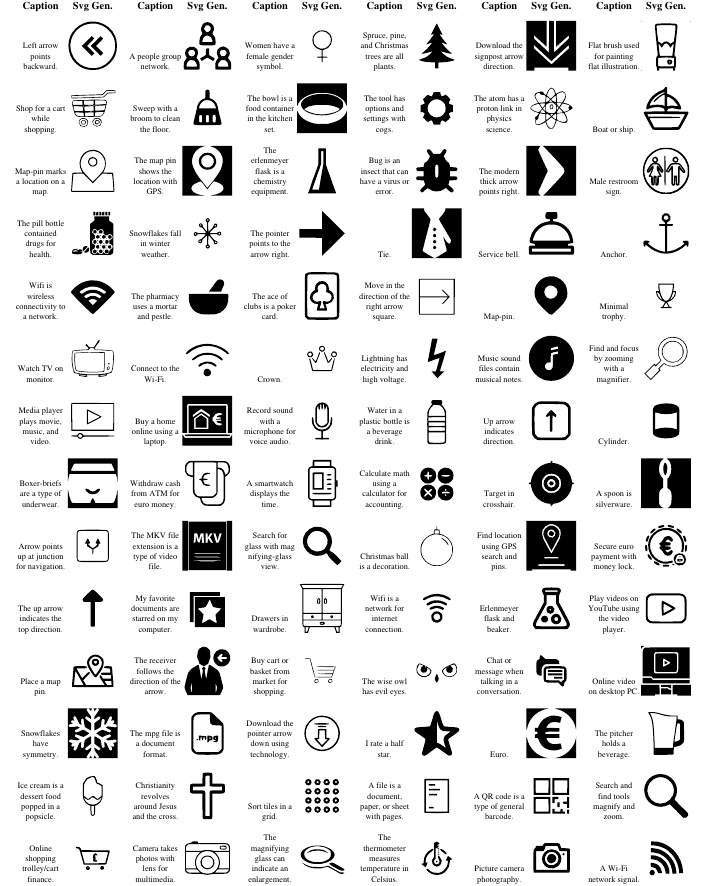}
    \caption{\textbf{Text-to-SVG generation} on FIGR-SVG test set (ii).}
    \label{fig:grid-figr-text2svg-2}
\end{figure*}

\begin{figure*}
    \centering
    \includegraphics[width=0.98\linewidth]{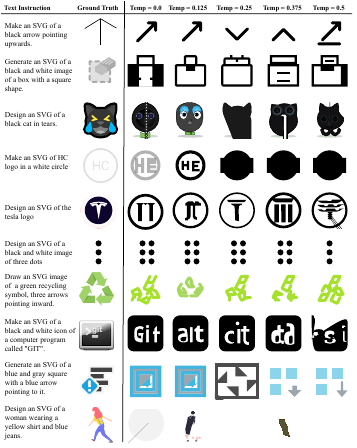}
    \caption{\textbf{Text-to-SVG Results}. We show failure Text-to-SVG results using StarVector-8B (cherry picked examples that show limitations). We sample 5 different temperatures as an ablation, showing the sensitivity of this parameter during generation. Results are presented in SVG (not raster images)}
    \label{fig:text2svg-fail}
\end{figure*}

Figures [\ref{fig:grid-svg-stack-text2svg-sv1-1} - \ref{fig:grid-figr-text2svg-2}] show additional qualitative results of StarVector when performing the task of text-conditioned SVG generation, performed on SVG-Stack and FIGR-SVG test sets. Our samples show reasonable effectiveness at this task, consistently grasping features like colors, shapes, and semantic concepts. However, sometimes some details required in the prompt are lost, e.g., an exact number of circles, shapes inside other shapes, or the direction of arrows. In some cases, some vector graphics shapes lose coherence, which we attribute to our model's current scale in terms of model parameters and context length. We suspect that these mistakes are due to the limited quality of the textual descriptions, sometimes lacking precision and grounding on the SVG images. See Figure~\ref{fig:text2svg-success} for successful cases of Text-to-SVG generation on SVG-Stack. Figure~\ref{fig:text2svg-fail} highlights some failure modes of StarVector-8B. These figures illustrate the impact of different generation temperatures. We rank the outputs generated at different temperatures based on their CLIP Score in relation to the text instruction.

Nevertheless, the StarVector approach of using LLMs for SVG code generation is the only method among baselines that allows us to create diverse vector graphics unrestrainedly, paving the way for more challenging and intricate designs.

\begin{table}[!ht]
    \centering
    \small 
        \caption{\textbf{Results of SVG Data Augmentation.} We ablate both our data augmentation pipeline and the use of a pretraining stage with SVG-Stack. These experiments are conducted on smaller datasets that are more susceptible to overfitting, using the StarVector-1B model. Vanilla refers to the StarVector model trained directly on the given dataset without SVG-Stack pretraining. Next, we introduce our data augmentation pipeline. Finally, we initialize training from an SVG-Stack pretrained checkpoint and fine-tune on the given dataset. The “+” symbol indicates that the methods from the previous rows are also included.}

    \resizebox{1.0\linewidth}{!}{
    \setlength{\tabcolsep}{3pt} 
    \begin{tabular}{@{}lccc|ccc@{}} 
        \toprule
        & \multicolumn{3}{c}{\textbf{SVG-Emojis}} 
        & \multicolumn{3}{c}{\textbf{SVG-Icons}} \\
         \cmidrule(lr){2-4} \cmidrule(lr){5-7} 
        \textbf{Method}  & \textbf{LPIPS $\downarrow$} & \textbf{SSIM $\uparrow$} & \textbf{MSE $\downarrow$}
         & \textbf{LPIPS$\downarrow$} & \textbf{SSIM$\uparrow$} & \textbf{MSE$\downarrow$} \\
        \midrule
        StarVector (vanilla) & 0.355 & 0.683 & 0.108  & 0.104 & 0.845 & 0.047 \\
        + Data Augmentation  & \underline{0.329} & \underline{0.706} & \underline{0.097}  & \underline{0.057} & \underline{0.905} & \underline{0.029} \\
        \rowcolor{gray!15}
        + SVG-Stack Pretrain  & \textbf{0.225} & \textbf{0.748} & \textbf{0.061} & \textbf{0.057} & \textbf{0.894} & \textbf{0.031} \\
        \bottomrule
    \end{tabular}
    }
    
    \label{tab:augmentations}
\end{table}

\begin{table}[!ht]
    \centering
        \caption{\textbf{Ablation of Image Encoders.} We ablate different image encoders with StarVector-1B, namely CLIP ViT-B/32~\citep{radford2021learning}, VQ-GAN~\citep{esser2021taming}, and ConvNext~\citep{liu2022convnet}. We experiment with training experiments on SVG-Fonts and SVG-Emojis datasets. CLIP gives the best results on all reconstruction metrics.}
    \resizebox{1.0\linewidth}{!}{
    \setlength{\tabcolsep}{3pt} 
    \begin{tabular}{@{}lccc|ccc@{}} 
        \toprule
        & \multicolumn{3}{c}{\textbf{SVG-Fonts}} 
        & \multicolumn{3}{c}{\textbf{SVG-Emojis}} \\
         \cmidrule(lr){2-4} \cmidrule(lr){5-7}
        \textbf{Encoder}  & \textbf{LPIPS $\downarrow$} & \textbf{SSIM $\uparrow$} & \textbf{MSE $\downarrow$}
         & \textbf{LPIPS $\downarrow$} & \textbf{SSIM $\uparrow$} & \textbf{MSE $\downarrow$} \\
        \midrule
        \rowcolor{gray!15}
        CLIP  & \textbf{0.026} & \textbf{0.955} & \textbf{0.021} & \textbf{0.202} & \textbf{0.778} & \textbf{0.051} \\
        VQGAN & 0.092 & 0.854 & 0.072 & 0.345 & 0.688 & 0.099 \\
        ConvNext & 0.085 & 0.854 & 0.073 & 0.311 & 0.708 & 0.088 \\
        
        \bottomrule
    \end{tabular}
    }
    \label{tab:results-ablation-fonts-emoji}
\end{table}

\subsection{Results on SVG-Diagrams}
Figure~\ref{tab:svg-diagrams} presents the results of StarVector-8B, along with comparisons to LIVE, VTracer, Potrace, and AutoTrace. StarVector-8B is the only approach that produces plausible results, as it effectively leverages appropriate SVG primitives. DinoScore aligns well with this visual assessment, accurately reflecting the quality of StarVector-8B's outputs.

In contrast, MSE consistently favors other baselines despite their limitations. This is because MSE prioritizes exact pixel matching, favoring models designed to fit curves and colors to the input image. However, these baselines fail to preserve the semantics of the original diagrams, resulting in outputs where the meaning and structure are completely lost.

\subsection{Analysis of SVG Primitives}\label{app:app-primitives}

This section examines how StarVector leverages SVG primitives to produce more compact and semantically accurate SVGs. In contrast to prior models constrained to using only \textit{path} primitives, StarVector effectively utilizes the entire range of SVG primitives, including parametrically defined shapes, gradients, and text elements.

This enhanced capability stems from its ability to operate directly within the SVG code space, facilitated by its multimodal, transformer-based architecture~\citep{alayrac2022flamingo, liu2023visual}, which integrates visual and textual inputs. StarVector generates SVG code that closely resembles the input raster image while maintaining semantic awareness, enabling the use of symmetry, parametric shapes, and text. Prior methods, limited to first-order \textit{path} primitives, lack this semantic understanding, resulting in less compact and less expressive SVG representations.

\definecolor{tagcolor}{rgb}{0.75, 0.13, 0.13}
\definecolor{attrcolor}{rgb}{0.25, 0.5, 0.9}
\definecolor{valcolor}{rgb}{0.12, 0.55, 0.12}

\lstset{
    language=XML,
    basicstyle=\ttfamily\tiny, 
    keywordstyle=\color{tagcolor}\bfseries, 
    identifierstyle=\color{attrcolor},
    stringstyle=\color{valcolor},
    morekeywords={svg, rect, polygon, ellipse, circle, line, path}, 
    breaklines=true,
    breakatwhitespace=true, 
    postbreak=\mbox{}, 
    frame=none, 
    columns=fullflexible
}

\begin{table*}[h] 
    \centering
        \caption{\textbf{Usage of SVG Primitives.} Image vectorization results of StarVector and VTracer applied to images containing basic shapes, such as circles, rectangles, and polygons, with varied colors and transparencies. The leftmost column shows the input images prompted for vectorization, and other columns show the output SVG code, with the SVG primitives in red color. StarVector accurately identifies and generates SVG code for each primitive, preserving their distinct characteristics. In contrast, VTracer relies on the \texttt{path} primitive, resulting in SVG code that captures the input image in terms of pixels, with less fidelity to individual shapes. Due to the length of VTracer's SVG output, only the initial lines are shown. VTracer serves as a baseline model, representative of other baselines, which are omitted for space but exhibit similar behavior, primarily using \texttt{path} without shape recognition.}
    \label{fig:primitives}
    \begin{tabular}{@{}cc@{}c@{}} 

    \textbf{Test Example} & \textbf{StarVector} & \textbf{VTracer} \\
        
    \hline  
    
    \begin{minipage}{0.1\textwidth}
    \includegraphics[width=2cm]{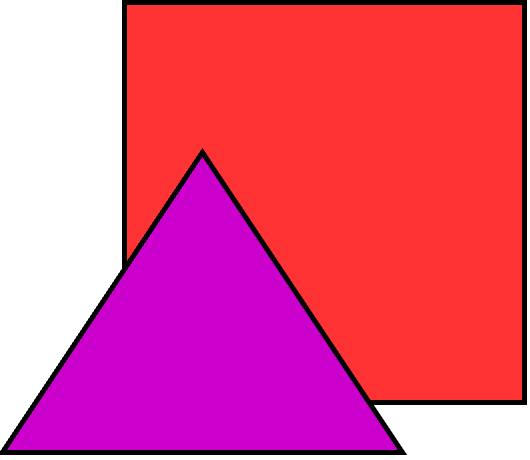}
    \end{minipage}
    &
    \begin{minipage}{0.4\textwidth} 
        \lstinputlisting[firstline=1, lastline=10]{figures/results/primitives-analysis/svg/sample1_starvector.svg}
    \end{minipage}
        
    &
    \begin{minipage}{0.45\textwidth} 
        \lstinputlisting[firstline=1, lastline=10]{figures/results/primitives-analysis/svg/sample1_vtracer.svg} 
    \end{minipage}
        
    \\

    \midrule
    
    \begin{minipage}{0.1\textwidth} 
        \includegraphics[width=2cm]{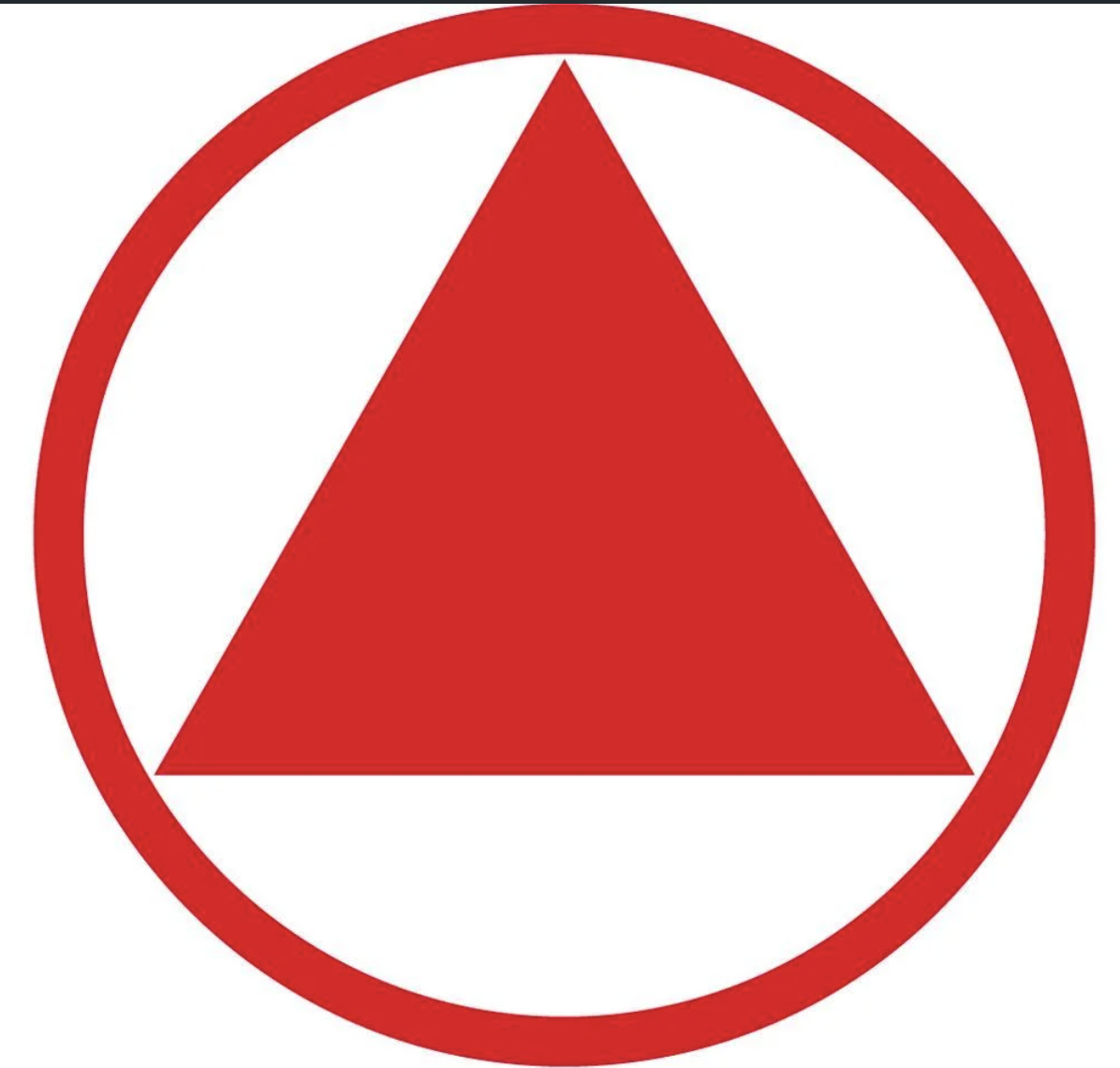}
        \end{minipage}
    &
    \begin{minipage}{0.4\textwidth} 
        \lstinputlisting[firstline=1, lastline=10]{figures/results/primitives-analysis/svg/sample2_starvector.svg}
    \end{minipage}
    &
    \begin{minipage}{0.45\textwidth} 
        \lstinputlisting[firstline=1, lastline=10]{figures/results/primitives-analysis/svg/sample2_vtracer.svg}
    \end{minipage}\\

     \midrule
         
    \begin{minipage}{0.1\textwidth} 
    \includegraphics[width=2cm]{figures/results/primitives-analysis/images/sample3.png}
    \end{minipage}
    &
    \begin{minipage}{0.4\textwidth} 
        \lstinputlisting[firstline=1, lastline=10]{figures/results/primitives-analysis/svg/sample3_starvector.svg}
    \end{minipage}
    &
    \begin{minipage}{0.45 \textwidth} 
        \lstinputlisting[firstline=1, lastline=10]{figures/results/primitives-analysis/svg/sample3_vtracer.svg}
    \end{minipage}
    
    \\

    \midrule
        
    \begin{minipage}{0.1\textwidth} 
    \includegraphics[width=2cm]{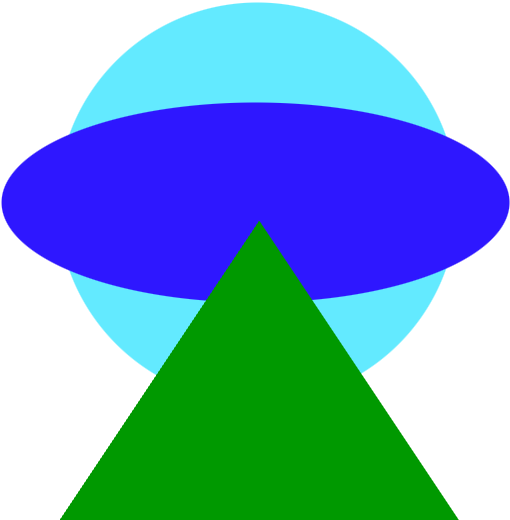}
    \end{minipage}
    &
    \begin{minipage}{0.4\textwidth} 
        \lstinputlisting[firstline=1, lastline=10]{figures/results/primitives-analysis/svg/sample4_starvector.svg}
    \end{minipage}
            &
    \begin{minipage}{0.45\textwidth} 
        \lstinputlisting[firstline=1, lastline=10]{figures/results/primitives-analysis/svg/sample4_vtracer.svg}
    \end{minipage}\\

    \end{tabular}

\end{table*}

\begin{figure*}[!h]
    \resizebox{0.99\textwidth}{!}{
    \includegraphics{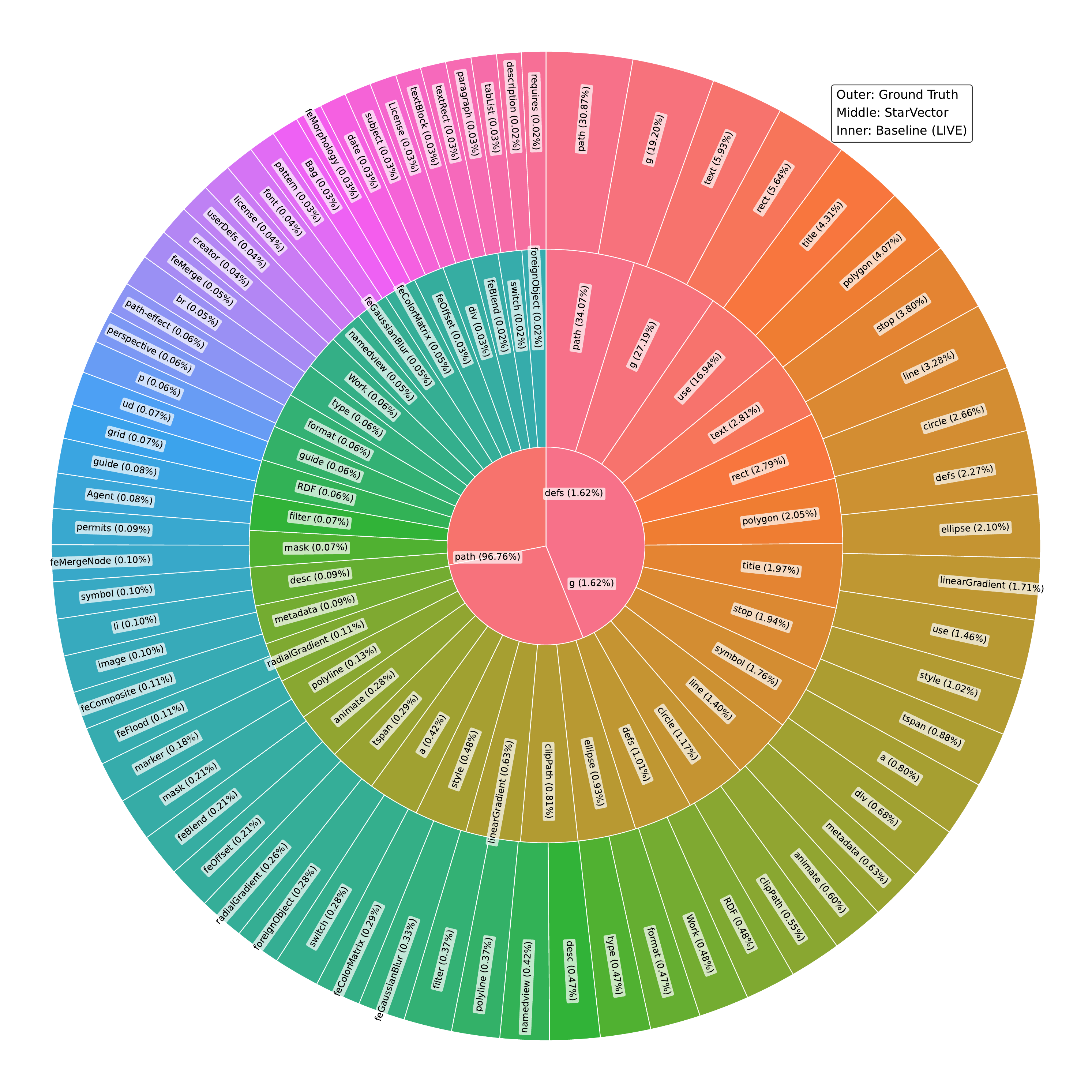}
    }
    \caption{\textbf{Distribution of SVG Primitives and SVG Tags.} We show the frequency of SVG tags that appear in the ground truth of SVG-Stack dataset (outer ring), compared to the frequency of tags generated by StarVector (middle ring), and compared to our most performant baseline (LIVE). For visualization purposes, we apply a logarithmic scaling over the counts and show the base percentage in parentheses. StarVector generates SVGs that contain tags with a similar distribution to the ground truth. The baseline is limited to \texttt{paths} or basic primitives.}
\label{fig:tag-stats}
\end{figure*}

\paragraph{Qualitative Analysis of SVG Primitives}\label{app:qualitative-primitives}

Table~\ref{fig:primitives} presents tests conducted on StarVector-8B and VTracer using simple designs composed of basic shapes such as circles, rectangles, and triangles, with variations in color, transparency, and levels of overlap. StarVector-8B demonstrates the ability to precisely identify the primitives that make up each design, producing visually accurate results while maintaining compact, concise, and interpretable SVG representations.

\paragraph{SVG Tag Distribution}\label{app:tag-stats}
Here, we show how StarVector can generate complex SVGs using the full syntax of the SVG language, in contrast with most of the literature methods, which are restricted to using only the \textit{path} command. Figure~\ref{fig:tag-stats} displays the distribution of SVG tags in the SVGs generated by StarVector along with the distribution in the SVG-Stack dataset, showcasing the strength of our method in using SVG tags and syntax in a similar way to the original in-the-wild dataset. We have computed the exact statistics on previous methods and found that they cannot come close to StarVector in this metric, as they are limited to using \textit{paths} and basic primitives. The effective usage of the large array of SVG tags and syntax makes our method the first SVG model to support these complexities.

\subsection{Human Evaluation}
We conducted a human evaluation to compare the outputs of StarVector-8B, our best model, with those of the most powerful baselines. Participants were selected from diverse backgrounds and carefully screened for conflicts of interest, with none of the key authors involved. The evaluation was performed through a web interface (shown in Figure~\ref{fig:osiris}) that provided anonymized outputs and randomized sample presentations to prevent pattern recognition or bias.

The results, presented in Figure~\ref{fig:human-eval}, demonstrate a strong preference for StarVector-8B across all settings, especially in SVG-Diagrams tasks. This highlights a disconnect between pixel-based metrics (e.g., MSE, SSIM) and human visual perception of SVGs. While baseline models often prioritize pixel-perfect reconstruction, human evaluators preferred StarVector's sharp, well-defined shapes and its effective use of primitives (Figure~\ref{fig:tag-stats}).

Spearman correlation analyses between model metrics and human evaluations further emphasize this gap. MSE shows weak correlations (0.0596 and -0.1002), indicating its inadequacy as a predictor of human preferences. In contrast, DinoScore exhibits significantly stronger correlations, with values of -0.6193 and 0.6214. Moreover, a robust correlation of 0.7577 between differences in DinoScore and human evaluation scores highlights DinoScore as a more reliable metric for assessing SVG quality in alignment with human judgment. 

\subsection{Comparing StarVector with Baselines}

Here, we discuss the results of each baseline individually and compare them to our proposed approach.

\vspace{1cm}
\begin{enumerate}
    \item \textbf{DeepSVG}~\cite{carlier2020deepsvg} is an elegant approach to learning a latent variable model for SVG. It proves effective at learning the task for the \textit{simplified} datasets (Figure~\ref{fig:results-icons-simplified}). It can accurately represent corners and edges. However, it only works in simplified datasets. This limitation restricts it from being a suitable solution in real applications.
    
    \vspace{1cm}
    \item \textbf{Im2Vec}~\cite{reddy2021im2vec} proposes a training procedure that does not require having SVG ground truth. It uses only pixel-based reconstruction loss with the input image, finding the optimal SVG parameters using a differentiable rasterizer like DiffVG~\cite{li2020differentiable}. This framework is appealing, as it aims to be used in images without SVG supervision. However, it requires hundreds of epochs with a reduced dataset to overfit the model to those examples, only working on modeling \textit{training} examples, as seen in \cite{reddy2021im2vec}. Im2Vec results on the datasets presented SVG-Bench are quite poor as it has bad generalization. Therefore, qualitative samples are not presented.  
    
    \vspace{1cm}
    \item \textbf{GPT-4 Vision}~\cite{openai2023gpt4v} excels at capturing the semantics of images and generating captions that accurately describe them. The SVG generated from this description is valid and effectively incorporates semantic concepts along with the accurate colors of the input image into the SVG code (see Figures [\ref{fig:results-icons-simplified} - \ref{fig:failure-emoji}]). However, it falls short in terms of reconstruction fidelity, as GPT-4V was not specifically trained for reconstruction tasks, making these results predictable.

    \vspace{1cm}
    \item \textbf{LIVE} achieves the best results in terms of pixel-based metrics like MSE, LPIPS, and SSIM (see Tables [\ref{tab:consolidated-image-vectorization-results}, \ref{tab:svg-diagrams}, \ref{tab:paths-and-inference}]). Using 32 paths, it effectively represents a wide range of images, making it highly versatile for vectorization tasks, including natural images, which StarVector cannot handle due to its specialized training. However, LIVE has notable limitations. It relies on a slow test-time optimization process (approximately 10 minutes per sample, using 32 paths, Table~\ref{tab:paths-and-inference}) to refine SVG outputs for good MSE scores, often introducing unwanted visual artifacts. Additionally, its exclusive use of \textit{path} primitives results in significantly larger SVG files (19k tokens), as shown in Table\ref{tab:consolidated-image-vectorization-results}. In contrast, StarVector produces compact (3k tokens) and professional-grade SVGs by leveraging a variety of SVG primitives beyond paths, achieving higher precision and efficiency.

    Notably, we find that this method, and the broader family of differentiable techniques it belongs to, is unsuitable for generating images that require specific primitives, such as those in diagrams. Its performance on the SVG-Diagrams benchmark is poor, as illustrated in Figure~\ref{fig:svg-diagrams-results}.
    
    \vspace{1cm}
    \item \textbf{DiffVG} offers comparable results as the other baselines in terms of reconstruction metrics, and it can produce suitable SVG image vectorization with 32 paths (same as LIVE), as seen in Tables [\ref{tab:consolidated-image-vectorization-results}, \ref{tab:svg-diagrams}, \ref{tab:paths-and-inference}]. This tables also show that DiffVG produces large SVGs as seen in the number of tokens (Tokens column), approaximately 19k tokens, similar to LIVE. This means that files are extremely large compared to the ones of StarVector. As mentioned before, this is due to StarVector leveraging understanding and SVG primitieves. Nevertheless, this method is substantially faster than LIVE, requiring 30 seconds per sample. 
    
    \vspace{1cm}
    
    \item \textbf{Image Processing Methods: VTracer, Potrace, Autotrace}
    \textit{Previous image processing methods for Image-to-SVG are powerful}, excelling at fitting vector images to raster inputs with near-zero MSE while reliably capturing the shapes and colors of the original image. However, \textbf{we identify several shared limitations across these methods:} (1) lack of SVG file compression, as they often generate excessively long paths (see the Tokens column in Tables~\ref{tab:consolidated-image-vectorization-results} and \ref{tab:svg-diagrams}); (2) susceptibility to visual artifacts, especially with challenging patterns; and (3) poor performance in vectorizing diagrams, as illustrated in Figure~\ref{fig:svg-diagrams-results}. On the SVG-Diagrams benchmark, \textbf{VTracer}, \textbf{Potrace}, and \textbf{AutoTrace} struggle with producing high-quality results.
    
    These methods perform best on images that can be segmented into distinct regions by color or texture but fail with complex patterns, such as small, closely spaced shapes or fine details. For example, in Figures \ref{fig:results-icons-simplified} and \ref{fig:svg-stack-comparison}, small polygons and intricate text are inadequately vectorized, with details often lost. All three methods are restricted to \textit{path} primitives, limiting their ability to support features like optical character recognition or rendering text with the \texttt{<text>} tag. Although \textbf{Potrace} sometimes better preserves text compared to others, it still cannot recognize or encode it semantically.
    
    Despite these limitations, these methods excel in generation speed, making them highly efficient for many tasks. As shown in Table~\ref{tab:paths-and-inference}, \textbf{VTracer} and \textbf{AutoTrace} can generate SVGs in under a second, while \textbf{Potrace} typically takes around 10 seconds. In comparison, StarVector requires over a minute, and other baselines can take anywhere from 10 to 20 minutes. While StarVector’s semantic richness and compact outputs make it better suited for certain applications, these image processing methods demonstrate a clear advantage in scenarios where speed is critical.
        
    \vspace{1cm}
    \item \textbf{IconShop} achieves remarkable results on the SVG-FIGR dataset, as demonstrated in Table~\ref{tab:merged_results_text2svg} and Figures of their original work~\citep{wu2023iconshop}. However, it is not designed to handle SVG-Stack due to its restriction to modeling only the \textit{path} primitive. StarVector outperforms IconShop, as shown in Tables~\ref{tab:merged_results_text2svg}, across metrics such as FID, FID CLIP, and CLIP Score. Qualitative examples further highlight StarVector's superior performance in Text-to-SVG generation within the FIGR-SVG dataset.

    \vspace{1cm}
    \item \textbf{LLMs for Code Generation.} \textit{Methods utilizing LLMs to directly generate SVG code present appealing advantages.} In our evaluation, we assessed GPT-4, GPT-4V, CodeLlama, and our proposed StarVector approach. By leveraging the code space, these models can utilize various SVG primitives based on their understanding of raster images, including path shapes and higher-order primitives like circles and text. This capability enables applications in new domains, such as diagram generation, as demonstrated by the results in Figure~\ref{fig:svg-diagrams-results}. However, most LLMs have not been specifically trained for the Image-to-SVG task, which limits their performance. In contrast, StarVector outperforms other LLMs due to its dedicated architecture and tailored training methodology, excelling in both image understanding and SVG generation.

Upon reviewing the complete set of results, we conclude that the StarVector approach is the only deep learning-based Image-to-SVG model capable of achieving results comparable to those of VTracer, Potrace, and AutoTrace. Furthermore, StarVector paves the way for novel research avenues in vector graphics generation, enabling applications such as diagram generation, Text-to-SVG generation, and potentially enhanced editing and understanding of vector images.

\subsection{Human Evaluation}

We conducted a human evaluation to compare the outputs of StarVector-8B, our best model, with those of the most powerful baselines. Participants were selected from diverse backgrounds and carefully screened for conflicts of interest, with none of the key authors involved. The evaluation was performed through a web interface (shown in Figure~\ref{fig:osiris}) that provided anonymized outputs and randomized sample presentations to prevent pattern recognition or bias.

\begin{figure*}[!ht]
    \centering
    \includegraphics[width=1.0\linewidth]{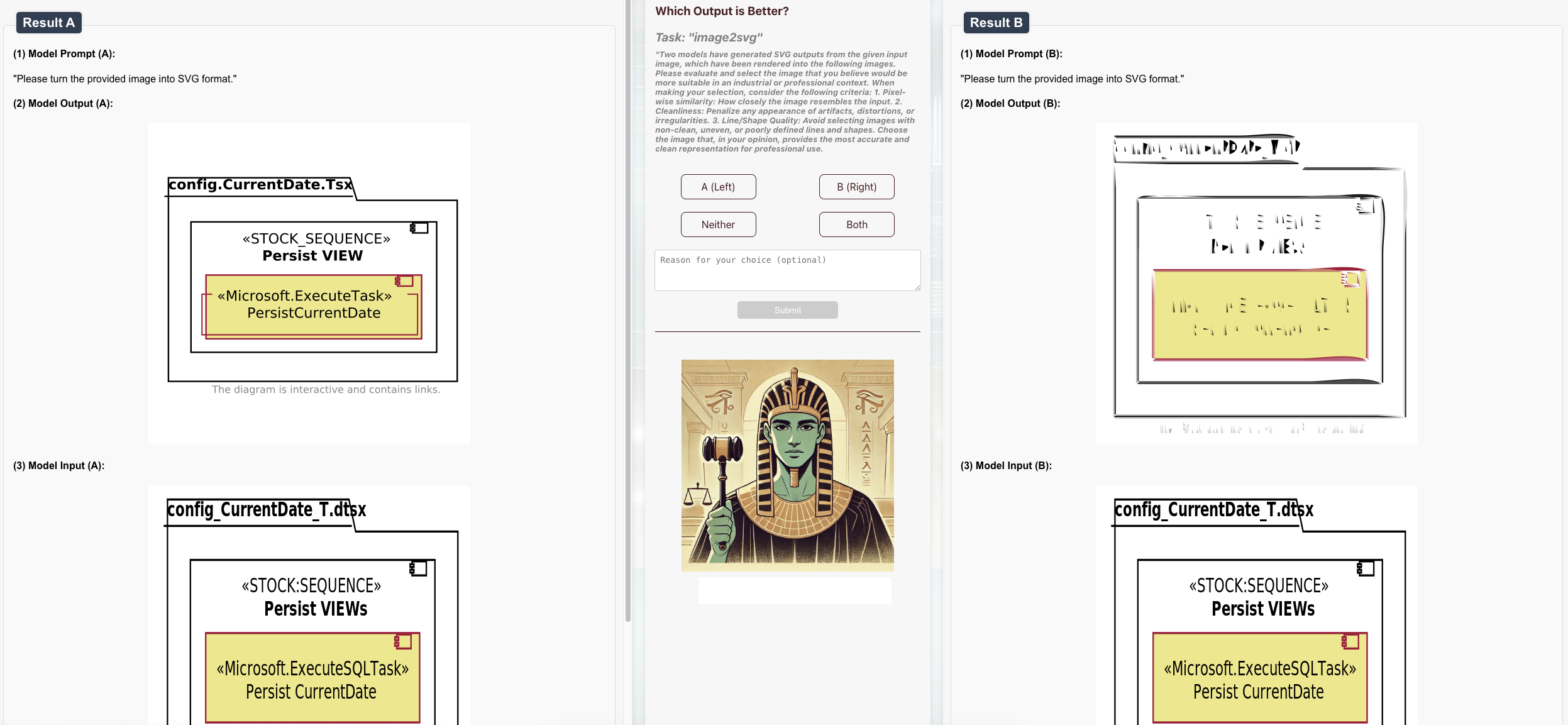}
    \caption{The web interface used during the human evaluation.}
    \label{fig:osiris}
\end{figure*}

The results, presented in Figure~\ref{fig:human-eval}, demonstrate a strong preference for StarVector-8B across all settings, especially in SVG-Diagrams tasks. This highlights a disconnect between pixel-based metrics (e.g., MSE, SSIM) and human visual perception of SVGs. While baseline models often prioritize pixel-perfect reconstruction, human evaluators preferred StarVector's sharp, well-defined shapes and its effective use of primitives (Figure~\ref{fig:tag-stats}).

Spearman correlation analyses between model metrics and human evaluations further emphasize this gap. MSE shows weak correlations (0.0596 and -0.1002), indicating its inadequacy as a predictor of human preferences. In contrast, DinoScore exhibits significantly stronger correlations, with values of -0.6193 and 0.6214. Moreover, a robust correlation of 0.7577 between differences in DinoScore and human evaluation scores highlights DinoScore as a more reliable metric for assessing SVG quality in alignment with human judgment. 

\end{enumerate}

\section{StarVector Method}
\label{app:starvector-architecture}

Here, we provide details on the StarVector architecture, training recipe, and generation process.

\subsection{Architecture}
\subsubsection{Large Language Model}
\label{app:starvector-codellm}
We consider several aspects when choosing the LLM to handle the SVG code generation. First, we require an LLM that can handle large token contexts during training, as SVG code samples are typically of long lengths (i.e., between 1,000-4,000 tokens for the most common SVG datasets but growing arbitrarily for much more complex vector graphics). Second, we need fast decoding during the generation of these large contexts. Finally, we would benefit from models that have been extensively pre-trained on general coding tasks to avoid early training costs. 

Some prior works offer open-source models that fit these requirements. We explored the open-source families of models CodeGen~\cite{nijkamp2022codegen}, StarCoder~\cite{li2023starcoder} and StarCoder2~\citep{lozhkov2024starcoder}. 

We empirically find the StarCoder family to be the most suitable choice for our requirements. StarCoder offers a pre-trained model with 1B parameters (\texttt{starcoderbase-1b}) and a context length of 8k tokens, making it ideal for smaller-scale experiments while maintaining efficient generation speeds. The model employs Multi-Query Attention and a Fill-in-the-Middle objective, trained on 1 trillion tokens, with a context window of 8192 tokens. Its compact size ensures compatibility with GPUs, facilitating data parallelism during training—a crucial benefit when fine-tuning all network parameters, including the memory-intensive image encoder.

The second generation, StarCoder2, extends the context length to 16,384 tokens, presenting an exciting avenue for exploring training on SVGs with longer context requirements. For this, we leverage the 7B parameter version (\texttt{starcoder2-7b}), which incorporates a sliding window attention mechanism of 4,096 tokens and was trained on over 3.5 trillion tokens of code using the same Fill-in-the-Middle objective~\citep{lozhkov2024starcoder}. This enhanced context capacity and training scale make it a promising candidate for scaling SVG-based experiments.

These two types of LLMs define the backbones of our two StarVector variants. StarVector-1B is based on the \texttt{starcoderbase-1b} architecture and weights, while StarVector-8B is based on the \texttt{starcoder2-7b} architecture and weights.

\subsubsection{Image Encoder}
\label{app:starvector-image-encoder}
Our image encoding pipeline computes the images' feature representations using a backbone image encoder and aligns them to the LLM via the Adapter module (see Figure~\ref{fig:starvector}). State-of-the-art image encoders are typically focused on natural images. However, our data contains images of logotypes, icons, fonts, or emojis, which usually contain no background (which we set to white) and mostly constant colors. 

Note that the image encoder is used exclusively for the Image-to-SVG task and is not employed during the Text-to-SVG task. For Image-to-SVG, images must be projected into a representation with the same dimensionality as that of the LLM. We train the model to enable the LLM to ingest these representations and generate SVG code sequentially.

To choose the best encoder, we draw inspiration from the success of pre-trained encoder backbones in downstream computer vision tasks such as classification~\cite{radford2021learning}, retrieval, and generation~\cite{esser2021taming}, including both convolutional and transformer-based models. Specifically, we experiment with CLIP ViT-B/32~\cite{radford2021learning}, ConvNext~\cite{liu2022convnet} (pre-trained on LAION-2B~\cite{schuhmann2022laion}), and VQGAN~\cite{esser2021taming}, which we pre-train on an image reconstruction task using raster images from SVG-Stack. For CLIP, we have $L_v=257$ embeddings, including the CLS token. For VQGAN, we use the pre-quantization layers and flatten them to obtain $L_v=196$ embeddings. For ConvNext, we flatten the last activation map to get $L_v=49$ embeddings.

We explore several image encoders based on different paradigms. VQGAN~\cite{esser2021taming} is based on learning to project images to discrete tokens. First, we fine-tune an Imagenet~\cite{deng2009imagenet}-pretrained VQGAN on the SVG-Stack dataset with the VQ-adversarial reconstruction task. We find that using the features before the quantization yields better results. ConvNext~\cite{liu2022convnet} is a convolutional backbone, which we extract features before pooling. We start from a LAION-2B~\cite{schuhmann2022laion}-pretrained checkpoint. Finally, ViT CLIP~\cite{radford2021learning} is based on the Visual Transformer (ViT) \cite{dosovitskiy2020image} and is well prepared for autoregressive tasks. We extract all output representations. We use a LAION-2B pre-trained model. During the training of StarVector, all the parameters of the image encoders are updated. We find that the best choice is using CLIP. We consider that the gains in performance come from CLIP using more visual tokens (257) than the other image encoders. 

The adapter first projects the features from the original dimensionality $D_v$ to a dimensionality $D_v\times 2$, followed by a Swish non-linear activation function and a linear projection to the LLM dimensionality $D_l$. Finally, we apply a layer normalization~\cite{ba2016layer}. We initialize the adapter parameters with Glorot~\cite{glorot2010understanding}. Dropout~\cite{srivastava2014dropout} of 0.1 is applied at the beginning. These hyperparameters were found using a random search on SVG-Fonts.

Our results show that image resolution is essential to capture fine-grained details like texts or high-frequency patterns. As seen in the SVG-Diagrams dataset in Figure~\ref{fig:diagrams}), diagrams and figures are part of the SVG-Stack dataset and present challenging horizontal or vertical aspect ratios. When images have these aspect ratios, we make the image fit in the $224\times224$ resolution, losing much detail, especially for the OCR capabilities of reading rendered texts and accurately displaying them. 

Additional results comparing image encoders can be found in Figures~\ref{fig:ablation-emoji} and \ref{fig:ablation-fonts}. These results show the boost in precision obtained when using CLIP. VGQAN and ConvNext often fail to capture the image's shape and the \textit{path}'s trajectory. We note that ConvNext performs better than VQGAN. These differences are also due to the differences in the number of parameters. The CLIP ViT-L/14 model that we use consists of 290M parameters, VQGAN consists of 29M, and ConvNext consists of 179M parameters.

Generating SVGs from natural images is out of the scope of this project. However, future work should focus on adapting this approach to natural images, drawing from~\cite{ma2022towards} and~\cite{cai2023leveraging} to create a dataset of natural images and SVG pairs.

The selected image encoder architectures for StarVector include two variants: one with fewer parameters and reduced image resolution, based on the CLIP ViT-B/32 model, which processes images at 224x224 pixels and is utilized in StarVector-1B. The second variant, SigLip ({{\texttt{siglip-so400m302}} patch14-384}), has a larger number of parameters and processes images at a higher resolution of 384x384 pixels, and is employed in StarVector-8B. Given the positive results from the ViT architecture, we chose the SigLip variant due to its demonstrated effectiveness~\citep{zhai2023sigmoid} and the enhanced resolution it provides.

\subsection{Training}
\label{app:training}
For training the StarVector model, we define the task of Image-to-SVG as an inverse rendering problem that converts a raster image (represented with visual tokens) into a sequence of SVG code. This can be viewed as a sequence-to-sequence problem that models the translation between the image and SVG code domains. As detailed in Section~\ref{app:starvector-architecture}, we utilize a CLIP ViT-B/32 for StarVector-1B and SigLip for StarVector-8B as image encoders, along with a non-linear adapter to generate a sequence of visual tokens.

The training process consists of two stages. In the first stage, the Image-to-SVG training phase, we construct sequences of visual tokens (produced by the image encoder and adapter) and SVG tokens, separated by a trigger token, \texttt{<svg-start>}. We train the LLM to learn these sequences on a large SVG-Stack dataset using a basic language modeling loss that calculates the cross-entropy loss in predicting the next token in a sequence based on the previous tokens.

This task enables the model to learn the concept of \textit{drawing} with SVG vectors that resemble the input image. Importantly, this training can occur without supervision in the image domain (i.e., without pixel loss), relying solely on categorical cross-entropy loss for the LLM vocabulary introduced by the next-token prediction task.

In the second stage, we fine-tune the checkpoint from the first stage, which has learned SVG syntax through the Image-to-SVG task, on the Text-to-SVG task. During this phase, the image encoder is disregarded, as it becomes a Text-to-text task where the text instructions and SVG codes can be tokenized and processed directly by the LLM.

\vspace{1cm}\noindent\textbf{{Training Details.}} We trained StarVector-1B on 1 node of 8 A100 80GB GPUs using Accelerate with DeepSpeed stage 2 and StarVector-8B on 8 nodes of 8 H100 80GB GPUs with Fully Shared Data Parallel (FSDP). For Image-to-SVG, we used total batch sizes of 128 and 512 for StarVector-1B and -8B, respectively, a learning rate of 1e-5, and the AdamW optimizer. To optimize memory and computation, we employed bf16 precision, FlashAttention2, and gradient checkpointing. StarVector-1B took 7 days to train, while StarVector-8B took 10 days, with both models completing 2 epochs.

We use HuggingFace Transformers~\cite{wolf2019huggingface} and PyTorch~
\cite{paszke2017automatic} for the implementation. We use a batch size of 2. Images are processed with a resolution of 224x224, as defined by the pre-trained CLIP image encoder, and process a maximum of 8192 tokens, considering the 257 visual tokens and the rest for the SVG tokens. We use gradient batch accumulation of 8 and train on a data parallel setup with 4 A100 80GB GPUs, having an effective batch size of 64. The learning rate is set to $5\times10^{-4}$ for training, using AdamW optimizer~\cite{loshchilov2017decoupled} for approximately five days of training on the SVG-Stack dataset. 

\subsection{Generation}
\label{app:starvector-sampling}

Here, we describe how to sample SVG code from our model. As a decoder-only LLM~\cite{li2023starcoder}, StarVector first computes the key-value (KV) cache using the visual tokens from the image and then produces the initial set of output logits. This stage is often quick because the model can process the entire visual token sequence simultaneously~\cite{spector2023accelerating}. The selected token from the output logits is then input back into the model, which generates logits for the subsequent token. This process is iteratively repeated until the model produces the desired quantity of tokens. Our approach uses architectural improvements for fast decoding, such as FlashAttention~\cite{dao2022flashattention} and Multi-Query Attention~\cite{shazeer2019fast}. We leverage vLLM to improve inference speed.

We perform a grid search on SVG-Emoji and SVG-Icons validation sets to select the correct sampling temperature. The choice of temperature does not strongly impact the results. However, a 1-point increase in performance is observed on CD for SVG-Emoji using temperatures close to 1.0. 

We also present an ablation study of StarVector-1B popular decoding techniques~\cite{holtzman2019curious, murray2018correcting, shao2017generating, vijayakumar2016diverse}. Specifically, we experiment with greedy decoding, beam search, and nucleus sampling with top-$p$. Results are shown in Table~\ref{tab:sampling}. The use of nucleus sampling with top-$p$=0.9 and temperature T=0.5 (no beam search) shows to be the best option. The beam search improves the greedy decoding baseline but does not work well when combined with nucleus sampling, increasing the inference time. In sum, we recommend nucleus sampling~\cite{holtzman2019curious} with top p=0.9 and temperature between 0.5 and 0.9 for the best performance. 

\vspace{1cm}
\noindent\textbf{Are the SVGs valid and compilable?} A common issue when generating SVGs with our approach is that the maximum token length of the LLM might not be sufficient to complete the SVG code, leading to compilation errors. We find that 85\% of the generated SVG fit within the context length and compile successfully. The remaining incomplete samples are post-processed with \texttt{cairosvg} to produce a complete and compilable SVG. However, in some cases, parts of the image may be lost during this process. \textit{With this technique, 100\% of the generated SVGs are valid and compilable}.

\vspace{1cm}
\noindent\textbf{Improving SVG Quality Through Sampling.} The generation process is stochastic, meaning the outputs may sometimes take an incorrect \textit{path}, leading to failed generations or repetitive patterns. To address this, we propose a simple baseline approach: generate $k$ SVG outputs with varying sampling parameters (e.g., by adjusting the temperature), then compare the outputs with the ground truth using a visual metric (we propose DinoScore) to select the most accurate result. In an ablation study conducted on SVG-Stack using StarVector-8B with $k=1$ and $k=5$, we observe a boost in DinoScore by $0.12$. Empirically, after sampling 100 test samples, we find that 32\% of the SVGs are more accurate when using $k=5$, though this increases the generation time by a factor of $k$. The use of vLLM helps mitigate the slower sampling process, as it operates much faster. For further improvement in code generation, previous work has used MCTS techniques~\citep{belouadi2023automatikz}, which leverage visual feedback more effectively to guide the generation and enhance the stochastic sampling process.

\end{document}